\documentclass[10pt,journal,compsoc]{IEEEtran}

\ifCLASSOPTIONcompsoc
  \usepackage[nocompress]{cite}
\usepackage[pagebackref=true,breaklinks=true,letterpaper=true,colorlinks,bookmarks=false]{hyperref}
\usepackage{microtype}

\usepackage{amssymb}
\usepackage{amsmath}
\usepackage{amsthm}
\usepackage{mathdots}
\usepackage{mathtools}
\usepackage{multicol}

\usepackage{tensor}

\usepackage{urwchancal}
\DeclareFontFamily{OT1}{pzc}{}
\DeclareFontShape{OT1}{pzc}{m}{it}{<-> s * [1.10] pzcmi7t}{}
\DeclareMathAlphabet{\mathpzc}{OT1}{pzc}{m}{it}

\usepackage{graphicx}
\usepackage{ifpdf}

\ifpdf
\usepackage{epstopdf}
\PrependGraphicsExtensions{.svg}
\fi

\usepackage{xypic}
\usepackage{booktabs}
\usepackage{subcaption}
\usepackage{array,multirow}
\usepackage{rotating}
\usepackage{makecell}
\usepackage{rotating}
\usepackage{array}
\usepackage{float}
\restylefloat{table}
\usepackage{physics}
\usepackage{algorithm}
\usepackage{algpseudocode}
\usepackage{bbding}
\usepackage{amsmath}
\newtheorem{theorem}{Theorem}[section]

\newtheorem{remark}[theorem]{Remark}

\usepackage{wrapfig}
\newcommand{\eventframe}{event-frame }
\newcommand{\EventFrame}{Event-Frame }
\newcommand{\etal}{\textit{et al}. }

\newcommand{\eg}{\textit{e}.\textit{g}., }
\newcommand{\overbar}[1]{\mkern 1.5mu\overline{\mkern-1.5mu#1\mkern-1.5mu}\mkern 1.5mu}

\newcommand{\kernelwidth}{0.25\textwidth}
\else
  \usepackage{cite}

\fi

\ifCLASSINFOpdf
\else
\fi

\usepackage{todonotes}

\usepackage{booktabs}
\usepackage{array,multirow}
\usepackage{rotating}
\usepackage{makecell}
\usepackage{rotating}
\usepackage{array}
\usepackage{float}
\usepackage{physics}
\usepackage{algorithm}
\usepackage{algpseudocode}
\usepackage{enumitem}
\usepackage{epsfig}
\usepackage{graphicx}
\usepackage{amsmath}
\usepackage{amssymb}
\usepackage{pifont}
\newcommand{\xmark}{\ding{55}}%
\usepackage{mathtools}
\usepackage{bm}
\usepackage{footnote}

\IEEEpubid{\makebox[\columnwidth]{%
		0162-8828 © 2023 IEEE. Personal use is permitted, but republication/redistribution requires IEEE permission.
		\hfill}%
	\hspace{\columnsep}\makebox[\columnwidth]{ }}

\begin{document}

\title{An Asynchronous Linear Filter Architecture for Hybrid Event-Frame Cameras}

\author{Ziwei Wang, Yonhon Ng, Cedric Scheerlinck, Robert Mahony
	\thanks{
	Manuscript received 20 August 2021; revised 28 April 2023; accepted 17 August 2023. Date of publication 4 September 2023; date of current version
	8 January 2024. This work was supported by the Australian Research Council
	through the “Australian Centre of Excellence for Robotic Vision” CE140100016.
	Recommended for acceptance by R. P. Wildes. (Corresponding author: Ziwei
	Wang.)
	\\
	The authors are with the Systems Theory and Robotics (STR) Group, College
	of Engineering and Computer Science, Australian National University, Can-
	berra, ACT 2601, Australia (e-mail: ziwei.wang1@anu.edu.au; yonhon.ng@
	anu.edu.au; cedric.scheerlinck@anu.edu.au; robert.mahony@anu.edu.au).
	\\	
	 Published in: IEEE Transactions on Pattern Analysis and Machine Intelligence (Volume: 46, Issue: 2, February 2024). \\
	Digital Object Identifier 10.1109/TPAMI.2023.3311534.	
}%
	\\
	{\tt\small Project URL: \color{red}\url{https://github.com/ziweiwwang/Event-Asynchronous-Filter}}
}

\IEEEtitleabstractindextext{%
\begin{abstract}
Event cameras are ideally suited to capture High Dynamic Range (HDR) visual information without blur but provide poor imaging capability for static or slowly varying scenes.
Conversely, conventional image sensors measure absolute intensity of slowly changing scenes effectively but do poorly on HDR or quickly changing scenes.
In this paper, we present an asynchronous linear filter architecture, fusing event and frame camera data, for HDR video reconstruction and spatial convolution that exploits the advantages of both sensor modalities.
The key idea is the introduction of a state that directly encodes the integrated or convolved image information and that is updated asynchronously as each event or each frame arrives from the camera.
The state can be read-off as-often-as and whenever required to feed into subsequent vision modules for real-time robotic systems.
Our experimental results are evaluated on both publicly available datasets with challenging lighting conditions and fast motions, along with a new dataset with HDR reference that we provide.
The proposed AKF pipeline outperforms other state-of-the-art methods in both absolute intensity error (69.4\% reduction) and image similarity indexes (average 35.5\% improvement).
We also demonstrate the integration of image convolution with linear spatial kernels Gaussian, Sobel, and Laplacian as an application of our architecture.
\end{abstract}

\begin{IEEEkeywords}
Hybrid Event Cameras, High Dynamic Range, Asynchronous Filter, Video Reconstruction, Spatial Convolutions
\end{IEEEkeywords}}

\maketitle

\IEEEdisplaynontitleabstractindextext

%
\IEEEpeerreviewmaketitle

\IEEEraisesectionheading{\section{Introduction}\label{sec:introduction}}

\IEEEPARstart{E}vent cameras offer distinct advantages over conventional frame-based cameras: high temporal resolution (HTR), high dynamic range (HDR) and minimal motion blur \cite{lichtsteiner2008128}.
However, event cameras provide poor imaging capability for slowly varying or static scenes, where despite some work on developing `grey-level' event cameras that measure absolute intensity \cite{posch2010qvga,Chen19cvprw}, most sensors predominantly measure only the relative intensity change.
Conventional imaging technology, conversely, is ideally suited to imaging static scenes and measuring absolute intensity.
Hybrid sensors such as the Dynamic and Active Pixel Vision Sensor (DAVIS) \cite{brandli2014240} or custom-built systems \cite{wang2020joint,han2020neuromorphic,tulyakov2021time,wang2021stereo} combine event and frame-based cameras, and there is an established literature in video reconstruction fusing conventional and event camera data \cite{Brandli14iscas,Pan20pami,han2020neuromorphic,tulyakov2021time} that the present paper builds on.
The potential of such algorithms to enhance conventional video to overcome motion blur and increase dynamic range has applications from robotic vision systems (\eg autonomous driving), through film-making, to smartphone applications for everyday use.

\begin{figure}
	\centering
	\resizebox{0.49\textwidth}{!}{
		\begin{tabular}{ c c }
		\includegraphics[width=0.49\linewidth, trim={1cm 1cm 1cm 1cm},clip]{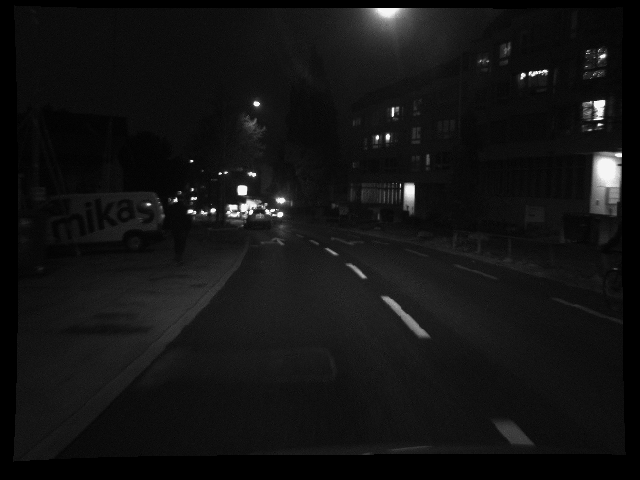} &
		\includegraphics[width=0.49\linewidth, trim={1cm 1cm 1cm 1cm},clip]{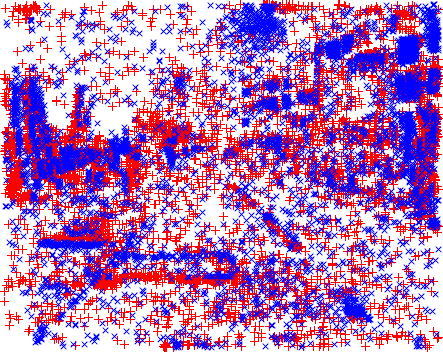}
		\\
		(a) Input LDR & (b) Input Events
		\\
		\includegraphics[width=0.49\linewidth, trim={1cm 1cm 1cm 1cm},clip]{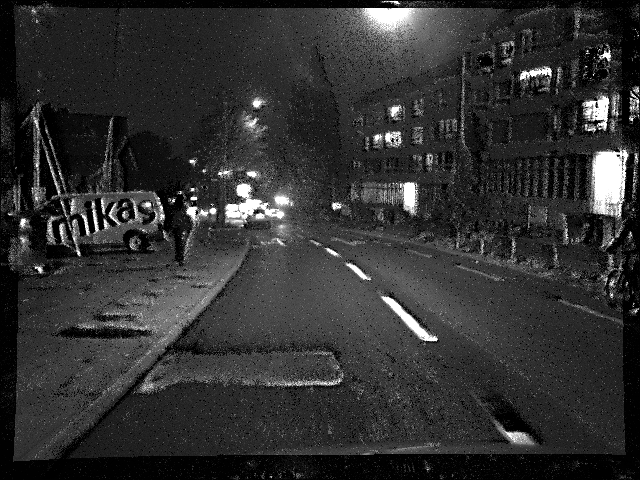} &
		\includegraphics[width=0.49\linewidth, trim={1cm 1cm 1cm 1cm},clip]{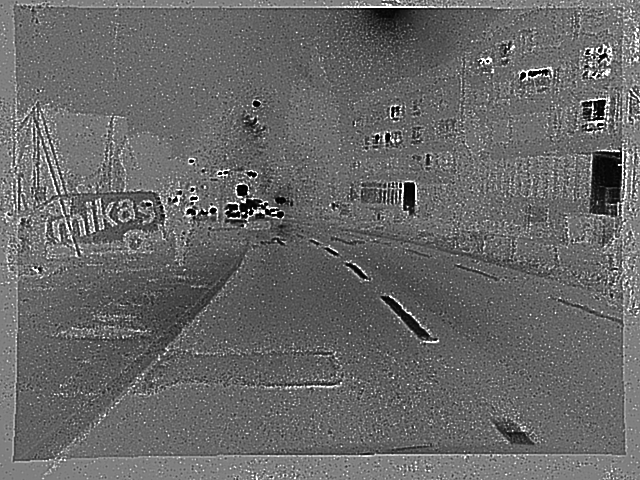}
		\\
		(c) HDR Image Reconstruction  & (d) Laplacian Spatial Convolution
		\end{tabular}
	}
	\caption{
		An example HDR image reconstruction and Laplacian spatial convolution on a driving sequence taken from the open-source stereo \eventframe dataset DSEC~\cite{Gehrig21ral}, city night sequence.
	Image (a) is a raw image from a conventional camera that is low dynamic range (LDR) and blurry.
	Image (b) are the events from a co-located event camera, red plus for positive events and blue cross for negative events.
	Image (c) is our HDR reconstruction that clearly reconstructs sharp objects in a challenging low-light condition.
	Image (d) is our Laplacian spatial convolution result, detecting many edges that are not visible in the input LDR image.
}
	\label{fig:front page}
	\vspace{-2mm}
\end{figure}
Event cameras respond to local changes in brightness rather than absolute brightness levels.
They inherently have a high intra-scene dynamic range since the relative change in irradiance is measured pixel-by-pixel as a ratio yielding log-sensitivity.
Changes in irradiance trigger `events' that are asynchronously measured with microsecond precision.
The minimum brightness change required to trigger an event is called the contrast threshold and this acts as a log-scale quantisation of relative intensity as measured by an event camera.
Blur artefacts caused by motion during the exposure time of a conventional camera sensor are almost eliminated since there is no exposure time, although timing offset due to finite response time of the photoreceptors does lead to some blur like effects, especially in low light conditions \cite{2021_Hu_V2E}.

Current state-of-the-art event-based HDR video reconstruction methods are based on deep learning methods~\cite{Rebecq20pami,Stoffregen20eccv,zhang2020learning,han2020neuromorphic,Zou_2021_CVPR}.
These algorithms rely on accumulation and batching of events in the pre-processing step and do not have the low-latency of true asynchronous algorithms.
Learning algorithms that advertise low latency and high frame rate video do so by reprocessing events multiple times to provide input for each frame reconstruction~\cite{Rebecq20pami,Stoffregen20eccv,Zou_2021_CVPR}.
More recently, deep-learning methods use both events and frames for image reconstruction are developed.
They use intensity map reconstructed from events to guide frame-based HDR reconstruction~\cite{han2020neuromorphic}, align bracket LDR images taken at different exposures~\cite{messikommer2022multi} or interpolate between frames~\cite{tulyakov2021time,Tulyakov22cvpr}.
These deep-learning methods require a large amount of training data  and quality of the model depends heavily on the data source.
Training with large amounts of synthetic data tends to lead to poor generalisability \cite{Stoffregen20eccv} increasing the sim-to-real gap.
This is particularly important for event cameras where parameters in the camera significantly change the characteristics of the event data.

In contrast to learning-based methods, explicit algorithms for event-based image reconstruction leverage classical vision and signal processing techniques~\cite{Brandli14iscas,wang19acra,Pan20pami,han2020neuromorphic,tulyakov2021time,Scheerlinck18accv,munda2018real,wang2021asynchronous}.
Since such algorithms are model based, their performance is coupled to the quality of the data (not the training data).
Improved event sensors that are presently underdeveloped will lead to immediate performance gains for explicit algorithms.
As event sensor quality continues to improve, stochastic fusion algorithms, implemented at the chip level, and integrated with hybrid event-frame sensors, have the potential to provide a significant improvement of image quality across a wide range of imaging applications.

In this paper, we propose an asynchronous linear filter architecture to reconstruct continuous-time videos and spatial convolutions.
The core of our architecture is based on a Complementary filter (CF)~\cite{Scheerlinck18accv} that fuses frames and event data in real-time at a pixel-by-pixel level.
The method does not depend on a motion-model, and works well in highly dynamic, complex environments.
Our approach retains the HTR and HDR information in the event data, producing an image state with greater temporal resolution and dynamic range than the input frames.
The result is a continuous-time estimate of intensity that can be queried locally or globally at any time, effectively at infinite temporal resolution, although the state is only updated when new events or frames are available, introducing some level of temporal quantisation.
To obtain better HDR reconstruction performance, we introduce a unifying noise model incorporating both event and frame data.
The noise model provides a stochastic framework in which the pixel intensity estimation can be solved using an Extended Kalman Filter (EKF) algorithm \cite{kalman1960new,kalman1961new}.
By exploiting the temporal quantisation of the event stream, we propose an exact discretisation of the EKF equations leading to a fully asynchronous implementation that updates whenever event or frame data is available.
We term the resulting algorithm the \emph{Asynchronous Kalman Filter (AKF)}~\cite{wang2021asynchronous} and note that it can be viewed as dynamically updating the gains of the CF~\cite{Scheerlinck18accv} based on the principles of stochastic data fusion.
In addition, we propose a novel temporal interpolation scheme and apply the established deblurring algorithm \cite{Pan20pami} to preprocess frame data in a step called \emph{frame augmentation} to improve temporal resolution.

Our proposed architecture can also compute asynchronous image spatial convolution from event camera outputs without generating pseudo-images.
Exploiting linearity, the spatial convolution is factored through the linear filter architecture and applied directly to the input event and frame data.
Each event is spatially convolved with a linear kernel to generate a neighbouring collection of events, all with the same timestamp.
The convolved event stream is fed into the pixel-by-pixel AKF and fused with convolved frame data to generate high quality HTR and HDR estimates of the linearly convolved input scene.
We demonstrate our approach using a variety of common kernels, including Gaussian, Sobel and Laplacian kernels, bypassing the intermediate step of generating a reconstructed image.
Convolved image primitives such as these are core input to many general computer vision and robotics applications such as detection, tracking, scene understanding, etc.
As our asynchronous architecture is suitable for implementation on front-end hardware (\eg FPGA and ASIC), it has great potential for high-performance embedded systems such as virtual reality systems.

Despite recent interest in event-based HDR image reconstruction \cite{wang2020joint, Zou_2021_CVPR}, to the best of the author's knowledge, there are no open-source targeted event-frame datasets with HDR reference on which to quantitatively evaluate HDR reconstruction.
We addressed this by building a stereo hybrid event frame sensor consisting of an RGB frame-based camera mounted alongside an event camera to collect events, frames and HDR ground truth.
Our system provides data with higher resolution, higher dynamic range, higher framerate, and the stereo configuration overcomes `shutter noise' of the popular existing datasets obtained with hybrid monocular event-frame cameras.

In summary, our contributions are:
\begin{itemize}
	\item An asynchronous linear filter architecture for event cameras to reconstruct continuous HDR videos and spatial convolutions, consisting of four separable modules in the architecture: (i) a Complementary Filter (CF) or Asynchronous Kalman Filter (AKF) for asynchronous hybrid \eventframe video reconstruction, (ii) an asynchronous Kalman gain solver that dynamically adjusts filter gain to produce better HDR videos, (iii) a frame augmentation module for video deblur and temporal interpolation,
	(iv) a real-time spatial convolution module designed for hybrid event-frame cameras, with examples shown for Gaussian, Sobel and Laplacian kernels.
	\item Bringing three asynchronous linear filters~\cite{Scheerlinck18accv,wang2021asynchronous,Scheerlinck19ral} into a unified framework and providing detailed empirical analysis.
	\item A unifying \eventframe uncertainty model.
	\item A novel HDR hybrid \eventframe dataset with reference HDR images for qualitative performance evaluation.
\end{itemize}
The proposed algorithm demonstrates state-of-the-art hybrid \eventframe HDR video reconstruction and spatial convolution as shown in Fig.~\ref{fig:front page}.

\section{Related Work}
Recognising the limited ability of pure event cameras (DVS) \cite{lichtsteiner2008128} to detect slow/static scenes and absolute brightness, hybrid \eventframe cameras such as the DAVIS \cite{brandli2014240} were developed.
Image frames and events are captured through the same photodiode allowing the two complementary data streams to be precisely registered \cite{Brandli14iscas}.
This has led to significant research effort into image reconstruction from hybrid \eventframe and pure event cameras including SLAM-based methods \cite{Kim16eccv,Rebecq17ral}, deblurring \cite{pan2019bringing,Pan20pami}, tracking \cite{Tedaldi16ebccsp,Liu16iscas,Gehrig19ijcv}, disparity estimation~\cite{wang2021stereo} and video reconstructions \cite{Brandli14iscas,Pini19iciap,Rebecq20pami,Scheerlinck20wacv,Stoffregen20eccv}.

Video and image reconstruction methods may be grouped into (i) per-event asynchronous algorithms that process events upon arrival \cite{Brandli14iscas,wang19acra} and (ii) batch (synchronous) algorithms that first accumulate a significant number (\eg 10k) of events before processing the batch in one go \cite{Pini19iciap, Rebecq20pami,Scheerlinck20wacv}.
While batch methods have achieved high accuracy, they incur additional latency depending on the time interval of the batch (\eg 50ms).
Asynchronous methods, if implemented on appropriate hardware and software, have the potential to run on a timescale closer to that of events $< 1$ms, which has recently shown great potential in high-speed optical communication~\cite{wang22iros} and high-frequency signal processing~\cite{sekikawa2019eventnet, wang22icra, ng2023asynchronous}.
A further distinction may be made between pure event reconstruction methods and hybrid \eventframe methods that use a mix of (registered) events and image frames.

\noindent
\textbf{Pure event reconstruction:}
Images and video reconstruction using only events is a topic of significant interest in the community that can shed light on the information content of events alone.
Early work focused on a moving event camera in a static scene, either pure rotations \cite{Cook11ijcnn,Kim14bmvc} or full 6-DOF motion \cite{Kim16eccv,Rebecq17ral}.
Hand-crafted approaches were proposed including joint optimisation over optic flow and image intensity \cite{Bardow16cvpr}, periodic regularisation based on event timestamps \cite{Reinbacher16bmvc}.
Some work~\cite{mondal2021moving, schaefer2022aegnn, zhang2022eventmd} built sparse event graphs for moving object detection using event data.
Recently, learned approaches have achieved state-of-the-art high quality video reconstruction \cite{Rebecq19cvpr,Rebecq20pami,Scheerlinck20wacv,Stoffregen20eccv,Zou_2021_CVPR} at significantly higher computational cost.

\noindent
\textbf{Event-frame reconstruction:}
The invention of the DAVIS \cite{brandli2014240} and its ability to capture frames alongside events (and even IMU measurements) has widened the community's perspective from pure event cameras to hybrid sensors and how best to combine modalities.
An early algorithm interpolated between frames by adding events scaled by the contrast threshold until a new frame is received  \cite{Brandli14iscas}.
The contrast threshold is typically unknown and variable, so \cite{Brandli14iscas} included a method to estimate it based on surrounding image frames from the DAVIS.
Wang \etal \cite{wang19acra} proposed both pure event and hybrid contrast threshold estimation algorithms and reconstruct videos using direct integration.
Pan \etal \cite{pan2019bringing,Pan20pami} devised the event double integral (EDI) relation between events and a blurry image, along with an optimisation approach to estimate contrast thresholds to reconstruct high-speed de-blurred video from events and frames.
Reconstructed video can also be obtained by warping still images according to motion computed via events \cite{Shedligeri19jei,Liu17tvc}, or by letting a neural network learn how to combine frames and events \cite{Pini20visigrapp,Wang19iccvw,Pini19iciap,Lin20eccv,Jiang20cvpr,zhang2020learning}.
Recognising the limited spatial resolution and noise of the DAVIS, some researchers built custom-built systems with separate event-frame sensors.
Han \etal \cite{han2020neuromorphic} built a hybrid event-frame system consisting of an RGB camera and a DAVIS240 event camera registered via a beam-splitter.
They achieved state-of-the-art event-frame-based HDR reconstruction results by using pure events to reconstruct intensity images (using a known algorithm~\cite{Rebecq20pami}) and fusing this HDR information with LDR absolute intensity images. 
Wang \etal \cite{wang2021stereo} mounted an event camera and an RGB camera side-by-side on a UR5 robot arm to collect indoor disparity estimation datasets with camera pose and disparity ground truth.
Recently, TimeLens~\cite{tulyakov2021time} and TimeLens++~\cite{Tulyakov22cvpr} achieved the state-of-the-art Video Frame Interpolation (VFI) performance fusing events and frame data.
They interpolate forwards and backwards using events in-between timestamps between two consecutive input frames with novel warping and synthesis modules to generate high speed frame rate video.
The input frames are treated as key frames and the algorithms do not improve HDR or undertake image deblurring.

\noindent
\textbf{Multi-exposure image fusion (MEIF):}
Ignoring the events initially it is possible to use multi-exposure frames to construct HDR images and then use the event to temporally interpolate.
Ma~\emph{\etal}~\cite{Ma17TIP} proposed the use of structural patch decomposition to handle dynamic objects in the scene.
Kalantari and Ramamoorthi~\cite{Kalantari17ACMTG} proposed a deep neural network and a dataset for dynamic HDR MEIF.
More recent work also deals with motion blur in long exposure images~\cite{Wang18ICSIP, Li18ICIP}.
These methods directly compute images that do not require additional tone mapping to produce nice-looking images~\cite{Que19TIM}.
Recently, for event-frame high dynamic range reconstruction,
Messikommer \etal \cite{messikommer2022multi} aligned and fused bracketed LDR images with different exposure times and events are used to improve the alignment of LDR images in scenes.		
However, all these works require multiple images at different exposures of the same scene and do not apply to the real-time reconstruction scenarios considered in this paper.

\noindent
\textbf{Spatial Event-based Convolutions:}
Camunas-Mesa \etal \cite{camunas2011event} introduced an event-driven multi-kernel convolution processor module and provided a proof-of-concept prototype in a CMOS process.
Ieng \etal \cite{Ieng14ieee} proposed a method to asynchronously compute spatial convolutions that relies on grey-level events (provided by the ATIS \cite{Posch11ssc}), and showed that beyond 3 frames per second, asynchronous convolutions outperform frame-based in computational cost.
Sabatier \etal \cite{Sabatier17tip} performed an asynchronous Fourier transform with the ATIS, demonstrating improved computational efficiency compared to frame-based methods.
Huang \etal \cite{Huang18iscas} proposed an on-chip module that causes events to trigger neighbouring pixels for gradient computation.
An alternative approach to event-based, 2D image filtering is the VLSI architecture proposed in \cite{serrano1999aer}, capable of implementing any convolutional kernel that is decomposable into $x$ and $y$ components.

\section{Sensor Model and Uncertainty}
\subsection{Event Camera Model}\label{sec:Event Camera Mode}
Event cameras measure the relative log intensity change of irradiance of pixels.
New events $e_{\bm{p}}$ are triggered when the log intensity change exceeds a preset contrast threshold $c$.
In this work, we model event stream as a Dirac delta or impulse function $\delta$ \cite{hassani2008mathematical} to allow us to apply continuous time systems analysis for filter design.
That is,
\begin{align}
e_{\bm{p}}(t) &=\sum_{s=1}^\infty
(c \sigma_{\bm{p}}^s + \eta^s_{\bm{p}}) \delta(t-t^s_{\bm{p}}), \label{eq:continuous_event}
\\
\eta^s_{\bm{p}} &\sim
\mathcal{N}\left(
0, Q_{\bm{p}}(t)
\right), \notag
\end{align}
where $t^s_{\bm{p}}$ is the time of the $s^{th}$ event at the $\bm{p} = (\bm{p}_x ,\bm{p}_y)^{T}$ pixel coordinate, the polarity $\sigma^s_{\bm{p}} \in \{-1,+1\}$ represents the direction of the log intensity change, and the noise $\eta^s_{\bm{p}}$ is an additive Gaussian uncertainty at the instance when the event occurs.
We model the noise covariance $Q_{\bm{p}}(t)$ as the sum of three contributing noise processes: process noise, isolated pixel noise, and refractory period noise.
That is
\begin{align}
\label{eq:event_covariance}
Q_{\bm{p}}(t) :=\sum_{s=1}^\infty\left(Q^\text{proc.}_{\bm{p}}(t) +
Q^\text{iso.}_{\bm{p}}(t) + Q^\text{ref.}_{\bm{p}}(t)
\right)
\delta(t - t^s_{\bm{p}}).
\end{align}
More details are provided in the supplementary material \S 1.
Note that the noise or jitter in event timestamps is found experimentally to be insignificant compared to other noise and is not modelled in this work.

\subsection{Conventional Camera Model}\label{sec:Conventional Camera Model}
The photo-receptor in a CCD or CMOS circuit from a conventional camera converts incoming photons into charge that is then converted to a pixel intensity by an analogue-to-digital converter (ADC).
In a typical camera, the camera response is linearly related to the pixel irradiance for the correct choice of exposure, but can become highly non-linear where pixels are overexposed or underexposed \cite{madden1993extended}.
In particular, effects such as dark current noise, CCD saturation, and blooming destroy the linearity of the camera response at the extreme intensities \cite{kim2008characterization}.
In practice, these extreme values are usually trimmed, since the data is corrupted by sensor noise and quantisation error.
However, the information that can be gained from this data is critically important for HDR reconstruction.
The mapping of the scaled sensor irradiance (a function of scene radiance and exposure time) to the camera response is termed the Camera Response Function (CRF) \cite{grossberg2003space,robertson2003estimation}.
To reconstruct the scaled irradiance $I_{\bm{p}}(\tau^k)$ at pixel $\bm{p}$ at time $\tau^k$ from the corresponding raw camera response $I^F_{\bm{p}}(\tau^k)$ one applies the inverse CRF
\begin{align}
I_{\bm{p}}(\tau^k) &= CRF^{-1}( I^F_{\bm{p}}(\tau^k) )  + \bar{\mu}_{\bm{p}}^k, \label{eq:I_F}
\\
\bar{\mu}_{\bm{p}}^k &\sim \mathcal{N}(0,\bar{R}_{\bm{p}}(\tau^k)), \notag
\end{align}
where $\bar{\mu}_{\bm{p}}^k$ is a noise process that models noise in $I_{\bm{p}}(\tau^k)$ corresponding to noise in $I^F_{\bm{p}}(\tau^k)$ mapped back through the inverse CRF.
This inverse mapping of the noise is critical in correctly modelling the uncertainty of extreme values of the camera response.
More details about the noise covariance $\bar{R}_{\bm{p}}$  is provided in the supplementary material \S 2.

We define the continuous log image intensity function by taking the log of $I_{\bm{p}}(\tau^k)$.
However, the log function is not symmetric and mapping the noise from $I_{\bm{p}}$ will bias the log intensity.
Using Taylor series expansion, the biased log intensity is approximately
\begin{align}
L_{\bm{p}}^F(\tau^k) &\approx \log (I_{\bm{p}}(\tau^k) + I_0) -\frac{\bar{R}_{\bm{p}}(\tau^k)}{2(I_{\bm{p}}(\tau^k) + I_0)^2}  + \mu_{\bm{p}}^k,  \label{eq:L_F} \\
\mu_{\bm{p}}^k &\sim \mathcal{N}(0,R_{\bm{p}}(\tau^k)),
\end{align}
where $I_0$ is a fixed offset introduced to ensure intensity values remain positive and
$R_{\bm{p}}(\tau^k)$ is the covariance of noise associated with the log intensity.
The covariance is given by
\begin{align}
\label{eq:frame_covariance}
R_{\bm{p}}(t) &= \frac{\bar{R}_{\bm{p}}(t)}{(I_{\bm{p}}(\tau^k) + I_0)^2}.
\end{align}
Generally, when $I_{\bm{p}}(\tau^k)$ is not extreme then
$\frac{\bar{R}_{\bm{p}}(t)}{2(I_{\bm{p}}(\tau^k) + I_0)^2} \ll \log(I_{\bm{p}}(\tau^k) + I_0)$ and
$L_{\bm{p}}^F(\tau^k) \approx \log(I_{\bm{p}}(\tau^k) + I_0)$.

\section{Method}\label{sec:method}

\begin{figure*}
	\centering
	\includegraphics[trim=3 5 2 5, clip, width=0.85\linewidth]{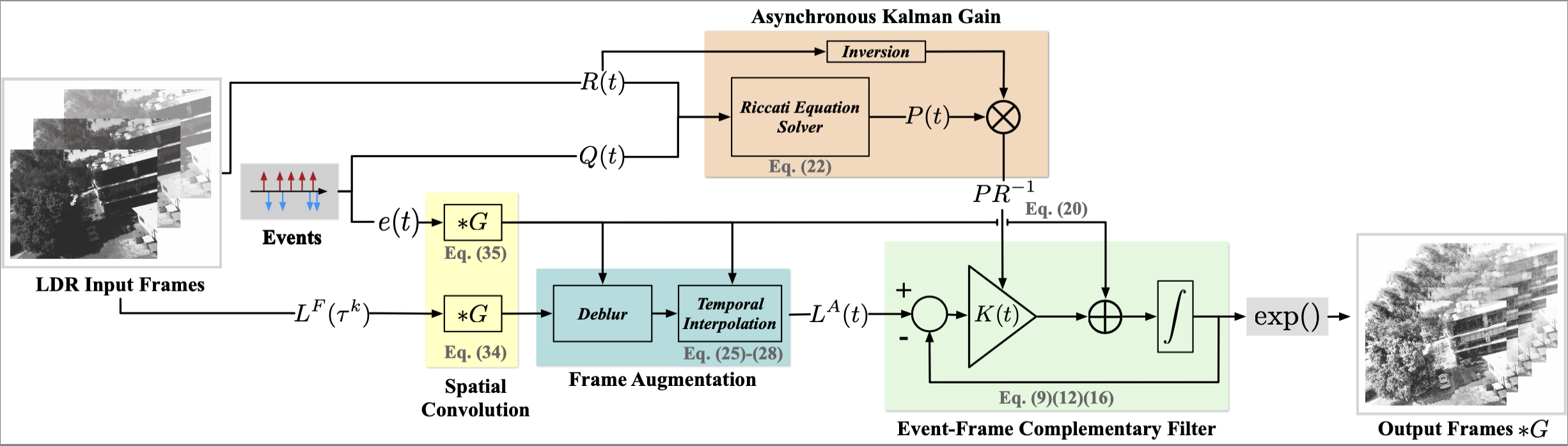}
	\caption{
		Block diagram of the image processing architecture discussed in \S\ref{sec:method}.}
	\label{fig: pipeline}
\end{figure*}

Our image processing architecture consists of four modules.
The block diagram in Fig. \ref{fig: pipeline} visualises the sequence of operations and data flow between modules.
In this section, we will introduce the methodology of each module:	
\begin{itemize}
	\item The Complementary Filter (CF) module fuses the augmented frames with event data to generate continuous-time video in real-time;
	\item The asynchronous Kalman gain module dynamically estimates the filter gain under a unifying noise model for both frame and event sensors;
	\item The frame augmentation module pre-processes the raw frame data to remove blur and increase temporal resolution of the input signal;
	\item The asynchronous spatial convolution module is optional and is used to compute image convolution of the output without reconstructing pseudo-images.
\end{itemize}
The classical Complementary Filter (CF)~\cite{Scheerlinck18accv} is the backbone of the proposed architecture for video reconstruction.
Using only CF, the filter gain is a user-defined constant, while the frame images are not augmented and the output images are not convolved. 

The Asynchronous Kalman Filter (AKF)~\cite{wang2021asynchronous} is obtained by adding the asynchronous Kalman gain module to the classical CF.
Computing the Kalman gain asynchronously over all events allows the filter to dynamically trade-off between event and frame data depending on the proposed unifying event-frame noise model.
As such, the AKF provides superior HDR and sharp video reconstruction at the cost of additional complexity.

Both the CF and AKF architectures depend on the input frame data.  
When the input frames are blurry and have a low-frame rate, it is advantageous to first augment the frames to minimise blur and increase the temporal resolution to the required output frame rate with the frame augmentation module.

The final module, the event-based spatial convolution is optional.
We show that it allows output of Sobel, Gaussian and Laplacian filtered images for general computer vision and robotics applications without computation of intermediary pseudo-frames.

\subsection{Event-frame Complementary Filter}\label{sec: CF}
In this section, we introduce a complementary filter structure~\cite{higgins1975comparison,mahony2008nonlinear,Franklin98digital} to fuse the event stream $e_{\bm{p}}(t)$ with augmented log-intensity frames $L_{\bm{p}}^A(t)$ (see Fig.~\ref{fig: pipeline}).

Complementary filtering is ideal for fusing signals that have complementary frequency noise characteristics; for example, where one signal is dominated by high-frequency noise and the other by low-frequency disturbance.
Events are a temporal derivative measurement and do not contain reference intensity $L_{\bm{p}}^A(0)$ information.
Integrating events to obtain $L_{\bm{p}}^A(t)$ amplifies low-frequency disturbance (drift), resulting in poor low-frequency information.
However, due to their high-temporal-resolution, events provide reliable \emph{high-frequency} information.
Classical image frames are derived from discrete, temporally-sparse measurements and have poor high-frequency fidelity.
However, frames typically provide reliable \emph{low-frequency} reference intensity information.
The proposed complementary filter combines a temporal \emph{low-pass} version of $L_{\bm{p}}^A(t)$ with a temporal \emph{high-pass} version of $\int_{0}^{t} e_{\bm{p}}(\gamma) d\gamma$ to reconstruct an (approximate) all-pass version of $\hat{L}_{\bm{p}}(t)$ using a constant gain $\alpha$ for all pixels.

The proposed filter is written as a continuous-time ordinary differential equation (ODE)
\begin{align}
\boxed{
	 \dot{\hat{L}}_{\bm{p}}(t) =  e_{\bm{p}}(t) - \alpha \big(\hat{L}_{\bm{p}}(t) - L_{\bm{p}}^A(t)\big),
}
\label{eq:complementary}
\end{align}
where $\hat{L}_{\bm{p}}(t)$ is the continuous-time log-intensity state estimate and $\alpha$ is the complementary filter gain, or crossover frequency \cite{mahony2008nonlinear}.
The filter can be understood as integration of the event field with an innovation term
$-\alpha \big(\hat{L}_{\bm{p}}(t) - L_{\bm{p}}^A(t)\big)$, that acts to reduce the error between $\hat{L}_{\bm{p}}(t)$ and $L_{\bm{p}}^A(t)$.

The key property of the proposed Complementary filter (CF) in \eqref{eq:complementary} is that although it is posed as a continuous-time ODE, one can express the solution as a set of asynchronous-update equations.
The algorithm is fully decoupled and the intensity state for each pixel $\bm{p}$ is separately computed.

Consider a combined sequence of monotonically increasing \emph{unique} timestamps $t^i_{\bm{p}}$ obtained by interleaving the event data $\{t^s_{\bm{p}}\}$ \eqref{eq:continuous_event} with the timestamps of the frame data $\{\tau_k\}$ timestamps \eqref{eq:I_F} and relabelling.
Within a time-interval \mbox{$t \in [t^i_{\bm{p}}, t^{i+1}_{\bm{p}})$} there are (by definition) no new events or frames, and the ODE \eqref{eq:complementary} is a constant coefficient linear ordinary differential equation
\begin{align}
\dot{\hat{L}}_{\bm{p}}(t)
& = -\alpha \big( \hat{L}_{\bm{p}}(t) - L_{\bm{p}}^A(t) \big), \quad t \in [t^i_{\bm{p}}, t^{i+1}_{\bm{p}}).
\end{align}
The solution to this ODE is given by
\begin{align}
\hat{L}_{\bm{p}}(t) &= e^{- \alpha (t - t^i_{\bm{p}})} \hat{L}_{\bm{p}}(t^i_{\bm{p}}) + (1 - e^{- \alpha  (t - t^i_{\bm{p}})}) L_{\bm{p}}^A(t), \label{eq:solution} \\
t &\in [t^i_{\bm{p}}, t^{i+1}_{\bm{p}}). \notag
\end{align}
It remains to paste together the piece-wise smooth solutions on the half-open intervals $[t^i_{\bm{p}}, t^{i+1}_{\bm{p}})$ by considering the boundary conditions.
Let
\begin{align}
(t^{i+1}_{\bm{p}})^- &:= \lim_{t \to (t^{i+1}_{\bm{p}})} t, \quad \text{for } t < t^{i+1}_{\bm{p}}
\\
(t^{i+1}_{\bm{p}})^+ &:= \lim_{t\to (t^{i+1}_{\bm{p}})} t,  \quad \text{for }  t >  t^{i+1}_{\bm{p}},
\end{align}
denote the limits from below and above.
There are two cases to consider:

\medskip \noindent \textbf{New frame:}
When the index $t^{i+1}_{\bm{p}}$ corresponds to a new image frame then the right hand side (RHS) of \eqref{eq:complementary} has bounded variation.
It follows that the solution is continuous at $t^{i+1}_{\bm{p}}$ and
\begin{align} \label{eq:solution_frame}
\hat{L}_{\bm{p}}(t^{i+1}_{\bm{p}}) = \hat{L}_{\bm{p}}(t^{i+1}_{\bm{p}})^{-}.
\end{align}

\medskip \noindent \textbf{Event:}
When the index $t^{i+1}_{\bm{p}}$ corresponds to an event then the solution
of \eqref{eq:complementary} is \emph{not} continuous at $t^{i+1}_{\bm{p}}$ and the Dirac delta function of the event must be integrated.
Integrating the RHS and LHS of (\ref{eq:complementary}) over an event
\begin{align}
\int_{(t^{i+1}_{\bm{p}})^-}^{(t^{i+1}_{\bm{p}})^+} \frac{d}{d \gamma} \hat{L}&_{\bm{p}}(\gamma)  d \gamma \\
= \int_{(t^{i+1}_{\bm{p}})^-}^{(t^{i+1}_{\bm{p}})^+}& e_{\bm{p}}(\gamma) - \alpha \big(\hat{L}_{\bm{p}}(\gamma) - L_{\bm{p}}^A(\gamma) \big) d \gamma,
\end{align}
which yields a unit step scaled by the contrast threshold and sign of the event as
\begin{align}
\hat{L}_{\bm{p}}(t^{i+1}_{\bm{p}})^+& -
\hat{L}_{\bm{p}}(t^{i+1}_{\bm{p}})^-
= c \sigma^{i+1}_{\bm{p}}.
\end{align}\label{eq:update event}
Note the integral of the second term $\int_{(t^{i+1}_{\bm{p}})^-}^{(t^{i+1}_{\bm{p}})^+} \alpha \big(\hat{L}_{\bm{p}}(\gamma) - L_{\bm{p}}^A(\gamma) \big) d \gamma$ is zero since the integrand is bounded.
We use the solution
\begin{align} \label{eq:solution_event}
\hat{L}_{\bm{p}}(t^{i+1}_{\bm{p}}) =
\hat{L}_{\bm{p}}(t^{i+1}_{\bm{p}})^- + c \sigma^{i+1}_{\bm{p}},
\end{align}
as initial condition for the next time-interval.
Equation \eqref{eq:solution}, \eqref{eq:solution_frame} and \eqref{eq:solution_event} characterise the full solution to the filter equation \eqref{eq:complementary}.

\begin{remark}
The filter can be also run on \emph{events only} without image frames by setting $L_{\bm{p}}^A(t) = 0$ in \eqref{eq:complementary}, resulting in a high-pass filter with corner frequency $\alpha$
	\begin{align}
		 \dot{\hat{L}}_{\bm{p}}(t) =  e_{\bm{p}}(t) - \alpha \hat{L}_{\bm{p}}(t).
	\label{eq:high-pass}
\end{align}
This method can efficiently generate a reasonable quality image state estimate from pure events and is useful in a range of applications where high image quality is not critical.
\end{remark}

\subsection{Asynchronous Kalman Filter}
In this section, we introduce the Kalman gain module that integrates the uncertainty models of both event and frame data to compute the filter gain dynamically.
We propose a continuous time stochastic model of the log intensity state
\begin{align*}
\mathrm{d} L_{\bm{p}}
&= e_{\bm{p}}(t)\mathrm{d}t + \mathrm{d} w_{\bm{p}}, \\
L^A_{\bm{p}}(t^{i}_{\bm{p}})  &=
L_{\bm{p}}(t^{i}_{\bm{p}}) + \mu_{\bm{p}}^i,
\end{align*}
where $\mathrm{d} w_{\bm{p}}$ is a Wiener process (continuous-time stochastic process) and $\mu_{\bm{p}}^i$ is the log intensity frame noise \eqref{eq:L_F} in continuous time associated with the models introduced in \S\ref{sec:Event Camera Mode} and \S\ref{sec:Conventional Camera Model} and the supplementary material \S 1 and \S 2.
Here $L^A(t)$ is the augmented image (see Fig.~\ref{fig: pipeline})
and the notation serves also as the measurement equation where $L_{\bm{p}}(t^{i}_{\bm{p}})$ is the true (log) image intensity.
The approach taken is to implement a Kalman-Bucy filter to generate an estimate $\hat{L}_{\bm{p}}(t)$ for the true state $L_{\bm{p}}(t)$.
In common with all Kalman-Bucy filters the resulting algorithm consists of a copy of the system model with correction that computes the estimate $\hat{L}_{\bm{p}}(t)$ depending on a Kalman gain $K_{\bm{p}}(t)$ that is computed in parallel by a Riccati equation.

The continuous time model for the state estimate is
\begin{align}
\dot{\hat{L}}_{\bm{p}}(t) &= e_{\bm{p}}(t) - K_{\bm{p}}(t)[\hat{L}_{\bm{p}}(t) - L_{\bm{p}}^A(t)],
\label{eq:filter_ODE}
\end{align}
where $K_{\bm{p}}(t)$ is the Kalman gain defined below \eqref{eq:kf}.
Analogously to the complementary filter we solve this ordinary different equation in a series of time intervals $t \in [t^i_{\bm{p}}, t^{i+1}_{\bm{p}})$.
Substituting the Kalman gain $K_{\bm{p}}(t)$ from \eqref{eq:kf} and \eqref{eq:P_full}, the analytic solution of \eqref{eq:filter_ODE} between frames or events is
\begin{align} \label{eq:L_hat_full}
\hat{L}_{\bm{p}}(t) = \frac{ [\hat{L}_{\bm{p}}(t^{i}_{\bm{p}}) - L_{\bm{p}}^A(t^{i}_{\bm{p}})] \cdot P_{\bm{p}}^{-1}(t^{i}_{\bm{p}})}{P_{\bm{p}}^{-1}(t^{i}_{\bm{p}}) + R_{\bm{p}}^{-1}(t) \cdot  (t - t^{i}_{\bm{p}})} + L_{\bm{p}}^A(t).
\end{align}
A detailed derivation of $\hat{L}_{\bm{p}}(t)$ is provided in the supplementary material \S 7.
The discrete update at event and frame times $t^i_{\bm{p}}$ is identical to the complementary filter; either \eqref{eq:solution_frame} for frame data or \eqref{eq:solution_event} for event data.

\subsubsection{Asynchronous Kalman Gain}\label{sec:KalmanGain}
The Asynchronous Kalman filter computes a pixel-by-pixel gain $K_{\bm{p}}(t)$ derived from estimates of the state and sensor uncertainties.
The Kalman gain is given by \cite{kalman1960new,kalman1961new}
\begin{align}
\label{eq:kf}
K_{\bm{p}}(t) = P_{\bm{p}}(t) R_{\bm{p}}^{-1}(t),
\end{align}	
where $P_{\bm{p}}(t) > 0$ denotes the covariance of the state estimate in the filter and $R_{\bm{p}}(t)$ \eqref{eq:frame_covariance} is the log-intensity frame covariance of pixel $\bm{p}$.
The standard Riccati equation \cite{kalman1960contributions,zaitsev2002handbook} that governs the evolution of the filter state covariance~\cite{kalman1961new} is given by
\[
\dot{P}_{\bm{p}} = -P_{\bm{p}}^2 R_{\bm{p}}^{-1}(t) + Q_{\bm{p}}(t),
\]
where $Q_{\bm{p}}(t)$ \eqref{eq:event_covariance} is the event noise covariance.
Here the choice of event noise model \eqref{eq:event_covariance} as a discrete noise that occurs when the update of information occurs means that the Riccati equation can also be solved during the time interval $t \in [t^i_{\bm{p}}, t^{i+1}_{\bm{p}})$ and at new event timestamp $t^{i+1}_{\bm{p}}$ separately.

In the time interval $t \in [t^i_{\bm{p}}, t^{i+1}_{\bm{p}})$ (no new events or frames occur), the
state covariance $P_{\bm{p}}(t)$ is asynchronously updated by the ordinary differential equation
\begin{align}
\dot{P}_{\bm{p}}(t) = - P_{\bm{p}}^2(t) \cdot R_{\bm{p}}^{-1}(t). \label{eq:Riccati_ODE1}
\end{align}
Since $R_{\bm{p}}(t)$ is constant on this time interval then the solution of \eqref{eq:Riccati_ODE1} is
\begin{align}
P_{\bm{p}}(t) &= \frac{1}{P_{\bm{p}}^{-1}(t^{i}_{\bm{p}}) + R_{\bm{p}}^{-1}(t) \cdot  (t - t^{i}_{\bm{p}})}, \notag \\
\text{ for }
t & \in [t^i_{\bm{p}}, t^{i+1}_{\bm{p}}).
\label{eq:P_full}
\end{align}
At the new event timestamp $t^{i+1}_{\bm{p}}$, the
state covariance $P_{\bm{p}}(t)$ is given by
\begin{align}
P_{\bm{p}}(t^{{i+1}}_{\bm{p}}) & = P_{\bm{p}}(t^{{(i+1)}-}_{\bm{p}}) + Q_{\bm{p}}(t^{i+1}_{\bm{p}}). \label{eq:P_event}
\end{align}
The explicit solution of Kalman filter gain is obtained by substituting \eqref{eq:P_full} and \eqref{eq:P_event} to \eqref{eq:kf}. See derivation of $P_{\bm{p}}(t)$ in the supplementary material \S 8.
The solution is substituted into \eqref{eq:filter_ODE} to obtain \eqref{eq:L_hat_full}.

\subsection{Frame Augmentation}

\begin{figure}
	\centering
	\includegraphics[width=1\linewidth]{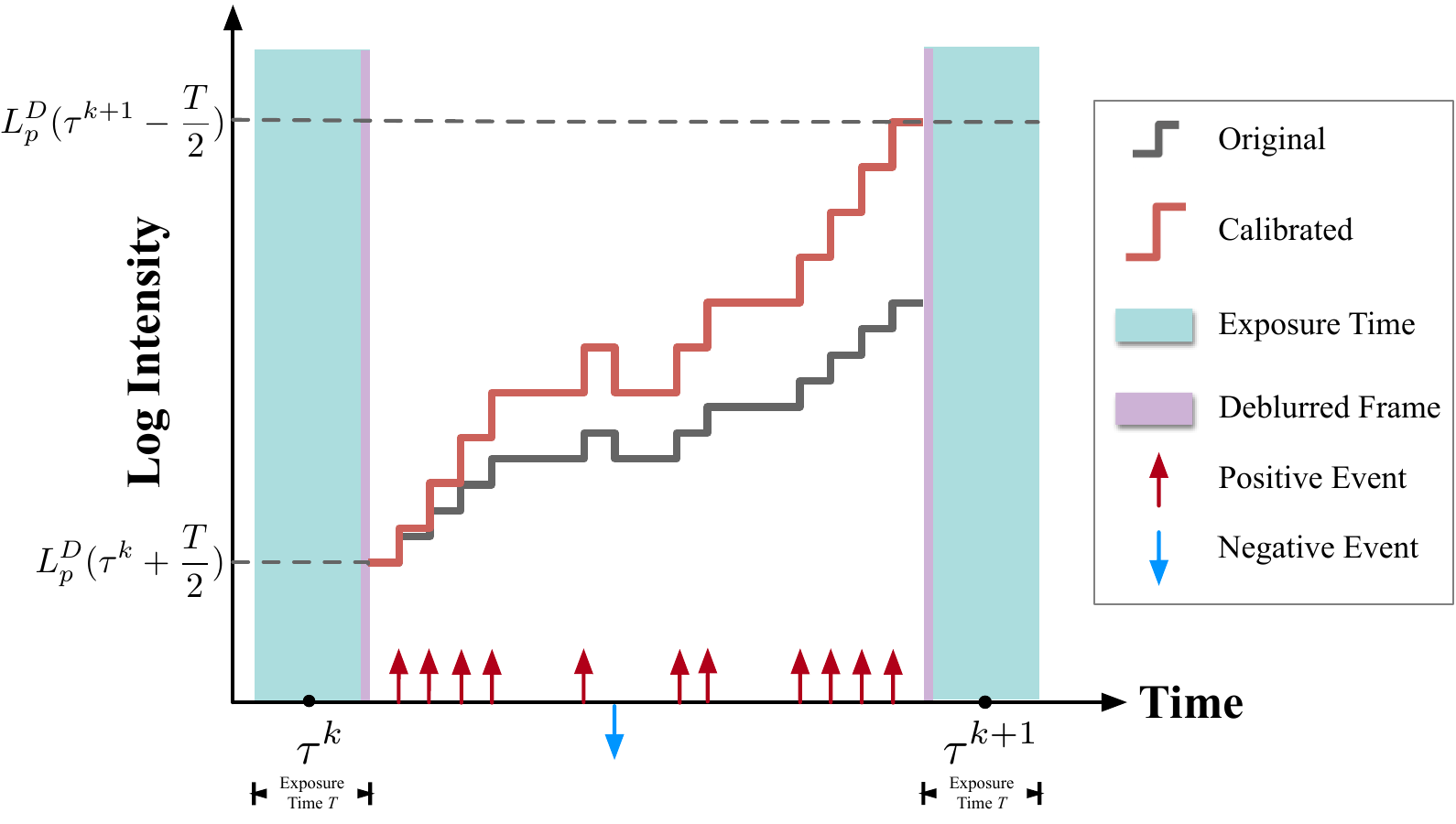}
	\caption{\label{fig: preprocessing}
		Frame augmentation.
		Two deblurred frames at times $\tau^k + \frac{T}{2}$ and $\tau^{k+1} - \frac{T}{2}$ are computed.
		The event stream is used to interpolate between the two deblurred frames to improve temporal resolution.
	} \vspace{-2mm}
\end{figure}

\subsubsection{Deblur}
Due to long exposure time or fast motion, the intensity images $L^F$ may suffer from severe motion blur.
We use the double integral model (EDI) from~\cite{Pan20pami} to sharpen the blurry low-frequency images to obtain two deblurred images at the centre of exposure period $\tau^k$ and $\tau^{k+1}$.
It can be extended to
$L_{\bm{p}}^D(\tau^k + T/2)$ at the end, and
$L_{\bm{p}}^D(\tau^{k+1} - T/2)$ at the beginning, of the exposure of frame at timestamp $\tau^k$ and $\tau^{k+1}$ by event direct integration (Fig.~\ref{fig: preprocessing}).

\subsubsection{Temporal Interpolation}
\label{sec:temporal_interpolation}
The goal of the interpolation module is to increase the temporal resolution of the frame data.
This is important to overcome the ghosting effects (see ablation study in Figure \ref{fig:ablation_cf} (a)-(b)) that are visible in our previous work where the image frames were interpolated using the zero-order-hold assumption \cite{Scheerlinck18accv, Scheerlinck19ral}.

The interpolation $L_{\bm{p}}^A(t)$ within exposure period is computed by event direct integration of the deblurred image $L_{\bm{p}}^{D}(\tau^k)$:

\begin{equation}
L_{\bm{p}}^A(t) =
\left\{
\begin{array}{l}
L_{\bm{p}}^{D}(\tau^k) - \int_{t}^{\tau^{k}}{e_{\bm{p}}}(\gamma) d\gamma, \\
\quad\quad\quad\quad\quad\quad \text{ if } t \in [\tau^k -T/2, \tau^k). \\
L_{\bm{p}}^{D}(\tau^k) + \int_{\tau^{k}}^{t}
{e_{\bm{p}}}(\gamma) d\gamma, \\
\quad\quad\quad\quad\quad\quad \text{ if } t \in [\tau^k, \tau^k + T/2). \\
\end{array}
\right.
\label{eq:interpolation3}
\end{equation}

To estimate intensity at the $i^{th}$ event timestamp at pixel $\bm{p}$ between two frame exposure period from $(\tau^k+T/2)$ to $(\tau^{k+1}-T/2)$, we integrate forward from a deblurred image $L_{\bm{p}}^D (\tau^k + T/2)$ taken from the end of the exposure (Fig.~\ref{fig: preprocessing}).
That is the forward interpolation is
\begin{equation} \label{eq:interpolation1}
L_{\bm{p}}^{A-}(t) = L_{\bm{p}}^D(\tau^{k} + T/2) + \int_{\tau^{k} + T/2}^{t}  {e_{\bm{p}}}(\gamma) d\gamma.
\end{equation}
where $L_{\bm{p}}^{A-}$ denotes the forward interpolated image.
Similarly, we interpolate backwards from the start of exposure $(\tau^{k+1} - T/2)$ to obtain
\begin{equation} \label{eq:interpolation2}
L_{\bm{p}}^{A+}(t) = L_{\bm{p}}^D(\tau^{k+1} - T/2) - \int_{t}^{\tau^{k+1} - T/2} {e_{\bm{p}}}(\gamma) d\gamma.
\end{equation}

Ideally, if there are no missing or biased events and the frame data is not noisy, then the forwards and backwards interpolation results $L_{\bm{p}}^{A-}(t^{i}_{\bm{p}})$ and $L_{\bm{p}}^{A+}(t^{i}_{\bm{p}})$ computed with the true contrast threshold should be equal.
However, noise in either the event stream or in the frame data will cause the two interpolations to differ.
We reconcile these two estimates by per-pixel calibration of the contrast threshold in each interpolation period.
Define the scaling factor of the contrast threshold
\begin{equation} \label{eq:scaling factor}
c^k_{\bm{p}} := \frac{L_{\bm{p}}^D(\tau^{k+1} - T/2) - L_{\bm{p}}^D(\tau^{k} + T/2)}{\int_{\tau^{k}+T/2}^{\tau^{k+1}-T/2} {e_{\bm{p}}}(\gamma) d\gamma}.
\end{equation}
This calibration can be seen as using the shape provided by the event integration between deblurred frames and changing the contrast threshold to vertically stretch or shrink the interpolation to fit the deblurred frame data (Fig.~\ref{fig: preprocessing}).
This is particularly effective at compensating for refractory noise where missing events are temporally correlated to the remaining events.

The inter-frame interpolation $L_{\bm{p}}^A(t)$ is defined as the weighted average of the forward and backward interpolation:
\begin{align*} \label{eq:scaling factor}
L_{\bm{p}}^A(t) &= (1-w) L_{\bm{p}}^{A+}(t)
+ w L_{\bm{p}}^{A-}(t), \\
t &\in [\tau^k+T/2, \tau^{k+1} - T/2),
\end{align*}
where $w$ is the weight between the forward and backward interpolation which is defined as

\begin{equation}
w =  \frac{t-(\tau^k+T/2)}{\tau^{k+1}-\tau^k-T}.
\end{equation}

\begin{remark}
For lower computational cost, the frame augmentation process can be skipped and the filter run on raw log-intensity frames with zero-order-hold (ZOH) analogous to the complementary filter~\cite{Scheerlinck18accv}:
	\begin{align}
	L^{A}_{\bm{p}}(t) := L^{F}_{\bm{p}}(\tau), \quad \tau \leq t < \tau + 1
	\end{align}
The resulting algorithm is the complementary filter with pixel-by-pixel Kalman gain tuning.
We evaluate this simplification in Section \ref{sec:results} and demonstrate that the frame augmentation is important in overcoming ghosting effects.
\end{remark}

\begin{figure*}
	\centering
	\resizebox{0.97\textwidth}{!}{
		\begin{tabular}{
				>{\centering\arraybackslash}m{5cm}
				>{\centering\arraybackslash}m{5cm} >{\centering\arraybackslash}m{5cm}
				>{\centering\arraybackslash}m{5cm}
				>{\centering\arraybackslash}m{5cm}
				>{\centering\arraybackslash}m{5cm}}	
			\includegraphics[width=1\linewidth]{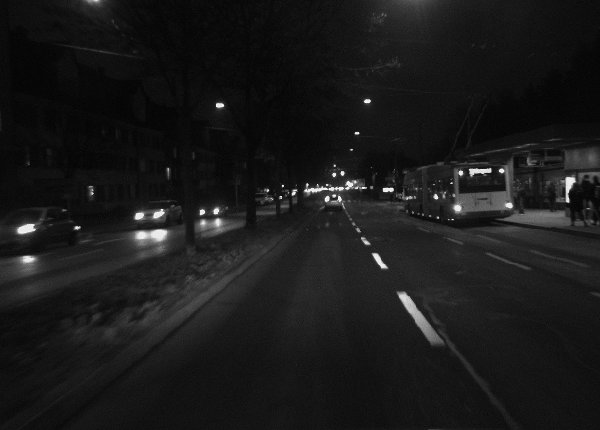}
			&
			\includegraphics[width=1\linewidth]{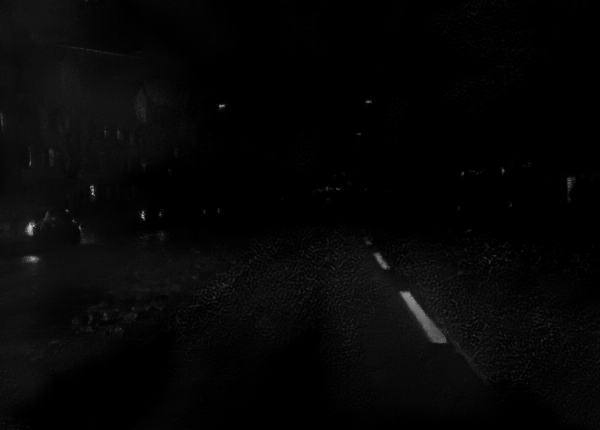}
			&
			\includegraphics[width=1\linewidth]{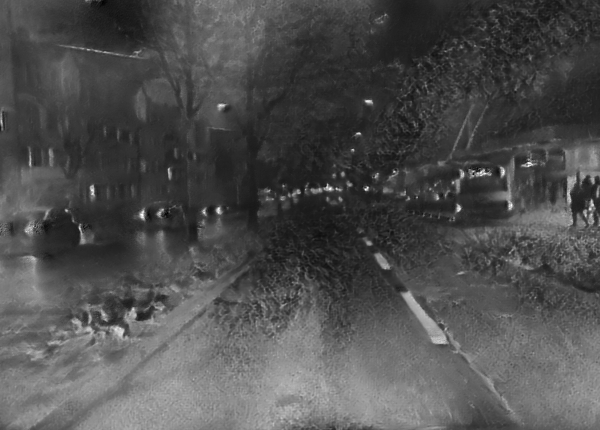}
			&
			\includegraphics[width=1\linewidth]{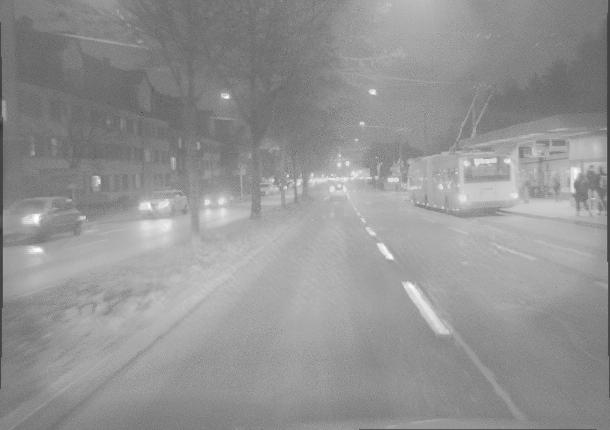}
			&
			\includegraphics[width=1\linewidth]{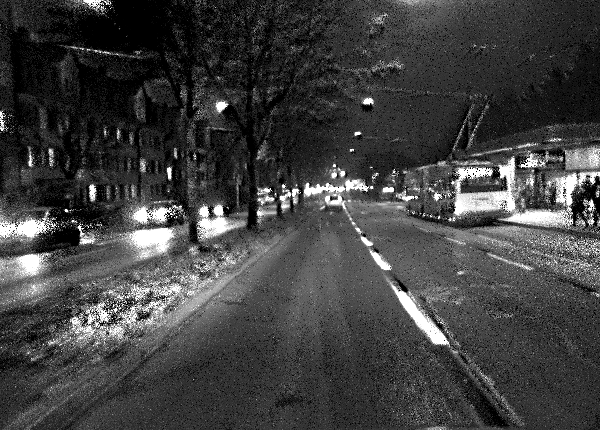}
			&
			\includegraphics[width=1\linewidth]{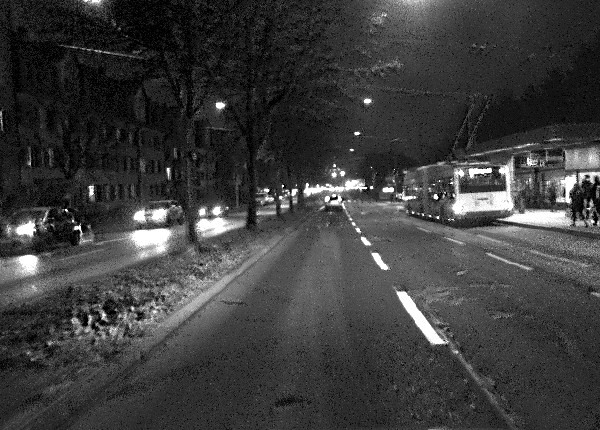}
			\\
			\includegraphics[width=1\linewidth]{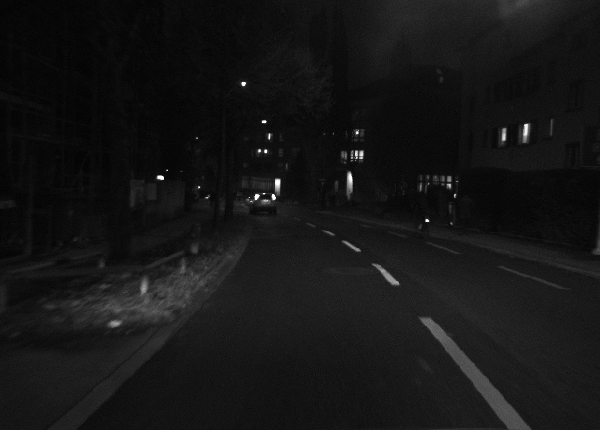}
			&
			\includegraphics[width=1\linewidth]{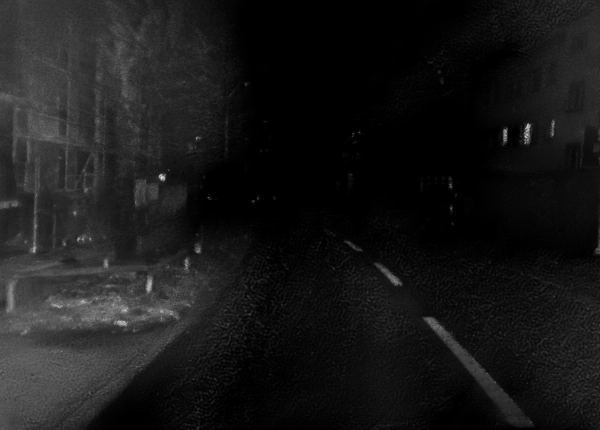}
			&
			\includegraphics[width=1\linewidth]{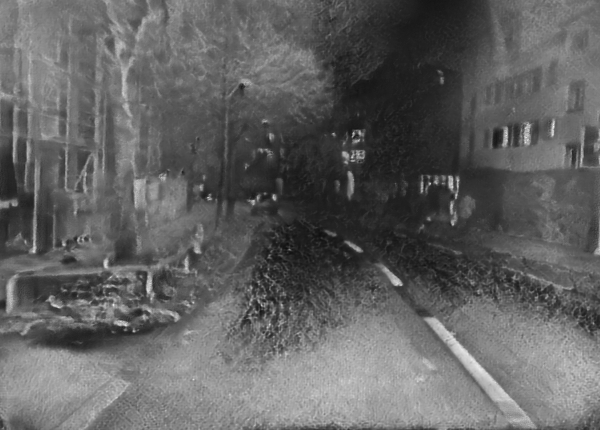}
			&
			\includegraphics[width=1\linewidth]{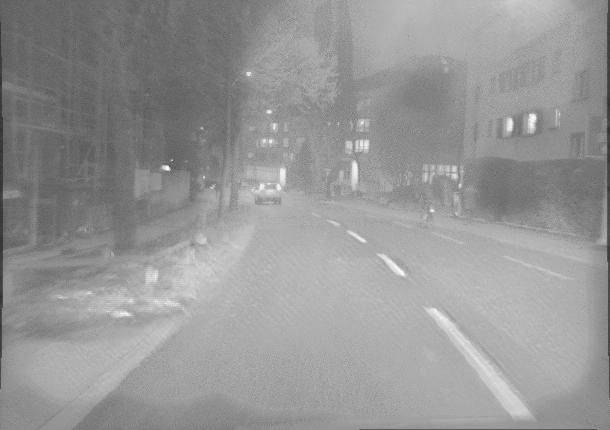}
			&
			\includegraphics[width=1\linewidth]{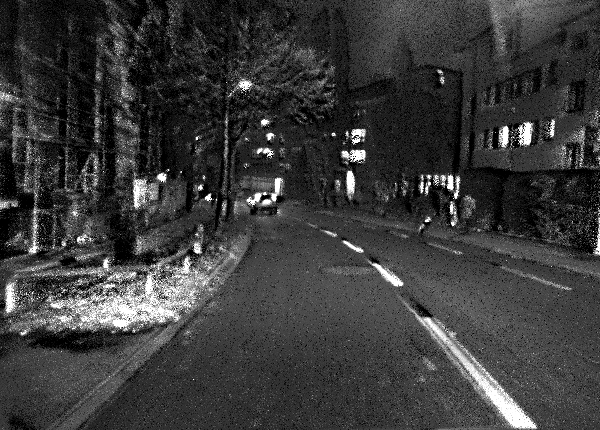}
			&
			\includegraphics[width=1\linewidth]{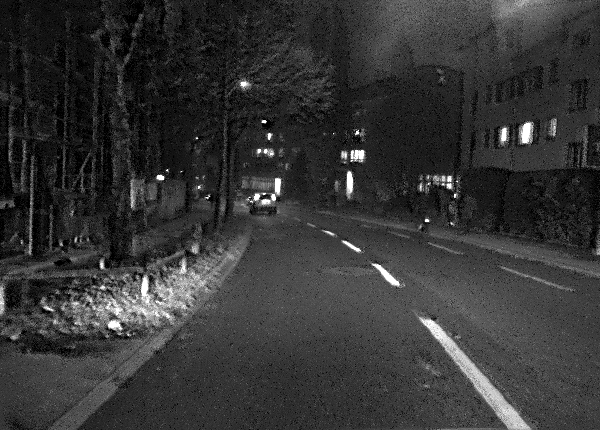}
			\\
			\includegraphics[width=1\linewidth]{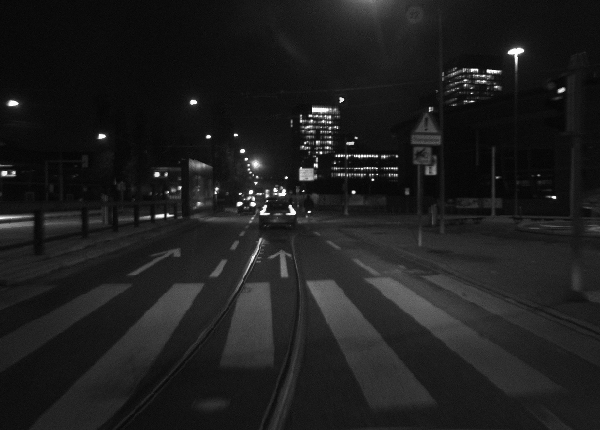}
			&
			\includegraphics[width=1\linewidth]{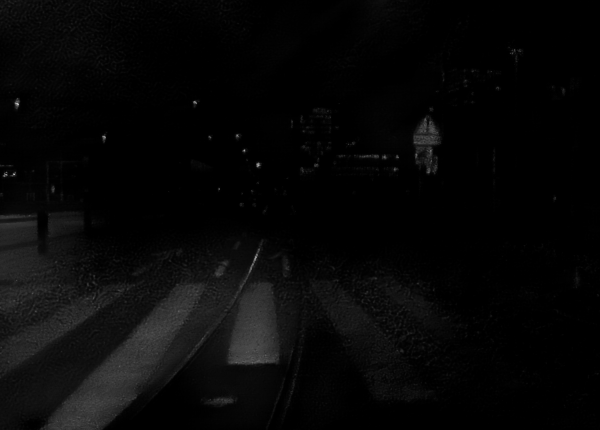}
			&
			\includegraphics[width=1\linewidth]{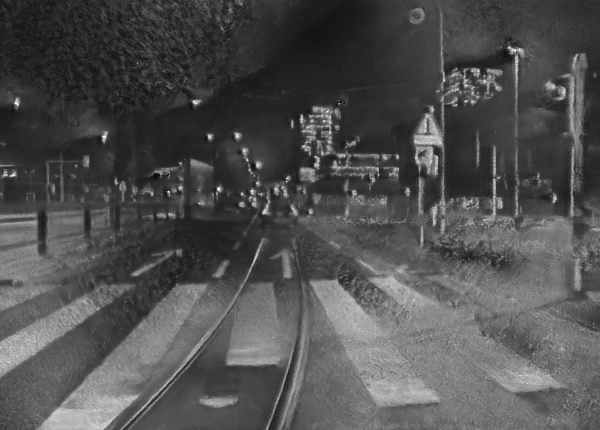}
			&
			\includegraphics[width=1\linewidth]{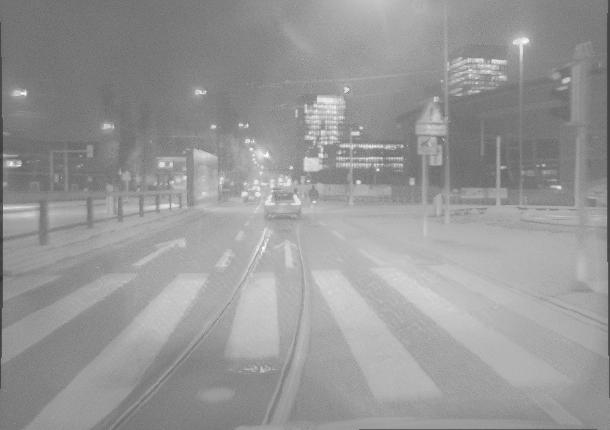}
			&
			\includegraphics[width=1\linewidth]{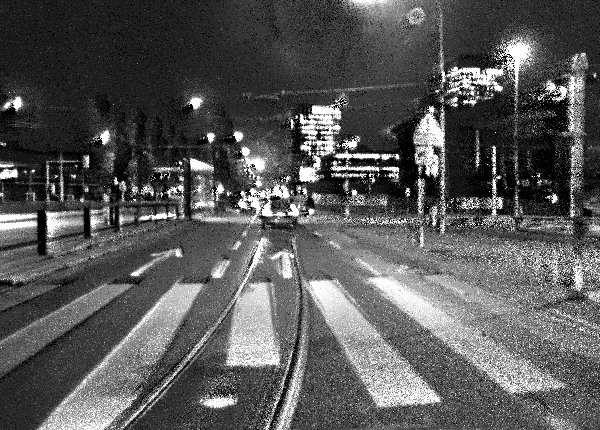}
			&
			\includegraphics[width=1\linewidth]{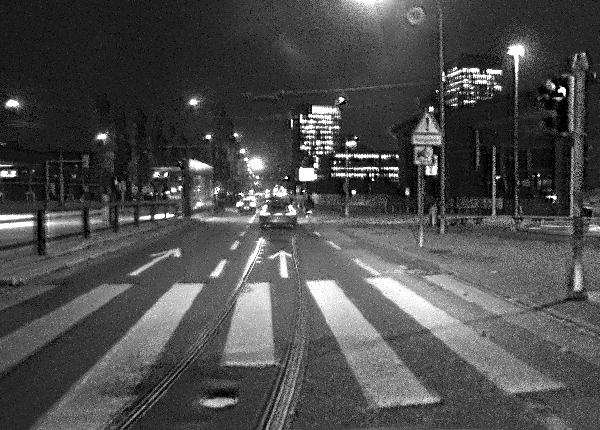}
			\\	
			\includegraphics[width=1\linewidth]{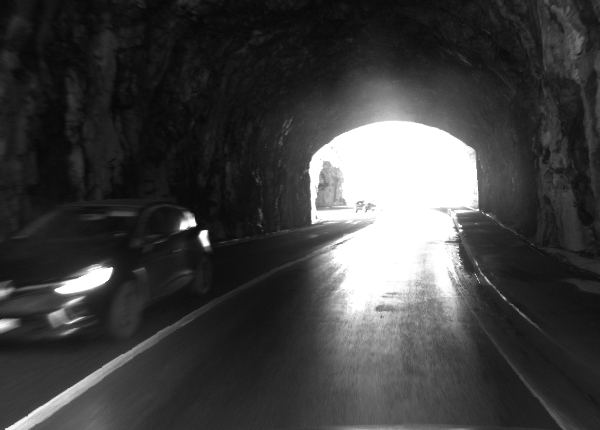}
			&
			\includegraphics[width=1\linewidth]{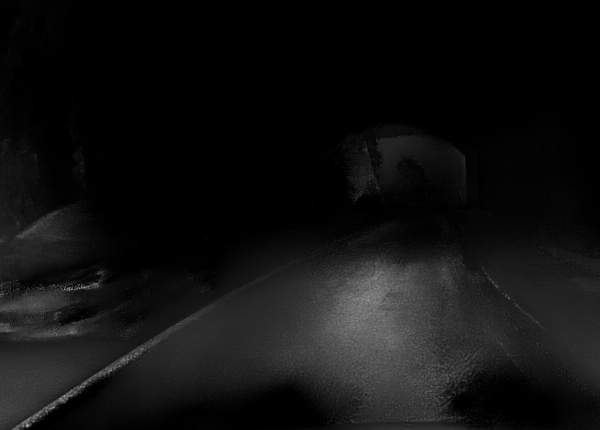}
			&
			\includegraphics[width=1\linewidth]{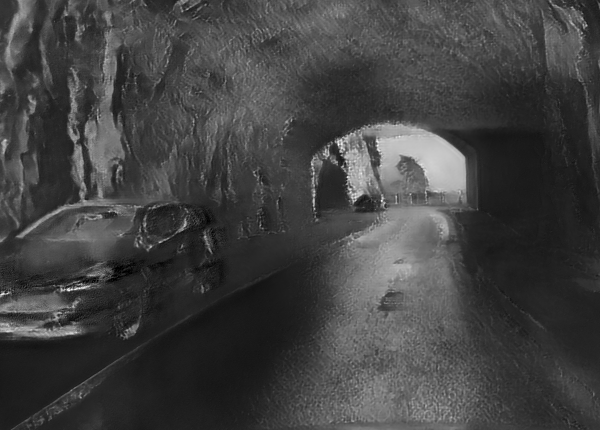}
			&
			\includegraphics[width=1\linewidth]{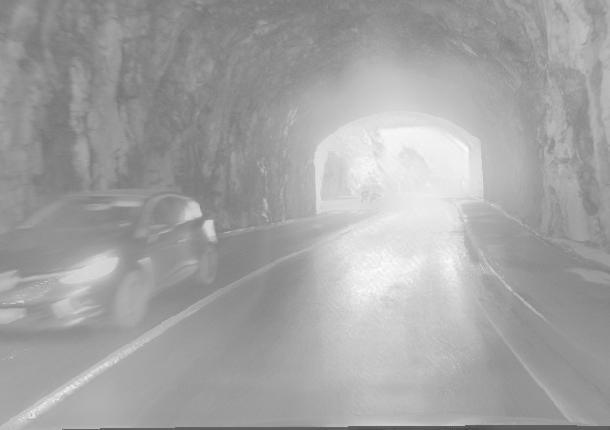}
			&
			\includegraphics[width=1\linewidth]{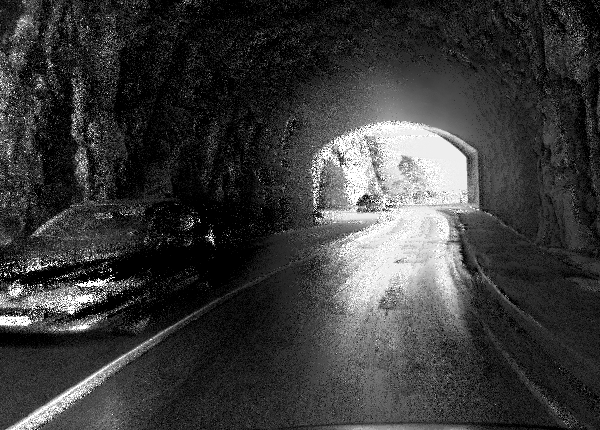}
			&
			\includegraphics[width=1\linewidth]{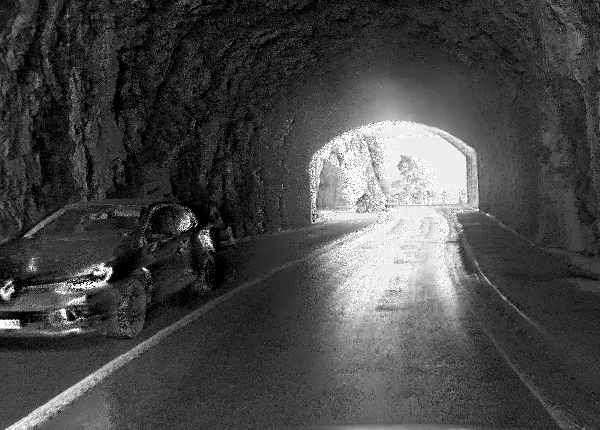}
			\\
			\includegraphics[width=1\linewidth]{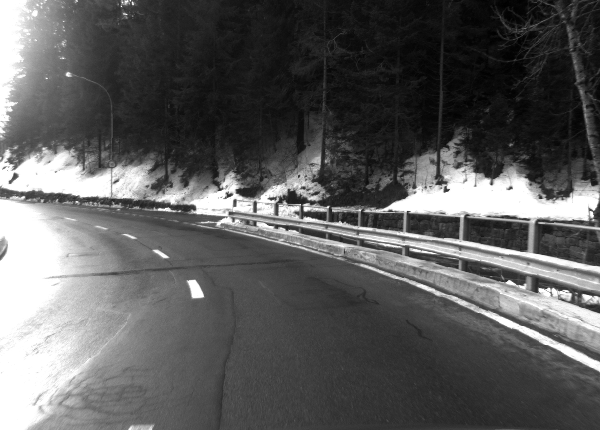}
			&
			\includegraphics[width=1\linewidth]{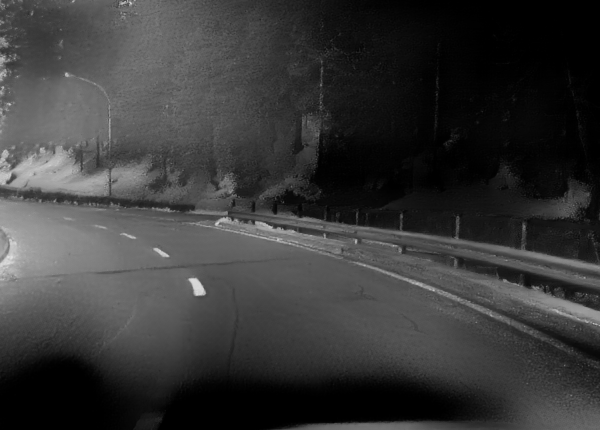}
			&
			\includegraphics[width=1\linewidth]{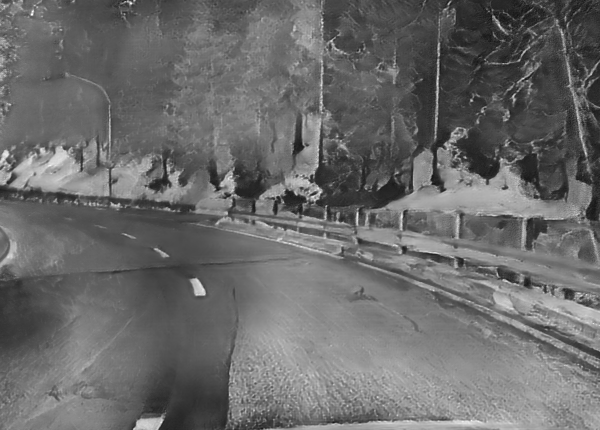}
			&
			\includegraphics[width=1\linewidth]{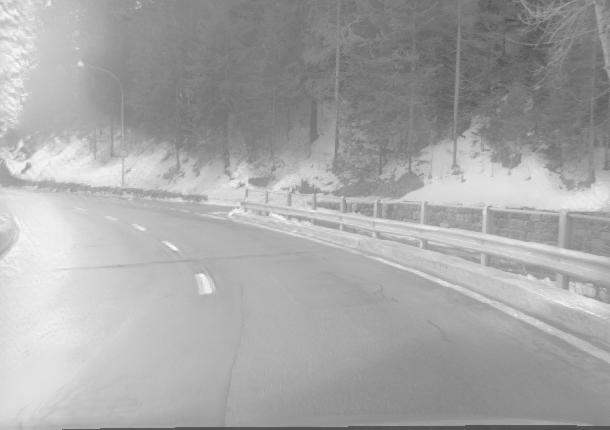}
			&
			\includegraphics[width=1\linewidth]{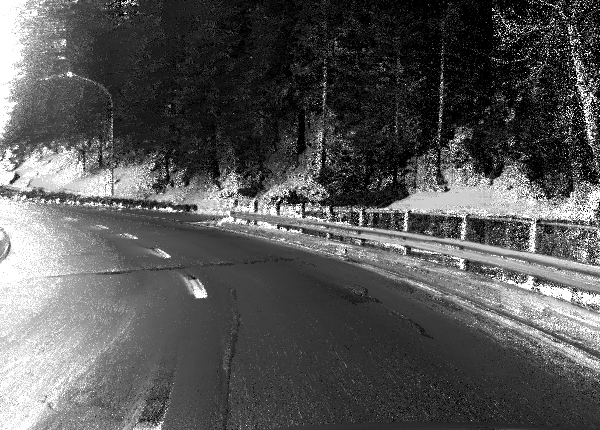}
			&
			\includegraphics[width=1\linewidth]{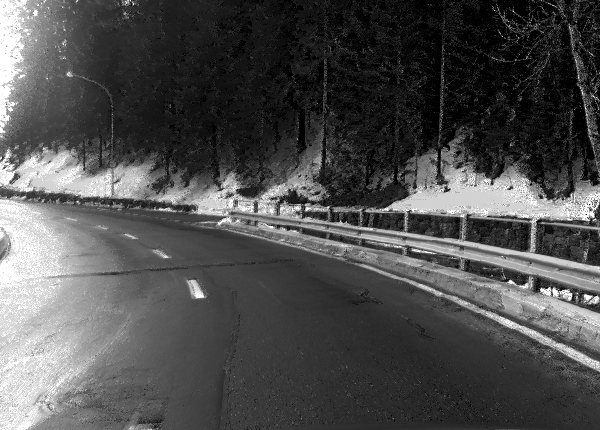}
			\\
			\Large (a) LDR input frame
			&\Large (b) E2VID \cite{Rebecq20pami}
			&\Large (c) ECNN \cite{Stoffregen20eccv}
			&\Large (d) Han \etal~\cite{han2020neuromorphic}
			&\Large (e) CF (ours)
			&\Large (f) \textbf{AKF (ours)}
		\end{tabular}
	}
	\caption{
		Comparison of state-of-the-art event-based video reconstruction methods on the newest open-source event camera driving sequences DSEC~\cite{Gehrig21ral}, with challenging city night scenes and high dynamic range outdoor scenes.
		E2VID~\cite{Rebecq20pami} fails on city night sequences, possibly because the DSEC dataset~\cite{Gehrig21ral} is based on a different type of event camera.
		ECNN~\cite{Stoffregen20eccv} achieves better HDR reconstruction but still sensitive to event noise.
		Han \etal~\cite{han2020neuromorphic} generates HDR images with clear details in both dark and bright scenarios but the images suffer from blur and washed-out artefacts.
		Our CF and AKF are able to compute sharp and clear objects in extreme lighting conditions and clearly outperform other methods.
	}
	\label{fig:dsec}
\end{figure*}

%
\begin{figure*}
	\newcommand{\colwidth}{3.3cm}
	\renewcommand{\tabcolsep}{0.4mm}
	\centering
	\resizebox{1\textwidth}{!}{ 
		\begin{tabular}
			{
				>{\centering\arraybackslash}m{4mm} >{\centering\arraybackslash}m{\colwidth} |
				>{\centering\arraybackslash}m{\colwidth} >{\centering\arraybackslash}m{\colwidth}
				>{\centering\arraybackslash}m{\colwidth}
				>{\centering\arraybackslash}m{\colwidth}
				>{\centering\arraybackslash}m{\colwidth} | >{\centering\arraybackslash}m{\colwidth}}
			\rotatebox{90}{\textbf{HDR} \texttt{Trees}}
			&
			\includegraphics[width=\linewidth]{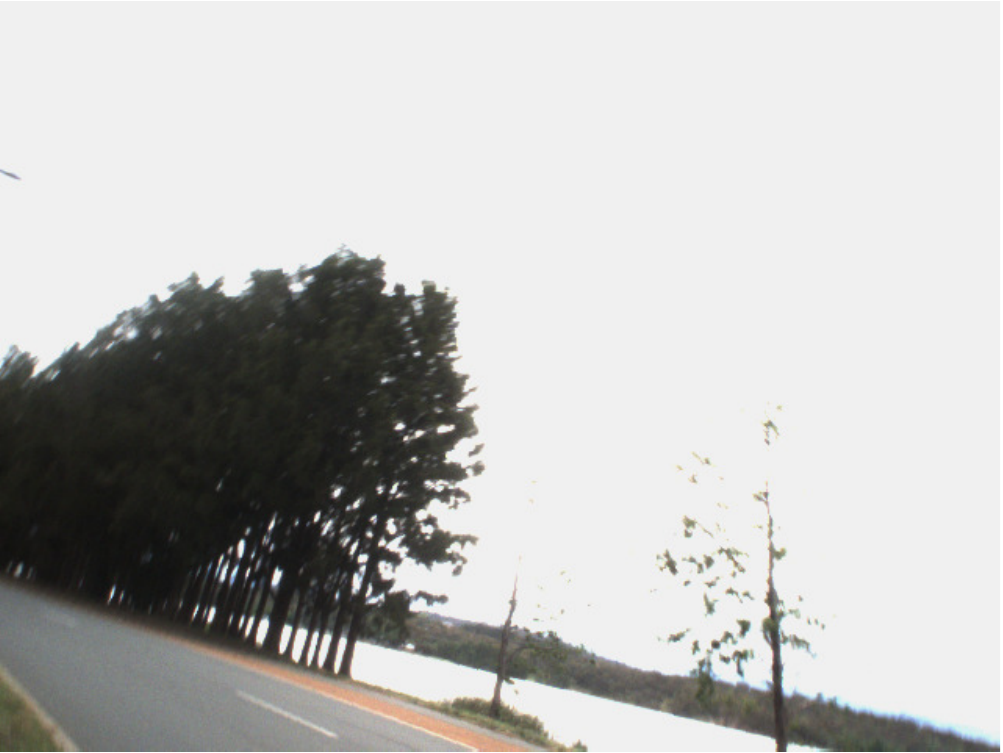}
			&
			\includegraphics[width=\linewidth]{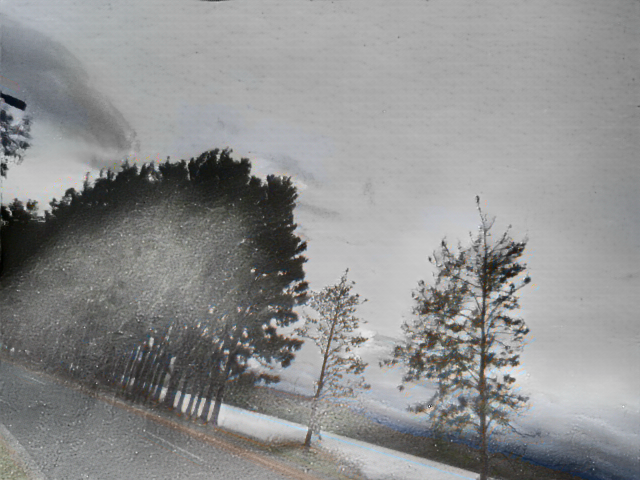}
			&
			\includegraphics[width=\linewidth]{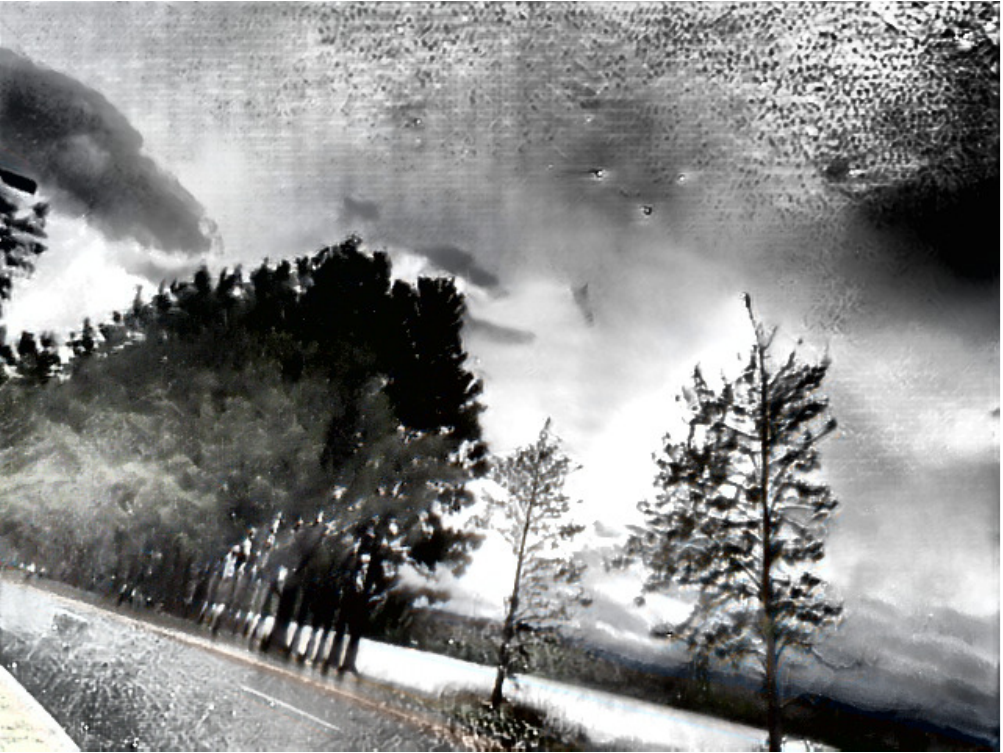}
			&
			\includegraphics[width=\linewidth]{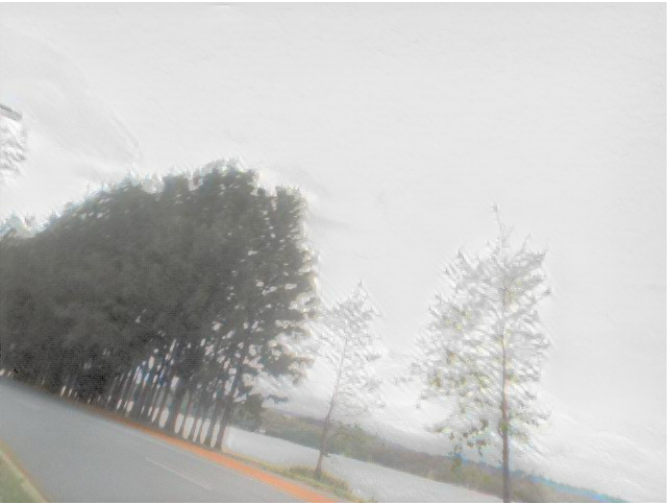}
			&
			\includegraphics[width=\linewidth]{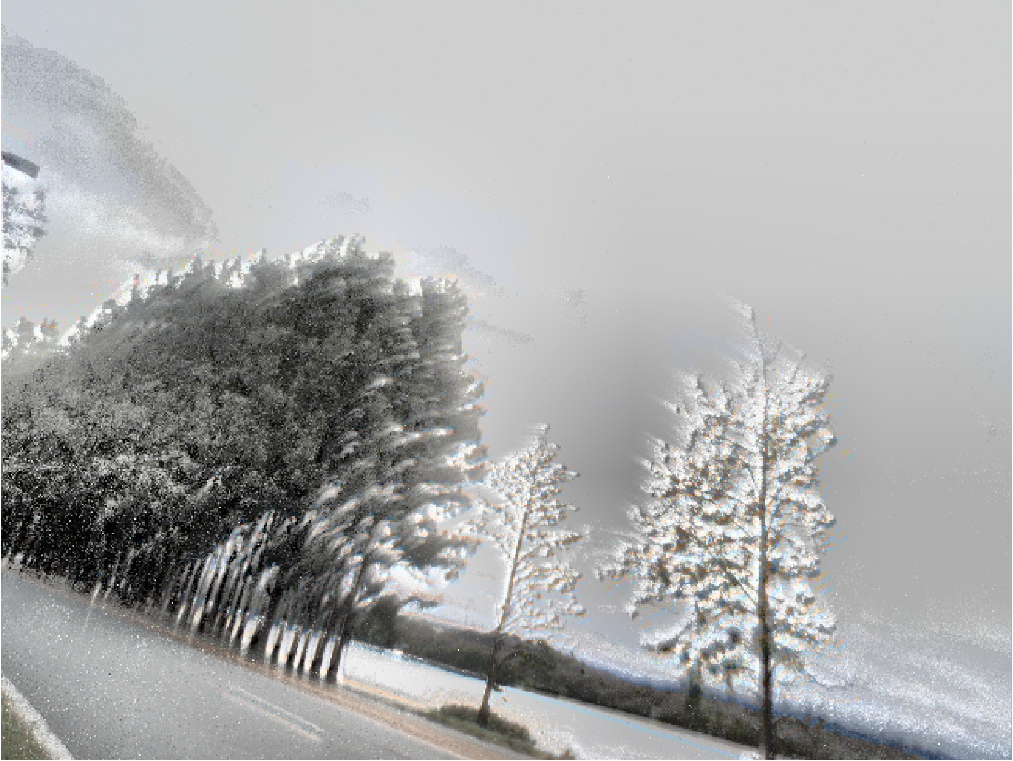}
			&
			\includegraphics[width=\linewidth]{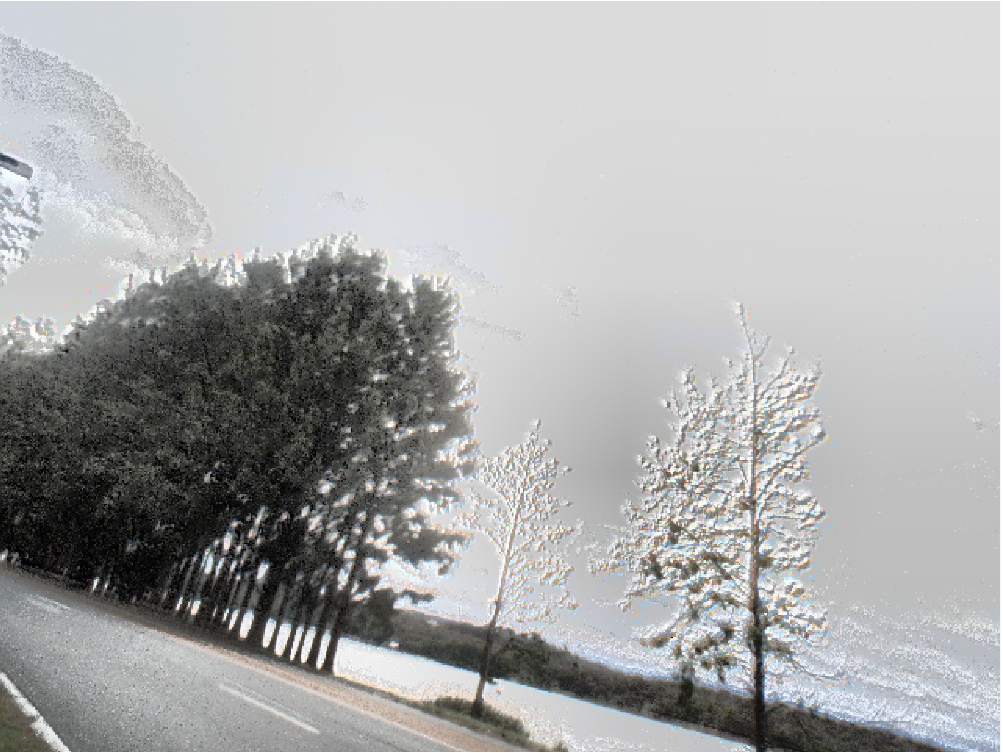}
			&
			\includegraphics[width=\linewidth]{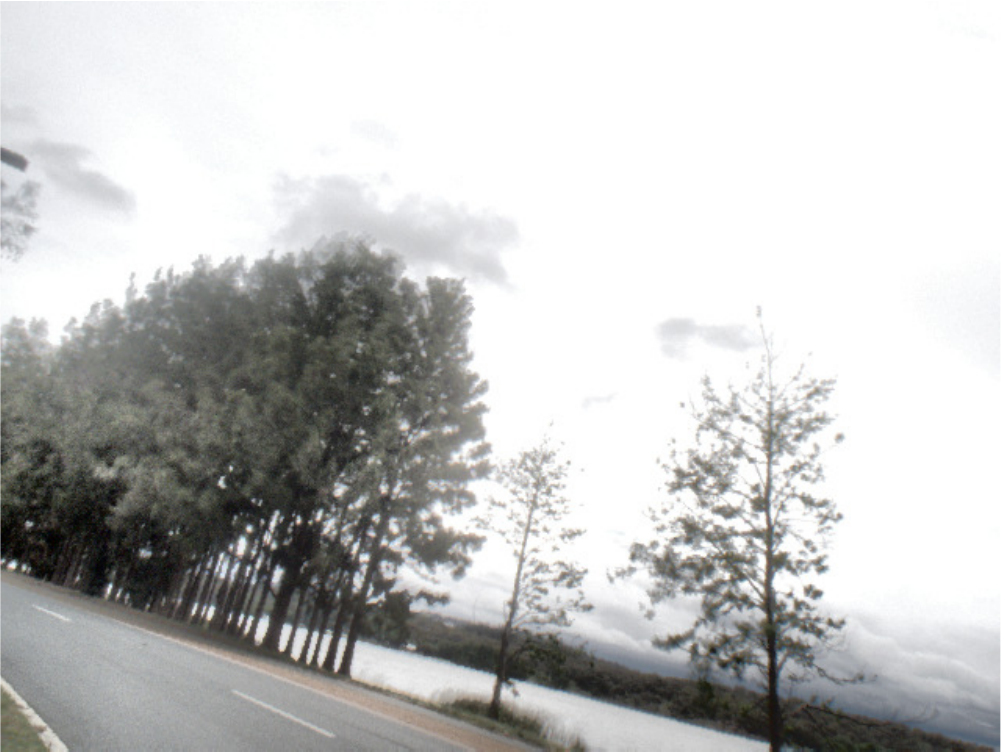}
			\\
			\midrule
			\rotatebox{90}{\textbf{AHDR} \texttt{Mountain}}
			&
			\includegraphics[width=\linewidth]{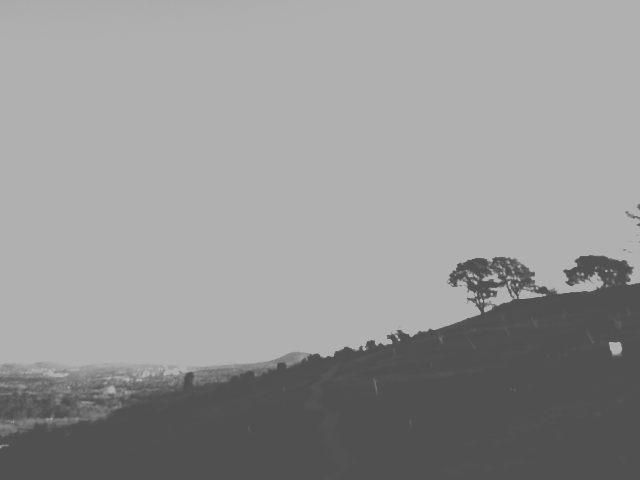}
			&
			\includegraphics[width=\linewidth]{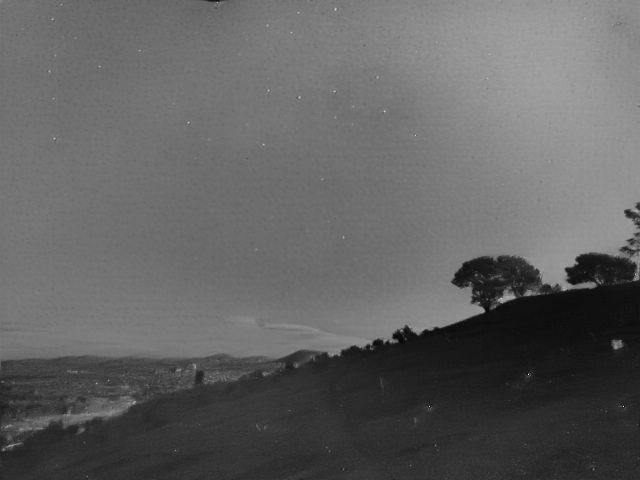}
			&
			\includegraphics[width=\linewidth]{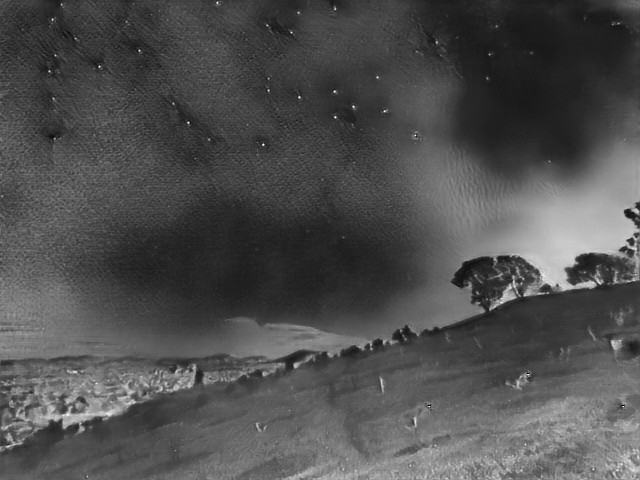}
			&
			\includegraphics[width=\linewidth]{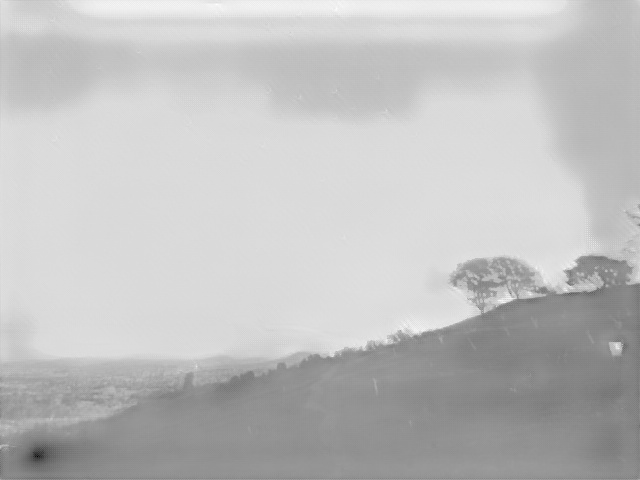}
			&
			\includegraphics[width=\linewidth]{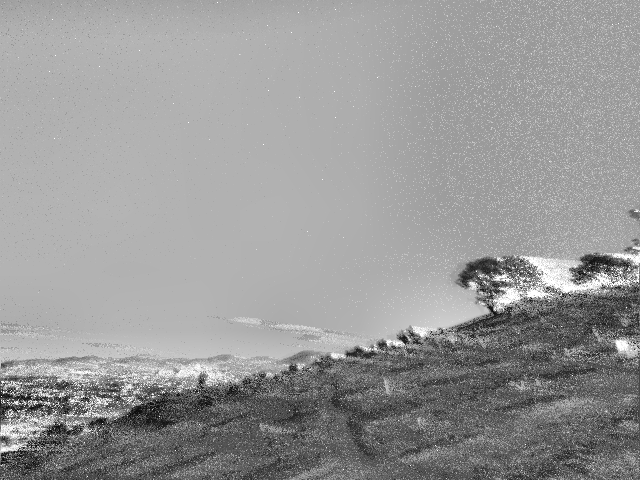}
			&
			\includegraphics[width=\linewidth]{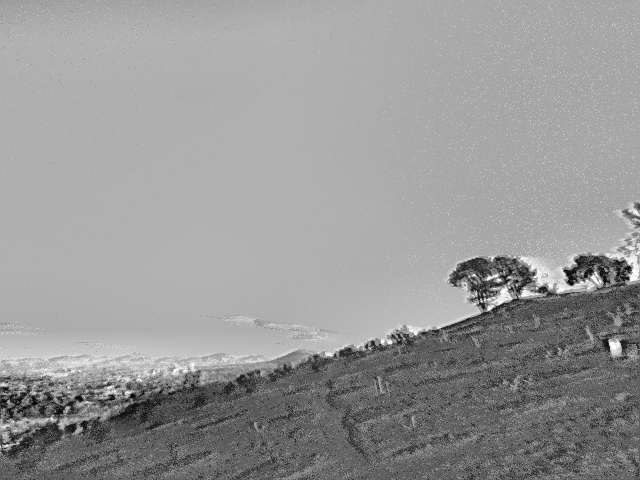}
			&
			\includegraphics[width=\linewidth]{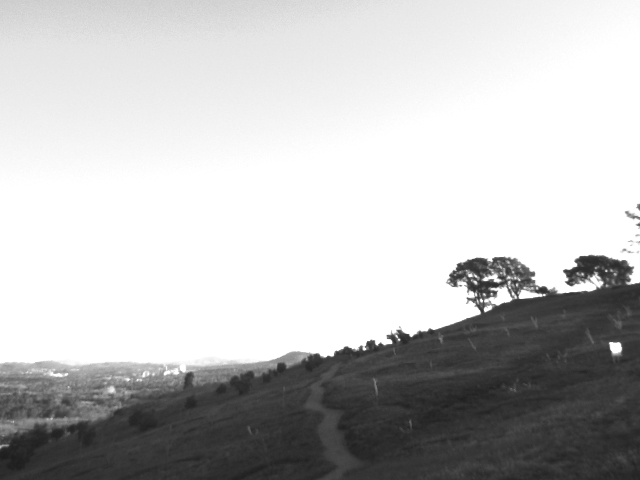}
			\\		
			& (a) LDR input image
			& (b) E2VID \cite{Rebecq20pami}
			& (c) ECNN \cite{Stoffregen20eccv}
			& (d) Han \etal~\cite{han2020neuromorphic}
			& (e) CF (ours)
			& (f) \textbf{AKF (ours)}
			& (g) Reference Image
		\end{tabular}
	}
	\caption{
		Typical results from the proposed HDR and AHDR dataset.
		Our HDR dataset includes referenced HDR images generated by fusing several images of various exposures.
		Our AHDR dataset is simulated by saturating the values of well-exposed real images, taking out most of the details.
		The original images are used as HDR references.
		E2VID~\cite{Rebecq20pami} and ECNN~\cite{Stoffregen20eccv} use event data only.
		They perform poorly on the dark trees in the HDR dataset and the road/sky in the AHDR dataset.
		Han \etal~\cite{han2020neuromorphic}  successfully reconstructs \texttt{Tree} but fails to reconstruct the road in \texttt{Mountain}.
		The output image also has unrealistic artefacts appearing near the image edges.
		Our CF fuses event data and low dynamic range frames properly but still suffers from motion blur (on the left-hand trees) and shadows on moving object edges.
		Our AKF correctly computes the underexposed and overexposed trees sharply in the HDR dataset and reconstructs the mountain road clearly in the artificially saturated regions.
	}
	\label{fig:hdr}
\end{figure*}


\subsection{Event-based Spatial Convolutions}
Spatial event convolution requires consideration of events in a local neighbourhood of a pixel as well as the events that occur at that pixel.
Consider a global index for all events across all pixels
	\begin{align}
	\text{event}^j = (t^j,\bm{p}^j,\sigma^j), j \in 1,2,3...
	\end{align}
Analogous to \eqref{eq:continuous_event}, the global continuous event field is represented as
	\begin{align}
	E({\bm{p}}, t) =\sum_{j=1}^\infty
	(c \sigma^j + \eta^j) \delta(t-t^j){\delta_{\bm{p}\bm{p}^j}},
	\label{eq:continuous_event_global}
	\end{align}
where $\delta(t)$ is a Dirac delta function and $\delta_{\bm{p}\bm{p}^j}$ is a Kronecker delta function; ${\delta_{\bm{p}\bm{p}^j}} = 1$ if $\bm{p} = \bm{p}^j$ and zero otherwise \cite{hassani2008mathematical}.
The integral of events is	
	\begin{align} \label{eq:L_}
	\int_{0}^t E(\bm{p}, \gamma) d \gamma \approx L(\bm{p}, t) - L(\bm{p}, 0),
	\end{align}
where $L(\bm{p}, t)$ is the log intensity seen by the camera with initial condition $L(\bm{p}, 0)$.
Let $K$ denote a linear spatial kernel with finite support.
Consider the convolution of $K$ with $L(\bm{p}, t)$.
Define
\begin{align} \label{eq:L^K_}
L^K(\bm{p}, t) := K * L(\bm{p}, t).
\end{align}
Note that the convolution of kernel $K$ is only applied spatially on the first dimension $\bm{p}$.
Using \eqref{eq:L_} and \eqref{eq:L^K_}, the convolution is derived as
\begin{align}
L^K(\bm{p}, t) & \approx K * L(\bm{p}, 0) + \int_{0}^t K * E(\bm{p},\gamma) d \gamma, \notag \\
	& \approx K * L(\bm{p}, 0) +   \int_{0}^t \sum_{j=1}^\infty c \, \sigma^j \, \delta(\gamma - t^j) \, {K * \delta_{\bm{p}\bm{p}^j}} d \gamma, \notag  \\
	& \approx K * L (\bm{p},0) + \int_{0}^t E^K(\bm{p},\gamma) d \gamma, \label{eq:KstarL}
	\end{align}
	where
\begin{align}
E^K(\bm{p}, t) := \sum_{j=1}^\infty c \, \sigma^j \, \delta(t - t^j) \, {K * \delta_{\bm{p}\bm{p}^j}}. \label{eq:_i}
\end{align}
Here $K * \delta_{\bm{p}\bm{p}^j}$ is the spatial convolution of the finite support kernel $K$ with an image $\delta_{\bm{p}\bm{p}^j}$ with a single non-zero pixel $\bm{p} = \bm{p}^j$ with unity value.
The result of such a convolution is an image with pixel values of zero everywhere except for a patch centred on $\bm{p}^j$ (the same size as $K$) with values drawn from the coefficients of $K$.
The convolved event field $E^K(\bm{p}, t)$ can be thought of as a finite (localised) collection of spatially separate events all centering at pixels $\bm{p}^j$, occurring at the same timestamps $t^j$.

\subsection{Continuous-time Filter for Convolved Events} \label{sub:ct-filter}
To compute the spatial convolution of an image by fusing events and frames we follow the steps outlined above.
The event stream is replaced by the convolved event stream~\eqref{eq:_i}, and the image frame is replaced by the convolved version of the log intensity frame~\eqref{eq:KstarL} as shown in Figure~\ref{fig: pipeline}.

Multiple different filters can be run in parallel.
For example, if gradient estimation is required, then two filter states ($\hat{G}_x, \hat{G}_y$) can be run in parallel for the $x$ and $y$ components using an appropriate directional kernels (Sobel, central difference, etc).
The convolved event field increases the number of events that must be processed at any given pixel and the convolution of the input images also adds to the computational complexity of the algorithm.
The algorithm, however, is ideally suited to an embedded implementation and has the potential to be highly effective in the future where more and more processing power is available at the pixel level.

\section{Experiments} \label{sec:results}

\noindent\textbf{Overview:}	
This section aims to evaluate the performance of our proposed methods.
We evaluate our video reconstruction method CF and AKF against state-of-the-art methods and provide a computational complexity analysis and an ablation study of the CF, AKF and the frame augmentation step that illustrate the efficacy and cost of each step.
In addition, we discuss the limitations of our video reconstruction methods and demonstrate the performance of event convolutions.

\noindent\textbf{Comparisons:}
There are two state-of-the-art event-based HDR image reconstruction methods for which open-source code is available:
E2VID~\cite{Rebecq20pami} and ECNN~\cite{Stoffregen20eccv}.
Both these methods are neural networks that use a window of spatio-temporal events, between two video frame time stamps, to generate an input tensor that is processed to generate a single HDR image in the video sequence.
The two neural networks are trained on combined simulated and DAVIS event camera dataset.
We also compare with a recent HDR image reconstruction method, Han \etal~\cite{han2020neuromorphic}, which use both events and frames as input.
In addition, the EDI algorithm~\cite{Pan20pami} combines events and frames to provide high quality image reconstruction within each exposure period that is then extended to inter-frame interpolation using direct integration for video reconstruction.
This algorithm deblurs and augments the LDR input data and the comparison is relevant particularly for inter-frame asynchronous reconstruction.
We do not compare with TimeLens~\cite{tulyakov2021time,Tulyakov22cvpr} since these algorithms treat the LDR input data as key frames and do not improve HDR or undertake image deblurring.

To benchmark the HDR performance of our algorithms, we compare our proposed complementary filter (CF) and Asynchronous Kalman filter (AKF) with the E2VID~\cite{Rebecq20pami}, ECNN~\cite{Stoffregen20eccv}, and Han \etal~\cite{han2020neuromorphic} algorithms on sequences drawn from three stereo hybrid \eventframe datasets (frame and event data come from separate sensors):
the newest open-source hybrid \eventframe camera driving dataset DSEC~\cite{Gehrig21ral} and two targeted HDR datasets (HDR and AHDR) that we have collected on a stereo hybrid event-frame system (discussed in the supplementary material).

\begin{figure*}
	\centering
	\resizebox{0.98\textwidth}{!}{
		\begin{tabular}{
				>{\centering\arraybackslash}m{1mm}
				>{\centering\arraybackslash}m{6cm}
				>{\centering\arraybackslash}m{6cm} >{\centering\arraybackslash}m{6cm}
				>{\centering\arraybackslash}m{6cm}
				>{\centering\arraybackslash}m{6cm}
				>{\centering\arraybackslash}m{6cm}}
			\rotatebox{90}{\texttt{\Large Night drive}}
			&
			\includegraphics[width=1\linewidth]{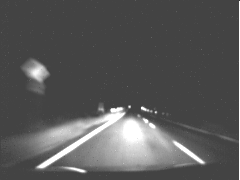}
			&
			\includegraphics[width=1\linewidth]{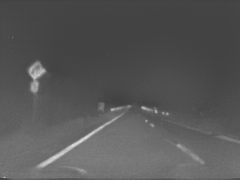}
			&
			\includegraphics[width=1\linewidth]{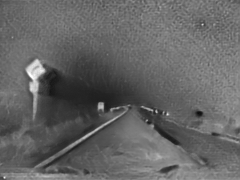}
			&
			\includegraphics[width=1\linewidth]{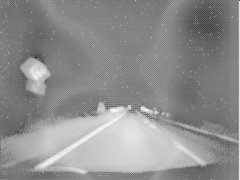}
			&
			\includegraphics[width=1\linewidth]{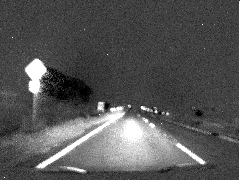}
			&
			\includegraphics[width=1\linewidth]{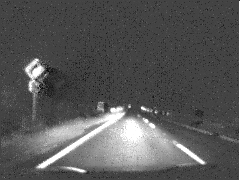}\\
			\\
			\rotatebox{90}{\Large \texttt{Outdoor running}}
			&	
			\includegraphics[width=1\linewidth]{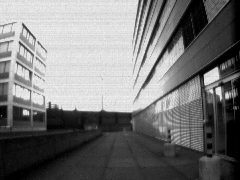}
			&
			\includegraphics[width=1\linewidth]{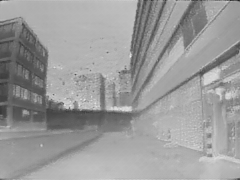}
			&
			\includegraphics[width=1\linewidth]{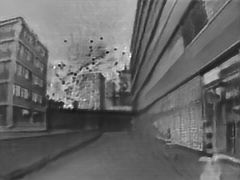}
			&
			\includegraphics[width=1\linewidth]{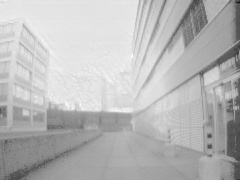}
			&
			\includegraphics[width=1\linewidth]{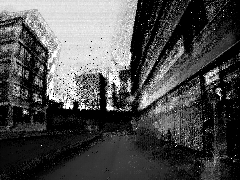}
			&
			\includegraphics[width=1\linewidth]{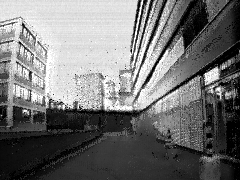}
			\\
			\rotatebox{90}{\Large \texttt{Box 6dof}}
			&
			\includegraphics[width=1\linewidth]{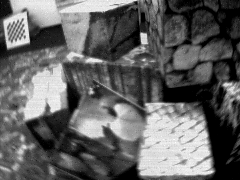}
			&
			\includegraphics[width=1\linewidth]{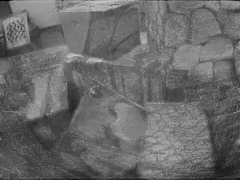}
			&
			\includegraphics[width=1\linewidth]{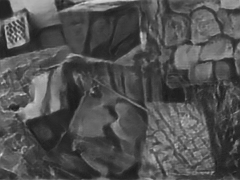}
			&
			\includegraphics[width=1\linewidth]{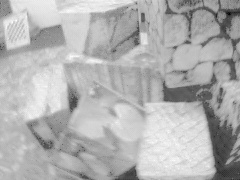}
			&
			\includegraphics[width=1\linewidth]{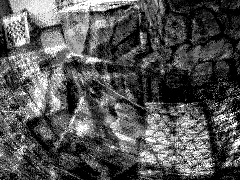}
			&
			\includegraphics[width=1\linewidth]{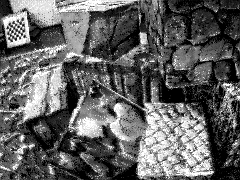}		
			\\
			&\Large (a) DAVIS frame
			&\Large (b) E2VID \cite{Rebecq20pami}
			&\Large (c) ECNN \cite{Stoffregen20eccv}
			&\Large (d) Han \etal~\cite{han2020neuromorphic}
			&\Large (e) CF (ours)
			&\Large (f) \textbf{AKF (ours)}
		\end{tabular}
	}
	\caption{Comparison of state-of-the-art event-based video reconstruction methods on sequences with challenging lighting conditions and fast motions, drawn from the open-source datasets ACD~\cite{Scheerlinck18accv}, CED~\cite{scheerlinck2019ced} and IJRR~\cite{Mueggler17ijrr}.
	E2VID \cite{Rebecq20pami} and ECNN \cite{Stoffregen20eccv} are able to reconstruct the blurry right turn sign in the high-speed, low-light \texttt{Night drive} dataset and the overexposed regions in \texttt{Outdoor running} and \texttt{Box 6dof} dataset.
	But without frame information, E2VID \cite{Rebecq20pami} only provides washed-out reconstructions in all three sequences and ECNN~\cite{Stoffregen20eccv} introduces artefacts at static background.
	ECNN~\cite{Stoffregen20eccv} is also highly sensitive to camera noise in \texttt{Outdoor running}.
	Han \etal~\cite{han2020neuromorphic} uses both frame and event data but the reconstructions are blurry and the brightness level is inaccurate.
	Our CF fails to capture details under extreme lighting conditions and suffers from a `shadowing effect' (white or black shadows trailing behind dark or bright moving objects).
	Our AKF overcomes the limitation of CF and outperforms the other methods in all challenging scenarios.
	Additional image and video comparisons are provided in the supplementary material.
	}
	\label{fig:davis}
\end{figure*}

\begin{figure}
	\centering
	\resizebox{0.5\textwidth}{!}{ 
		\begin{tabular}{ c c }
			\includegraphics[width=0.49\linewidth]{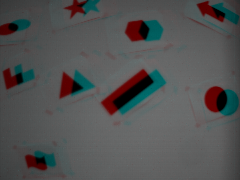} &
			\includegraphics[width=0.49\linewidth]{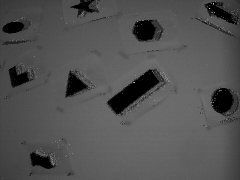}
			\\
			(a) Two Consecutive Raw Images & (b) EDI~\cite{Pan20pami}
			\\
			\includegraphics[width=0.49\linewidth]{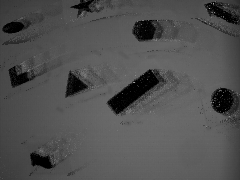} &
			\includegraphics[width=0.49\linewidth]{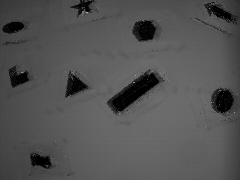}
			\\
			(c) Our CF  & (d) \textbf{Our Full AKF Pipeline}
		\end{tabular}
	}
	\caption{
		An example comparison between three asynchronous event-based video reconstruction methods, EDI~\cite{Pan20pami} and our CF and AKF.
		Image (a) shows two consecutive raw images from an event camera, red for the current frame and blue for the previous frame (shapes moving from right to left).
		Image (b)-(d) demonstrate the inter-frame reconstruction.
		The event direct integration of EDI~\cite{Pan20pami} in (b) fails to update some boundary pixels of the previous frame, leading to boundary shadows.
		CF in (c) simply interpolates between frames by holding intensity values of the previous frame until the next frame (zero-order-hold), which fails on fast motion scenarios.
		AKF reduces the gap between high temporal resolution events with low temporal resolution frames by a novel per-pixel contrast threshold scaling and temporal interpolation method. Therefore, AKF in (d) achieves better inter-frame reconstruction performance.
	}
	\label{fig:edi comp}
	\vspace{-2mm}
\end{figure}

\begin{figure*}[]
	\centering
	\resizebox{0.77\linewidth}{!}{ 
		\begin{tabular}{
				>{\centering\arraybackslash}m{4mm}
				>{\centering\arraybackslash}m{6cm}
				>{\centering\arraybackslash}m{6cm} >{\centering\arraybackslash}m{6cm}}
			\rotatebox{90}{\large  \texttt{Shadow}}
			&
			\includegraphics[width=1\linewidth]{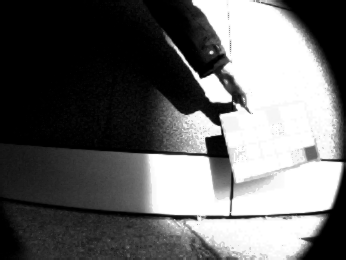} &
			\includegraphics[width=1\linewidth]{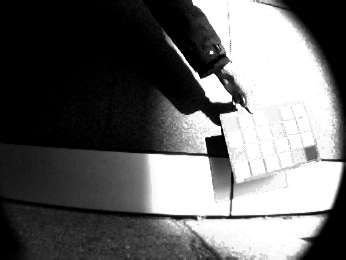} &
			\includegraphics[width=1\linewidth]{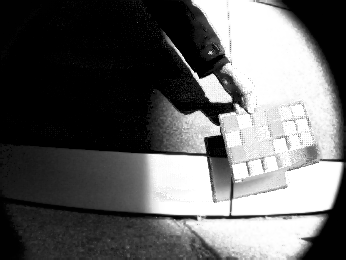}
			\\
			&  (a). LDR frame & (b). EDI~\cite{Pan20pami} in Frame Augmentation & (c). Full AKF Pipeline
			\\
			\rotatebox{
				90}{\large \texttt{Interlaken\_01a}}
			&
			\includegraphics[width=1\linewidth]{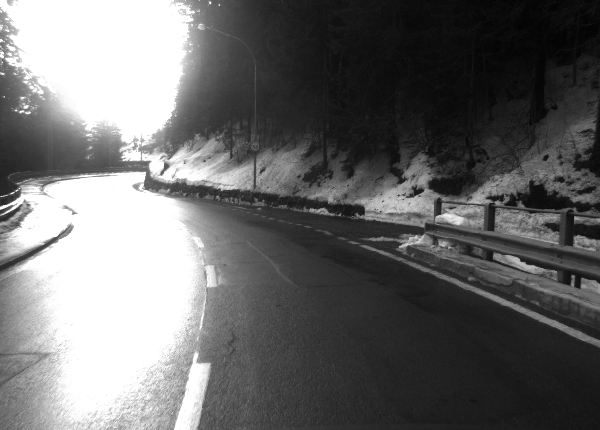} &
			\includegraphics[width=1\linewidth]{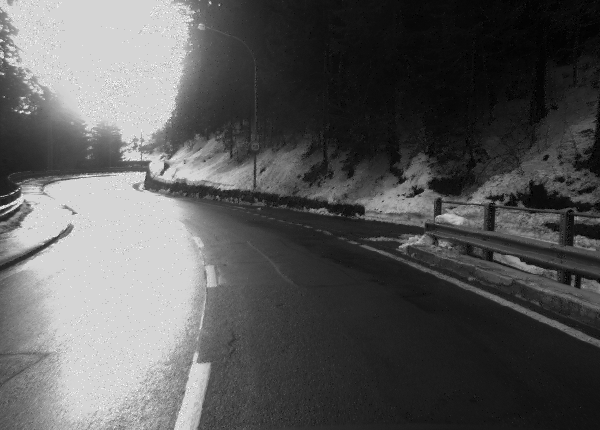} &
			\includegraphics[width=1\linewidth]{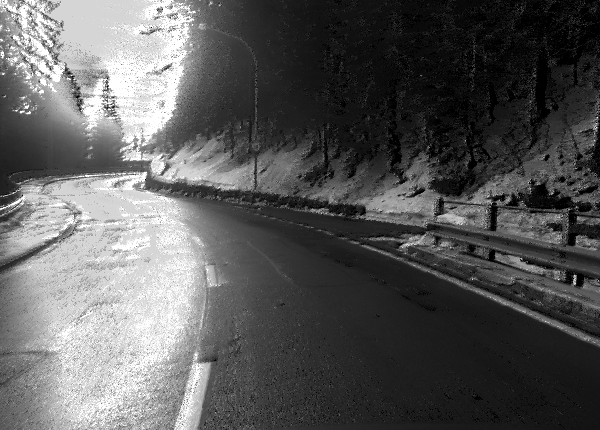}
			\\
			& (d). LDR frame & (e). EDI~\cite{Pan20pami} in Frame Augmentation & (f). Full AKF Pipeline
			\\
		\end{tabular}
	}
	\caption{\label{fig:ablation_edi}
		An example of image reconstruction between EDI~\cite{Pan20pami} in our frame augmentation and our full AKF pipeline.
		In figure (b), the deblurring method EDI~\cite{Pan20pami} computes the grid in the overexposed region using events that happen in the frame exposure time but fails to recover the intensity from white (overexposed) to grey.
		For dataset \texttt{Interlaken\_01a} in figure (e), EDI~\cite{Pan20pami} is not able to reconstruct texture in the highly reflective road region and tree boundaries in the bright sky. It is because the frame camera used in this dataset is set to auto exposure mode, and the extreme bright lighting conditions make the exposure time very short.
		There is insufficient events within the frame exposure time to recover the scene.
		However, our AKF pipeline is able to provide competitive result in the challenging lighting conditions.
		Figure (c) and (f) contains much more grid and road texture in the strong sun reflection area, which uses the state `memory' that associates with the previous frame/event data.
	}
\end{figure*}

\begin{figure*}[]
	\centering
	\resizebox{0.8\linewidth}{!}{ 
		\begin{tabular}{
				>{\centering\arraybackslash}m{6cm}
				>{\centering\arraybackslash}m{6cm} |
				>{\centering\arraybackslash}m{6cm} >{\centering\arraybackslash}m{6cm}}
			\includegraphics[width=1\linewidth]{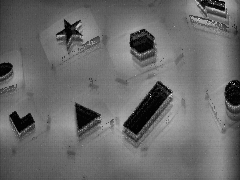} &
			\includegraphics[width=1\linewidth]{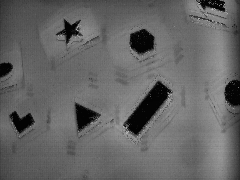} &
			\includegraphics[width=1\linewidth]{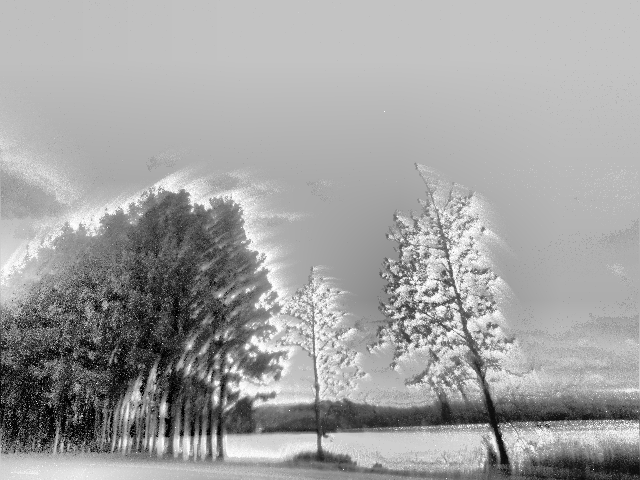} &
			\includegraphics[width=1\linewidth]{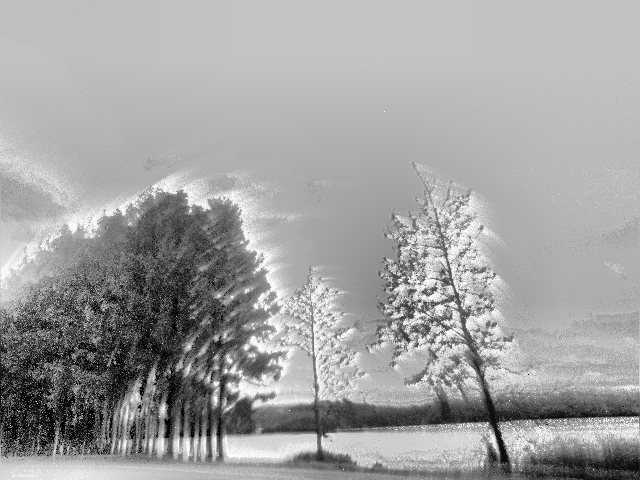}  \\
			\large(a). CF & \large (b). CF + frame augmentation & \large(c). CF  & \large (d). CF + frame augmentation \\
			\includegraphics[width=1\linewidth]{ablation_add_cf.png} &
			\includegraphics[width=1\linewidth]{ablation_add_cf_aug.png}&
			\includegraphics[width=1\linewidth]{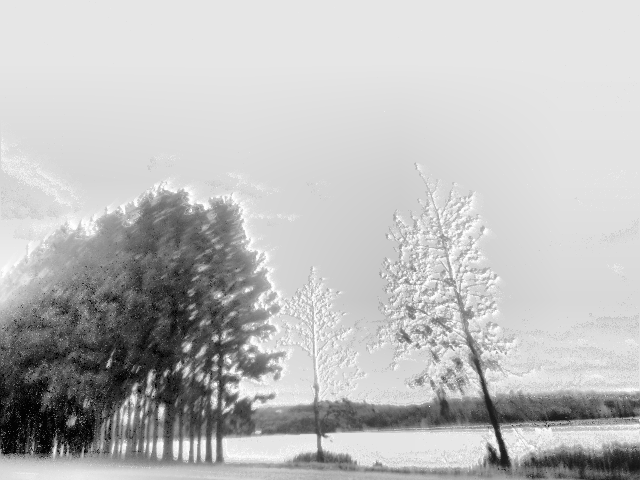}
			&
			\includegraphics[width=1\linewidth]{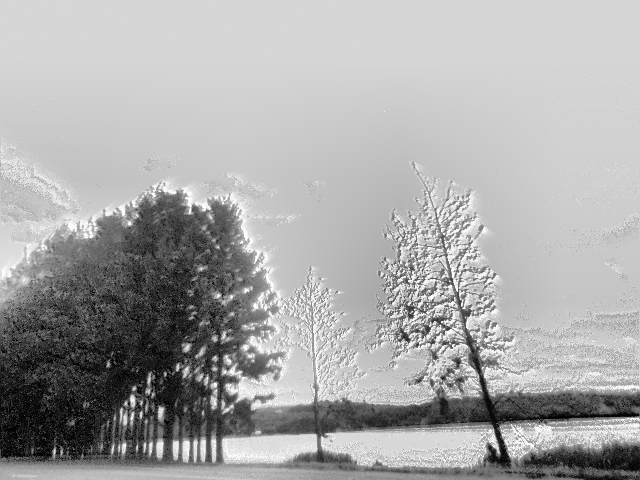}
			\\
			\large(e). AKF w/o frame augmentation  &\large (f). Full AKF pipeline &\large (g). AKF w/o frame augmentation  &\large (h). Full AKF pipeline
			\\
		\end{tabular}
	}
	\caption{\label{fig:ablation_cf}
		An example of applying our frame augmentation preprocessing algorithm on CF.
		Adding the frame augmentation step helps to reduce the `double edges' in the \texttt{Shapes\_6dof} with fast camera motion.
		However, it fails to recover the HDR region in (d) (\eg the white trialling shadows in \texttt{Tree\_rotation}).
		In the opposite, adding Kalman gain to CF, the same as AKF w/o frame augmentation, highly improves the HDR reconstruction in (g) but fails in high speed dataset in (e).
	}
\end{figure*}

Figure~\ref{fig:dsec} shows results of the DSEC dataset~\cite{Gehrig21ral} while Figure~\ref{fig:hdr} and Table~\ref{tab:hdr} present both the qualitative and quantitative results from our HDR dataset.
We also qualitatively analyse some challenging sequences from the popular open-source DAVIS event camera datasets ACD~\cite{Scheerlinck18accv}, CED~\cite{scheerlinck2019ced} and IJRR~\cite{Mueggler17ijrr} (Fig.~\ref{fig:davis}) and qualitatively evaluate on some benchmark image reconstruction sequences in IJRR~\cite{Mueggler17ijrr} (Table.~\ref{table: mean IJRR}).
To evaluate the asynchronous video reconstruction of the proposed algorithm, we analyse datasets drawn from Mueggler \etal \cite{Mueggler17ijrr} with fast camera motion.
We compare the EDI~\cite{Pan20pami}, CF and AKF methods in Figure~\ref{fig:edi comp}.
An ablation study of our AKF pipeline is provided in Figure~\ref{fig:ablation_edi} and~\ref{fig:ablation_cf}.

\noindent\textbf{Implementation details:}
The CF algorithm has a single parameter, the cutoff frequency.
This is chosen between 15-30 rad/s for the examples shown, details as noted in the discussion below.
For the AKF algorithm:
the event noise covariance $Q_{\bm{p}}$ \eqref{eq:event_covariance} is initialised to 0.01.
The event noise covariance tuning parameters \eqref{eq:event_covariance} are set to:
$\sigma_{\text{ref.}}^2 = 0.01$,
$\sigma_{\text{proc}}^2 = 0.0005$ and
$\sigma_{\text{iso.}}^2 = 0.03$.

\begin{table*}[t]
	\centering
	\caption{Comparison of state-of-the-art event-based video reconstruction methods E2VID \cite{Rebecq20pami}, ECNN \cite{Stoffregen20eccv}, Han \etal~\cite{han2020neuromorphic} and our CF and AKF on the proposed HDR and AHDR dataset.
		Metrics are evaluated over the full dataset of 10 sequences. 	
		Our AKF outperforms the compared methods on all metrics.
		Detailed evaluation of each sequence can be found in the supplementary material. Higher SSIM and Q-score and lower MSE indicate better performance.
	}
	\label{tab:hdr}
	\resizebox{1\textwidth}{!}{ 
		\begin{tabular}{ l | c c c c c | c c c c c | c c c c c }
			Metrics &  \multicolumn{5}{c|}{MSE ($\times 10^{-2}$) $\downarrow$}  &  \multicolumn{5}{c|}{SSIM~\cite{Wang04tip} $\uparrow$}
			&    \multicolumn{5}{c}{Q-score~\cite{Narwaria15jei} $\uparrow$}
			\\
			\midrule
			Methods & E2VID \cite{Rebecq20pami} & ECNN \cite{Stoffregen20eccv}  & Han \etal~\cite{han2020neuromorphic}  & CF (ours) & \textbf{AKF (ours)} & E2VID \cite{Rebecq20pami} & ECNN \cite{Stoffregen20eccv}  & Han \etal~\cite{han2020neuromorphic}  & CF (ours) & \textbf{AKF (ours)} & E2VID \cite{Rebecq20pami} & ECNN \cite{Stoffregen20eccv}  & Han \etal~\cite{han2020neuromorphic}  & CF (ours) & \textbf{AKF (ours)}
			\\
			\midrule
			\midrule
			HDR & $8.73 $ & $12.33 $ & $ 10.60  $  & $2.37 $  & $\textbf{1.84} $  & $0.58 $ & $0.37 $& $  0.64  $   & $0.84  $ & $\textbf{0.89} $  & $4.53 $ & $3.51 $ & $  5.02 $   & $5.17 $ & $\textbf{6.04} $
			
			\\
			\midrule
			AHDR & $11.57 $ & $21.17 $  & $  8.29  $   & $4.25 $  & $\textbf{3.94} $ & $0.50 $ & $0.04 $  & $  0.51 $   & $0.73  $ & $\textbf{0.79}$  & $5.25 $ & $3.36 $  & $  5.01 $   & $5.85 $ & $\textbf{6.48} $
			\\
			\bottomrule[\heavyrulewidth]
		\end{tabular}
	}
\end{table*}

\begin{table*}[t!]
	\centering
	\caption{\label{table: mean IJRR}Comparison of state-of-the-art methods of event-based video reconstruction on IJRR~\cite{Mueggler17ijrr} DAVIS datasets, showing only the mean values. See evaluation on each sequence in the supplementary material.
		Our CF and AKF perform well in the structural similarity metrics SSIM~\cite{Wang04tip} and LPIPS~\cite{Zhang18cvprLPIPS}.
		AKF outperforms other methods with a significant margin in the absolute intensity metrics MSE.
	}
	\label{tab:ijrr_mean}
	\resizebox{1\textwidth}{!}{ 
		\begin{tabular}{ l | c c c c | c c c c | c c c c }
			\toprule[\heavyrulewidth]\toprule[\heavyrulewidth]
			Metrics &  \multicolumn{4}{c|}{MSE ($\times 10^{-2}$) $\downarrow$}   &    \multicolumn{4}{c|}{SSIM~\cite{Wang04tip} $\uparrow$}  &    \multicolumn{4}{c}{LPIPS~\cite{Zhang18cvprLPIPS} $\downarrow$}
			\\
			\midrule
			Methods & E2VID \cite{Rebecq20pami} & ECNN \cite{Stoffregen20eccv} & CF (ours) & \textbf{AKF (ours)} & E2VID \cite{Rebecq20pami} & ECNN \cite{Stoffregen20eccv} & CF (ours) & \textbf{AKF (ours)} & E2VID \cite{Rebecq20pami} & ECNN \cite{Stoffregen20eccv} & \textbf{CF (ours)} & \textbf{AKF (ours)}
			\\
			\midrule
			Mean & $20.67 $ & $6.41 $  & $0.21 $ & $\textbf{0.18} $ & $0.43 $ & $0.56 $ & $0.85 $ & $\textbf{0.86} $ & $0.37 $ & $0.24 $  & $\textbf{0.19} $ & $\textbf{0.19} $ \\
			\bottomrule[\heavyrulewidth]
		\end{tabular}
	}
\end{table*}

\subsection{Main Results}
\noindent\textbf{DSEC Stereo Hybrid Event-Frame Datasets:}
In Figure~\ref{fig:dsec}, the five compared methods are tested on driving sequences in the city at night, high contrast tunnel and some other outdoor challenging scenes drawn from the DSEC dataset~\cite{Gehrig21ral}.

E2VID~\cite{Rebecq20pami} produces smoothed results but most of the night drive scenes are too dark.
It may be that E2VID~\cite{Rebecq20pami} network is trained on a simulated DAVIS dataset and the learned event noise model is not transferrable to a Prophesee event camera used in the DSEC dataset~\cite{Gehrig21ral}.
ECNN~\cite{Stoffregen20eccv} achieves a much better result than E2VID~\cite{Rebecq20pami} in this dataset.
It is likely that this is because ECNN~\cite{Stoffregen20eccv} used a training dataset with higher variation and a larger range of contrast threshold among sequences.
The algorithm still generates a `dirty mist' type of artefact in areas with noisy or insufficient event data, such as less textured road and dark background in the night driving sequences.
Han \etal~\cite{han2020neuromorphic} fuses paired E2VID~\cite{Rebecq20pami} reconstruction and LDR input images.	
	The network is able to combine the useful information from both images and events.
For example, the bright tunnel exit and the reflective road from events, and the dense intensity information in well-exposed areas from LDR frames.
	However, the HDR guidance they use is based on intensity images reconstructed by a pure-event network~\cite{Rebecq20pami} that has lost the true brightness changes of the scene.
As a consequence, the absolute intensity level of the raw output images are inaccurate.
The network also does not compensate for blur in the input LDR data, so the reconstructed images suffer from motion blur; for example, the roadside grasses and the car in the tunnel.
Our stochastic filter CF and AKF achieve much better results than the deep-learning-based methods.
CF is clearly able to reconstruct dark objects in the night scenes well, but the reconstructions are still blurry and noisy compared to AKF.
For example, AKF provides much cleaner reconstructions with sharp roadside buildings, trees, grass, cars, fences and people.
Video reconstructions from AKF also have less `shadowing effect' compared to CF, such as the black shadows trailing behind the white dividing lines in all driving datasets.

\noindent\textbf{HDR/AHDR Stereo Hybrid \EventFrame Datasets:}
The qualitative evaluation samples of LDR input, reconstructed and reference images of our proposed HDR/AHDR dataset are demonstrated in Fig.~\ref{fig:hdr}. The qualitative comparisons are shown in Table~\ref{tab:hdr}.

In the first row of Fig.~\ref{fig:hdr}, the proposed HDR dataset
\texttt{Trees} includes some underexposed trees (left-hand side) and two overexposed trees (right-hand side).
In the second row,
our AHDR sequence \texttt{Mountain} is artificially saturated (pixel values higher than 175 or lower than 80 of an 8-bit image, removing most of the detail.
E2VID~\cite{Rebecq20pami} reconstructs the two right-hand trees correctly, although the relative intensity of the tree is too dark.
But E2VID~\cite{Rebecq20pami} performs poorly in the dark area in \texttt{Trees} on the bottom left corner and skies/road in \texttt{Mountain} where it lacks events.
ECNN~\cite{Stoffregen20eccv} performs poorly because it is sensitive to noisy events in this dataset, \eg fewer events near the camera rotation centre in \texttt{Trees} and hot pixels in the sky of AHDR sequence \texttt{Mountain}.
Han \etal~\cite{han2020neuromorphic} is able to reconstruct the over- and under-exposed area in \texttt{Trees}, but the road in \texttt{Mountain} is not clear and the reconstructions suffer from inaccurate intensity level and artefacts near the image edges.
Our CF fuses event data and low dynamic range frames properly but exhibits a `shadowing effect' on object edges (trees and mountain road).
The artefacts are significantly reduced in reconstruction from our AKF by dynamically adjusting the per-pixel Kalman gain according to our proposed event and frame uncertainty model.
In addition, comparing to the obvious rotation motion that appears on the left-hand of CF reconstruction on \texttt{Trees}, the deblurring model in AKF also generates much sharper trunks.

We also quantitatively evaluated image reconstruction quality
with the HDR reference in the proposed HDR and AHDR dataset using the following metrics:
Mean squared error (MSE),
structural similarity Index Measure (SSIM) \cite{Wang04tip},
and Q-score~\cite{Narwaria15jei}.
SSIM measures the structural similarity between the reconstructions and references.
Q-score is a metric tailored to HDR full-reference evaluation.
All metrics are computed on the un-altered reconstruction and raw HDR intensities.

Table~\ref{tab:hdr} shows that our AKF outperforms other methods on the proposed HDR/AHDR dataset on MSE, SSIM and Q-score.
Unsurprisingly, our AKF outperforms E2VID~\cite{Rebecq20pami} and ECNN~\cite{Stoffregen20eccv} since it utilises frame information in addition to events.
The AKF outperforms state-of-the-art pure-event method E2VID~\cite{Rebecq20pami} and event-frame method Han \etal~\cite{han2020neuromorphic} in the absolute intensity error MSE with a significant reduction of 71.5\% and 69.4\% respectively.
The image similarity metrics SSIM and Q-score are improved by 55.6\% and 28.0\% for E2VID~\cite{Rebecq20pami}, and 46.1\% and 24.8\% for Han \etal~\cite{han2020neuromorphic}.

\noindent\textbf{DAVIS Datasets:}
The qualitative evaluation of the event-based reconstruction methods on challenging HDR scenes from the DAVIS datasets ACD~\cite{Scheerlinck18accv}, CED~\cite{scheerlinck2019ced} and IJRR~\cite{Mueggler17ijrr} are shown in Figure ~\ref{fig:davis}.
\texttt{Night drive} investigates extreme low-light, fast-speed, night driving scenarios with blurry and underexposed/overexposed DAVIS frames.
\texttt{Outdoor running} evaluates the overexposed outdoor scene with event camera noise.
\texttt{Box\_6dof} evaluates the highly textured, HDR indoor environment with fast camera motion.
E2VID~\cite{Rebecq20pami} and ECNN~\cite{Stoffregen20eccv} are able to capture HDR objects (\eg right turn sign in \texttt{Night drive}, box textures in \texttt{Box\_6dof}),
but the washed-out result of E2VID~\cite{Rebecq20pami} loses the overall scene intensity and ECNN~\cite{Stoffregen20eccv} introduces artefacts.
In \texttt{Outdoor running}, E2VID~\cite{Rebecq20pami} and ECNN~\cite{Stoffregen20eccv} are unable to reproduce the correct high dynamic range intensity between the dark road and bright left building and sky background.
Event noise in the middle area of \texttt{Outdoor running} is obvious in the reconstruction of  ECNN~\cite{Stoffregen20eccv} as well.
Han \etal~\cite{han2020neuromorphic} uses both frame and event data but the performance is affected by the blurry input LDR images.
For example, the `right-turn' sign is clear in the E2VID \cite{Rebecq20pami} reconstruction, but after fusing with the blurry input LDR image, the `right-turn' sign information is lost.
The output reconstructions lose the absolute brightness information of LDR images, generating `washed-out' images with inaccurate intensities.
Our CF also experiences blur from fast camera motion, and it is susceptible to hot pixel and event noise.
Similarly, CF exhibits a `shadowing effect' on object edges (on the trailing edge of road signs and buildings) due to using a single (constant value) cutoff frequency for all pixels.
Our AKF algorithm dynamically adjusts the Kalman gains governing the data fusion on a per-pixel uncertainty model.
As a consequence, it is able to clearly resolve the right turn arrow on the street sign and does not exhibit the same `shadowing effect' seen in the CF.
AKF is able to resolve distant buildings despite the fact that they are too bright and washed-out in the LDR DAVIS frame.
Furthermore, it is able to compensate for the hot pixels present in all sequences.
The frame augmentation sharpens the blurry DAVIS frame and reduces the temporal mismatch between the high data rate events and the low data rate frames.
AKF reconstructs the sharpest and most detailed HDR objects in all challenging scenes.

The IJRR dataset~\cite{Mueggler17ijrr} acts as a benchmark event camera dataset that is popularly used in video reconstruction.
Though the dataset is not targeted at HDR data and does not provide HDR references for quantitative evaluation, it is still possible to evaluate the ability of an algorithm to reconstruct an unknown ground truth image by sub-sampling the frame data.
To quantitatively evaluate on DAVIS datasets, we use every second image frame as input for the algorithm and then take the intermediate image frame as ground truth to evaluate the quality of reconstruction using the quantitative metrics.
Since the Han \etal~\cite{han2020neuromorphic} algorithm only reconstructs images at the timestamps of the LDR input data, we cannot evaluate it for Table~\ref{tab:ijrr_mean} which would require continuous-time video reconstruction.
In addition to the MSE and SSIM, we also evaluated on the learned perceptual image patch similarity (LPIPS) \cite{Zhang18cvprLPIPS}.
Table~\ref{tab:ijrr_mean} shows that for MSE and SSIM, the AKF is always the best, though CF is roughly equal on LPIPS.
We believe the reduced gap in performance between CF and AKF is due to `cleaner' frame data that is not under/overexposed as in our HDR dataset.
The large difference in MSE between AKF and E2VID~\cite{Rebecq20pami}/ECNN~\cite{Stoffregen20eccv} is not as apparent in SSIM and LPIPS, agreeing with the intuition that pure event reconstruction methods are relatively more faithful to scene structure than the absolute intensity.
See evaluation on each sequence in the supplementary material Table 4.

The AKF and CF achieve noticeably better performance in stereo hybrid event-frame dataset DESC~\cite{Gehrig21ral} and HDR/AHDR relative to the comparison algorithms.
This is due to the improved quality of the data in these sequences, where the event and frame data come from separate sensors with improved noise characteristics, resolution and frame rate.
A clear advantage of explicit algorithms such as the CF and the AKF is their ability to exploit high quality data directly, while learning algorithms such as E2VID~\cite{Rebecq20pami}, ECNN~\cite{Stoffregen20eccv} must be retrained.
Moreover, the authors believe that as the quality of the data improves the advantage of the learned methods, the ability to use priors in rebuilding images where there is poor information, will decrease.
The results presented demonstrate the potential of the CF and AKF to outperform learning based methods on good quality datasets.

\noindent\textbf{Asynchronous Reconstruction Methods Evaluation:}
In addition to evaluating image reconstruction under challenging lighting conditions with HDR methods E2VID~\cite{Rebecq20pami} and ECNN~\cite{Stoffregen20eccv}, in this section, we demonstrate the inter-frame reconstruction performance between EDI~\cite{Pan20pami}, CF and AKF in Fig.~\ref{fig:edi comp}.
In contrast to the batch image reconstruction method E2VID~\cite{Rebecq20pami} or ECNN~\cite{Stoffregen20eccv}, which accumulate sufficient events in a window to reconstruct a single image, asynchronous methods are able to update video continuously.

The dataset we used comes from the open-source event camera dataset IJRR~\cite{Mueggler17ijrr} with fast motions.
Fig.~\ref{fig:edi comp} (a) shows two consecutive intensity frames from an event camera in a single image. The current frame is shown in red and the previous frame is shown in blue.
The relatively big distance between images is because of fast camera motion, making inter-frame reconstruction challenging.
Fig.~\ref{fig:edi comp} (b)-(d) demonstrate image reconstruction between two frames by asynchronous reconstruction methods EDI~\cite{Pan20pami}, our CF and AKF methods generate video continuously and the performance is evaluated at the intermediate timestamp between two frames.
Fig.~\ref{fig:edi comp} (b) demonstrates the inter-frame reconstruction from EDI~\cite{Pan20pami}.
It deblurs the previous raw image to a sharp image
and then accumulates events to the deblurred image until the middle inter-frame timestamp.
The simple direct event integration model causes obvious image boundary shadows of the previous image.
This is because in the direct integration process, EDI\cite{Pan20pami} accumulates events with the same contrast threshold for all pixels,
missing or noisy events, and inaccurate event contrast threshold lead to a mismatch between frame and event data.
Hence, direct integration fails to update pixels at boundaries completely.

In Fig.~\ref{fig:edi comp} (c), our filter-based image reconstruction method CF `forgets' the previous information with a constant filter gain.
Its zero-order-hold assumption requires small motion between frames.
In fast motion scenarios, it integrates high temporal resolution events with the latest low temporal resolution frame.
Mismatched intensity changes lead to trailing shadows behind objects in motion direction.
The intensity interpolation and per-pixel contrast threshold scaling of frame augmentation in our full AKF pipeline overcome the limitations of the other two methods by increasing frame temporal resolution and matching intensity changes between frames and event data dynamically.
Fig.~\ref{fig:edi comp} (d) shows that our AKF achieves the best inter-frame reconstruction.

\subsection{Computational Complexity}
The time complexity of asynchronous filtering methods, AKF and CF, are both event-wise operations, where they perform a linear combination of several  $\mathcal{O}(1)$ operations, resulting in an overall $\mathcal{O}(N)$ complexity for $N$ events.
Within the frame augmentation step, the complexity of deblurring is $\mathcal{O}(P)$ and the complexity of contrast threshold calibration is $\mathcal{O}(N+M)$, where $P$ is the number of events that arrive during the exposure time, $N$ is the number of events between two images excluding exposure time, and each image contains $M$ pixels.
Notably, deblurring and contrast threshold calibration are only computed once per image.
The computational complexity of the in-frame augmentation step can be considered as $\mathcal{O}(P+N+M)$.
The framework is implemented in MATLAB.
Processing 10 images and more than 4 million events with resolution 480 $\times$ 640 during the period takes our full AKF pipeline around than 1 second per image in average, on a single Intel core i7-7700K CPU running at 4.20 GHz.
It should be noted that our AKF algorithm is well suited to a parallel pixel-by-pixel embedded implementation such as in FPGA or in sensor chip hardware.
The present code is purely targeted at demonstrating the potential of the approach and the question of real-time implementation is beyond the scope of the present work.

\subsection{Ablation Study}\label{sec:ablation study}
The HDR image reconstruction performance of our AKF benefits from three key contributions of the pipeline, the frame augmentation process, the complementary filter and the asynchronous Kalman gain.
To validate the efficacy of each module, we provide an ablation study as follows.

\noindent\textbf{Frame Augmentation:}
The frame augmentation includes a deblur schema by EDI~\cite{Pan20pami} and a sophisticated temporal interpolation.
For reconstruction on frame timestamps (not inter-frame),
the frame augmentation is based directly on the established performance of the EDI~\cite{Pan20pami}.
To analyse the efficacy of AKF above and beyond the frame augmentation step, EDI~\cite{Pan20pami} and the full AKF pipeline are compared in Fig.~\ref{fig:ablation_edi}.
In our frame augmentation step, EDI~\cite{Pan20pami} generates sharp images by associating frames to events that arrive within the exposure time.
Since EDI~\cite{Pan20pami} is not designed for HDR~\cite{Pan20pami} and the exposure time of each frame is relatively small, only pixels with events that occurred during the short time can be recovered properly, such as the grid edges in Fig.~\ref{fig:ablation_edi} (b).
Moreover, the frame camera of dataset  \texttt{Interlake\_01a} in Fig.~\ref{fig:ablation_edi} (e) was set to auto-exposure mode, and insufficient events arrive within the small exposure time in such a strong sun reflective scene.
In addition, in highly reflective regions in Fig.~\ref{fig:ablation_edi} (d), intensity changes can be extremely fast, so pixels in these areas are likely to reach the refractory period, leading to lots of skipped events.
Without using historic \eventframe information or dynamically adjusting contrast threshold in HDR regions as our full AKF pipeline, EDI~\cite{Pan20pami} only models the frame irradiance accumulation during the exposure time and events that happens at the time.
As a deblurring method, it recovers some of the HDR regions surprisingly well although the overall reconstruction is clearly still overexposed.
Fig.~\ref{fig:ablation_edi} (c) and (f) contains much more grid and road texture under the strong sun reflection.
The AKF recovers detailed intensities for the overexposed region by exploiting the inherent `memory' in the state of the filter.
Also, the uncertainty of pixels in the reflective region is much higher than the rest of the image, so AKF relies heavily on the historic filter state and event data, generating clear object textures in the challenging HDR scenarios.

\noindent\textbf{Kalman Gain:}
To investigate relative contribution of the frame augmentation and Kalman gain modules under different conditions, we conduct an ablation study on a fast motion \texttt{Shapes\_6dof} dataset and an HDR dataset \texttt{Tree\_rotation}.

Similar to Fig.~\ref{fig:edi comp} (c), the zero-order-hold assumption of CF fails in fast motion scenario in Fig.~\ref{fig:ablation_cf} (a).
The mismatched event data and low temporal resolution frames cause reconstruction error in inter-frame filter state update.
Comparing to Fig.~\ref{fig:ablation_cf} (b), the temporal intensity interpolation provides CF with accurate reference images for data fusion between frames.
Hence, most of the `double edges' that are caused by delayed reference images are removed.
It is the same for AKF.
Without frame augmentation, AKF fuses event data and the latest frame with per-pixel Kalman gain.
The result in Fig.~\ref{fig:ablation_cf} (e) also show obvious shadow associated to mismatched reference frame, which is similar to CF in Fig.~\ref{fig:ablation_cf} (a).
Because DAVIS frames are reliable in this dataset (with no over or underexposed region), dynamically computing Kalman gain does not make an obvious difference.
However, adding frame augmentation makes a difference in datasets with fast camera motion.
The full AKF pipeline achieves clear reconstruction in Fig.~\ref{fig:ablation_cf} (f).

An HDR scenario \texttt{Tree\_rotation} is shown in Fig.~\ref{fig:ablation_cf} (c)-(d) and (g)-(h).
Adding frame augmentation from Fig.~\ref{fig:ablation_cf} (c) to (d) minimally impacts the reconstruction performance of CF, but computing Kalman gain based on frame and event data uncertainty significantly improves the HDR reconstruction from (c) to (g).
This is because frames are not reliable under extreme lighting conditions; for example, most of the sky pixels are too bright and saturated in \texttt{Tree\_rotation}.
Including the Kalman gain can balance data fusion pixel-by-pixel by identifying pixels with large uncertainty in frames and relying more on event data and historic filter state.
But in the CF, a constant exponential decay rate is used for all pixels, leading to the white trailing shadows behind trees in Fig.~\ref{fig:ablation_cf} (c) and (d).
From Fig.~\ref{fig:ablation_cf} (d) and (h), replacing the constant complementary gain with Kalman gain also improves the HDR reconstruction performance.
Compared to AKF without frame augmentation (g), our full AKF pipeline (h) slightly improves performance by deblurring and increasing frame temporal resolution.
Even though \texttt{Tree\_rotation} is not a high speed dataset, the reconstruction of the full AKF pipeline in (h) is much sharper than (g).
Our ablation study shows that frame augmentation improves performance in fast motion scenes while Kalman gain estimation is important in HDR image reconstruction.

\subsection{Limitations}
As a preliminary model, our CF fuses zero-order-hold frames and high temporal resolution event data using a constant gain for all pixels.
The zero-order-hold assumption of low-frequency frames causes time delay between frame and event data in fast motion scenes, leading to `double edges' in reconstructions.
The constant filter gain also limits the HDR property of CF.
Our full AKF pipeline overcomes the limitation of low temporal resolution of frame data by event-based temporal interpolation between frames.
AKF achieves better HDR reconstruction than CF by using a dynamically adjusted filter gain according to the proposed frame-event noise model.
The superior performance of AKF is at the cost of additional complexity versus CF, and susceptibility to non-Gaussian noise in event data.

As an explicit algorithm, AKF reconstructs event-frame data to video directly based on a camera statistical noise model.
Since the AKF requires no training process, it cannot learn complex image priors from data like other machine learning approaches.
Even with our proposed noise model, the noise tolerance of AKF is generally lower than that of learning-based methods.
The authors believe, however, that as event sensor technology improves this limitation is likely to become less and less important.

\subsection{Event Convolutions}
The experiments were performed using a number of challenging open-source datasets~\cite{Mueggler17ijrr,Scheerlinck18accv,Gehrig21ral}.
The internal filter state of the system is asynchronous and for visualisation we display instantaneous snapshots taken at sample times.
The complexity of our algorithm scales linearly with the number of (non-zero) elements in the kernel.
We set the contrast threshold $c$ to be constant, except during frame augmentation in~\eqref{eq:scaling factor}.

\begin{figure*}[t]
	\centering
	\resizebox{0.83\textwidth}{!}{\begin{tabular}{ >{\centering\arraybackslash} m{2.5cm} >{\centering\arraybackslash} m{4.5cm} >{\centering\arraybackslash} m{4.5cm} >{\centering\arraybackslash} m{4.5cm} >{\centering\arraybackslash} m{4.5cm}}
			&\large \texttt{night\_drive} &\large \texttt{boxes\_6dof} &\large \texttt{city09d} &\large \texttt{interlaken\_01a}
			\\
			Identity
			
			&
			\includegraphics[width=\kernelwidth]{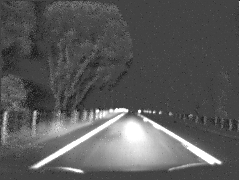}
			&
			\includegraphics[width=\kernelwidth]{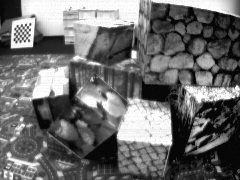}
			&
			\includegraphics[width=\kernelwidth, trim={1cm 1cm 1cm 1cm},clip]{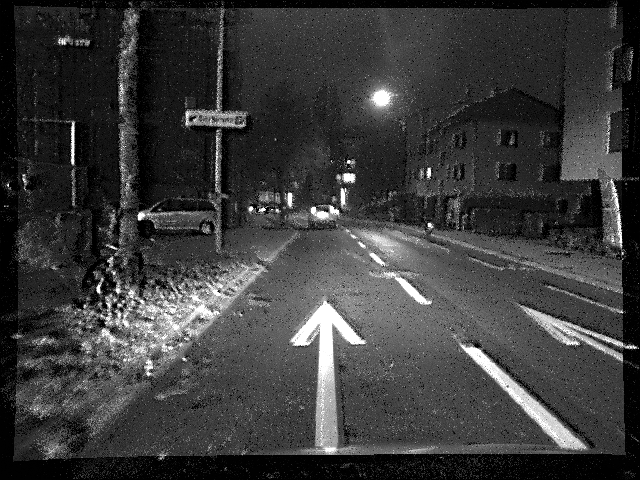}
			&
			\includegraphics[width=\kernelwidth, trim={1cm 1cm 1cm 1cm},clip]{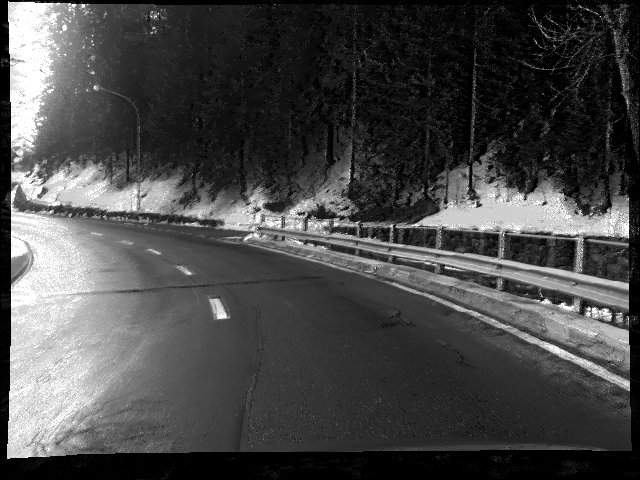}
			\\
			Gaussian
			
			\medskip
			$
			$ &
			\includegraphics[width=\kernelwidth]{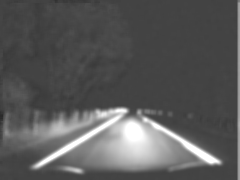}
			&
			\includegraphics[width=\kernelwidth]{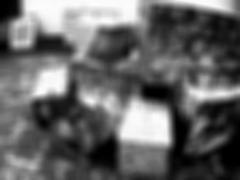}
			&
			\includegraphics[width=\kernelwidth, trim={1cm 1cm 1cm 1cm},clip]{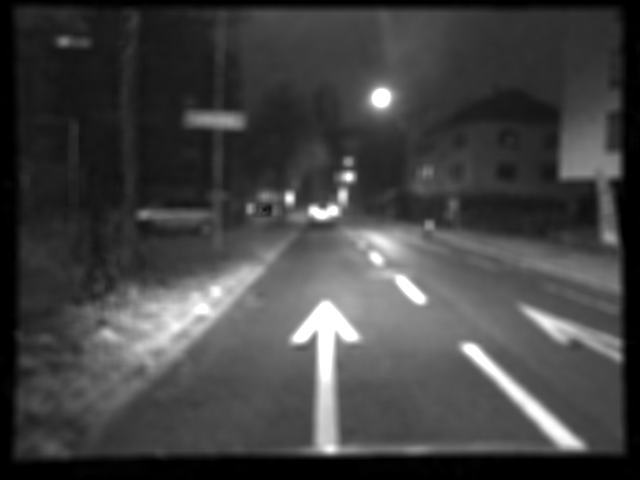}
			&
			\includegraphics[width=\kernelwidth, trim={1cm 1cm 1cm 1cm},clip]{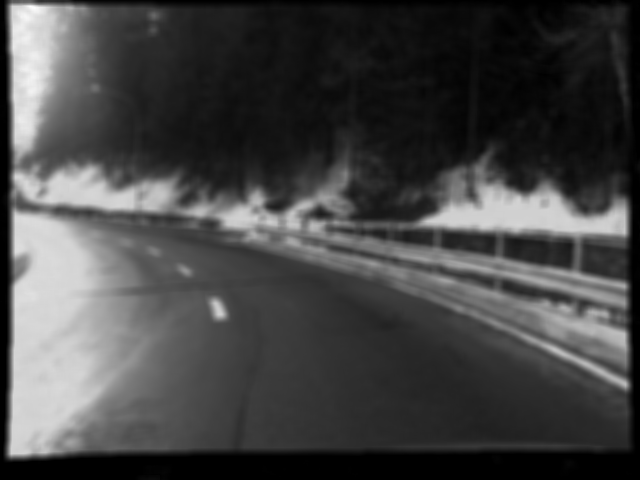}
			\\
			Sobel $x$
			
			&
			\includegraphics[width=\kernelwidth]{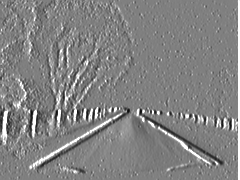}
			&
			\includegraphics[width=\kernelwidth]{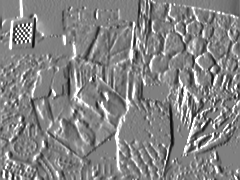}
			&
			\includegraphics[width=\kernelwidth, trim={1cm 1cm 1cm 1cm},clip]{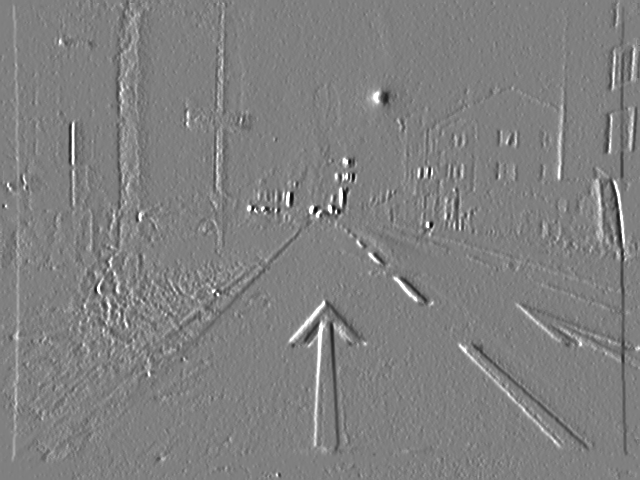}
			&
			\includegraphics[width=\kernelwidth, trim={1cm 1cm 1cm 1cm},clip]{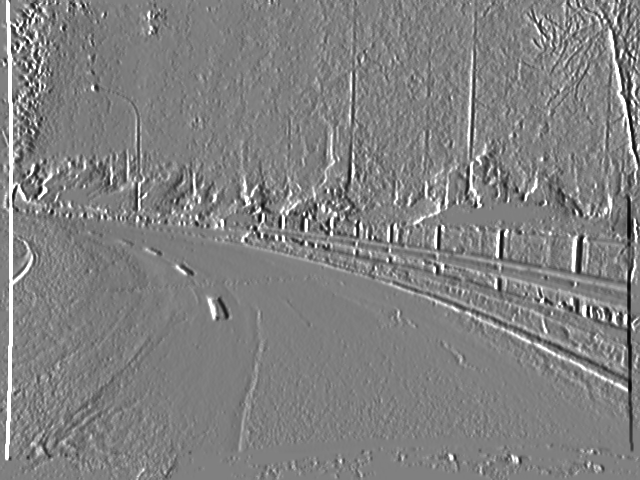}
			\\
			Sobel $y$
			
			&
			\includegraphics[width=\kernelwidth]{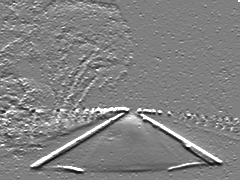}
			&
			\includegraphics[width=\kernelwidth]{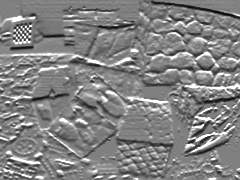}
			&
			\includegraphics[width=\kernelwidth, trim={1cm 1cm 1cm 1cm},clip]{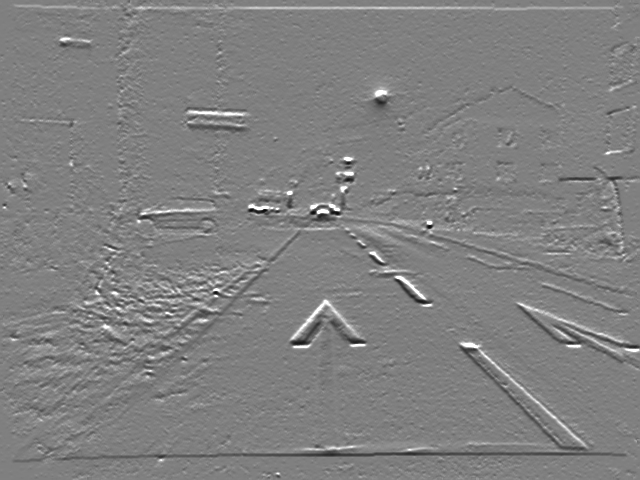}
			&
			\includegraphics[width=\kernelwidth, trim={1cm 1cm 1cm 1cm},clip]{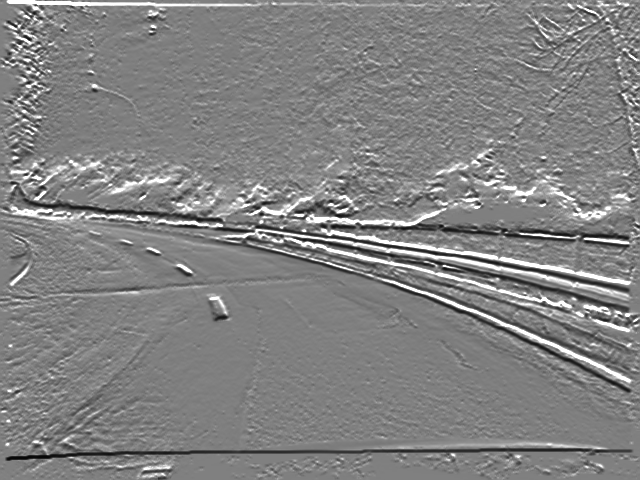}
			\\
			Colour coded gradient from Sobel
			
			\medskip
			
			\includegraphics[width=0.15\columnwidth]{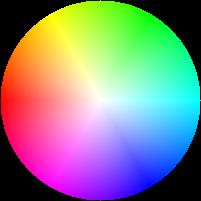}
			
			&
			\includegraphics[width=\kernelwidth]{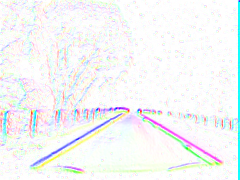}	
			&
			\includegraphics[width=\kernelwidth]{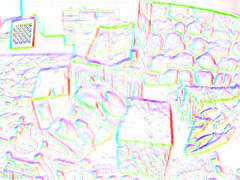}	
			&
			\includegraphics[width=\kernelwidth, trim={1cm 1cm 1cm 1cm},clip]{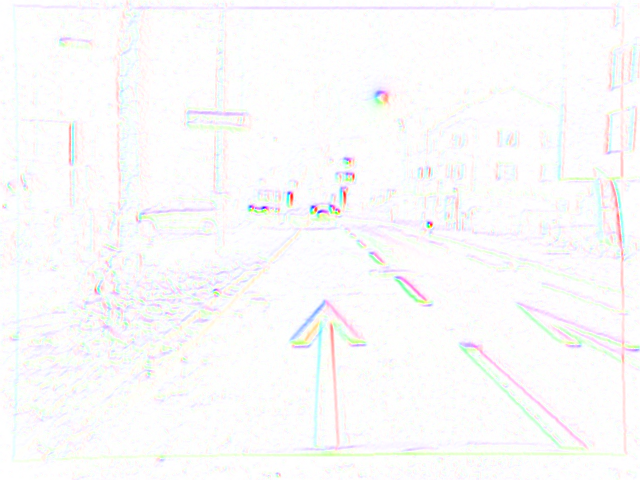}
			&
			\includegraphics[width=\kernelwidth, trim={1cm 1cm 1cm 1cm},clip]{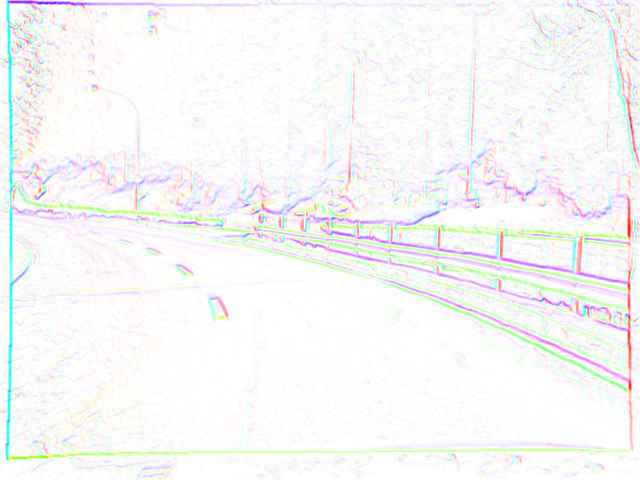}
			\\
			Laplacian

			&
			\includegraphics[width=\kernelwidth]{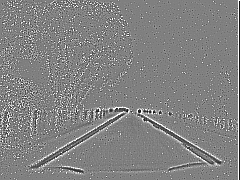}
			&
			\includegraphics[width=\kernelwidth]{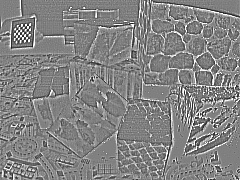}
			&
			\includegraphics[width=\kernelwidth, trim={1cm 1cm 1cm 1cm},clip]{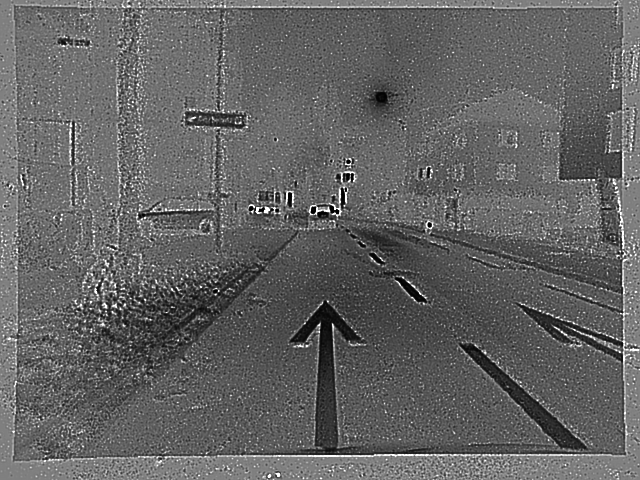}
			&
			\includegraphics[width=\kernelwidth, trim={1cm 1cm 1cm 1cm},clip]{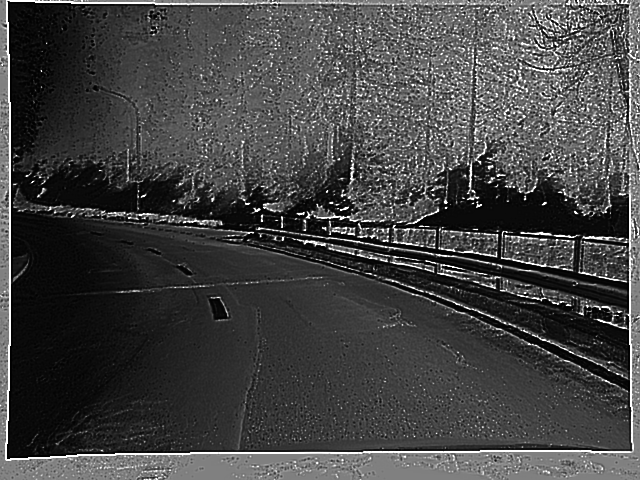}
		\end{tabular}	}
	\caption{Different kernels $K$ applied to events using our proposed pipeline~\ref{fig: pipeline}. \texttt{night\_drive}~\cite{Scheerlinck18accv} and \texttt{city09d}~\cite{Gehrig21ral} are captured in challenging night driving conditions, demonstrating excellent performance in low-light settings thanks to the event camera. \texttt{boxes\_6dof}~\cite{Mueggler17ijrr} is captured in an indoor setting with cluttered scene. \texttt{interlaken\_01a} is captured in day time driving with very dark and very bright regions. Note that the last four convolution results for \texttt{city09d} and \texttt{interlaken\_01a} are not as clear compared to the others due to the camera having a higher resolution.
	}
	\label{fig:kernels}
	\vspace{-2mm}
\end{figure*}

Fig.~\ref{fig:kernels} displays a range of different filtered versions of input sequences (\texttt{night\_drive}~\cite{Scheerlinck18accv}, \texttt{boxes\_6dof}~\cite{Mueggler17ijrr} and \texttt{city09d}~\cite{Gehrig21ral}).
The first row of Fig.~\ref{fig:kernels} shows the application of the identity kernel.
This kernel returns the identical result to our previously demonstrated full AKF pipeline.
The sequences that follow, for a range of different kernels, are generated using the proposed algorithm and convincingly appear as one would expect if the kernel had been applied to the image reconstruction from the top row.
The key advantage of the proposed approach is that it does not incur latency or additional computation associated with first requiring the image to be reconstructed before the computation of the desired convolution.
For example, to obtain a high-frequency convolution output, the computational complexity $\mathcal{O}(n)$ is significantly smaller when compared to applying convolution over the reconstructed image is $\mathcal{O}(qwh)$, where $n$ is the number of events, $q$ is the required output frequency, and $(w,h)$ are the width and height of the image.

The sequences in Fig \ref{fig:kernels} are:
\begin{itemize}
	\item \texttt{night\_drive} \emph{(left)}:
	Country road at night with no street lights or ambient lighting, only headlights.
	The car is travelling at 80km/h, causing considerable motion in the scene. This sequence focuses on performance in high-speed, low-light conditions.

\item \texttt{boxes\_6dof} \emph{(second left)}: A DAVIS240C camera move around a cluttered indoor environment in six degrees of freedom under normal lighting condition.
	\item \texttt{city09d} \emph{(second right)}: A new night driving dataset using higher resolution Prophesee event cameras and normal RGB cameras. The car is driven in a normally lit night driving scenario, showing performance in a high dynamic range environment.

\item \texttt{interlaken\_01a} \emph{(right)}: A new day time driving dataset with higher resolution Prophesee event cameras and normal RGB cameras. The car is driving along a road, on a bright day with reflective road surface, snow covering the slope by the road, and a lot of shadows from the trees. It shows the performance of our convolution results in a very high dynamic range environment.
\end{itemize}

\noindent Despite the noise in the event stream, our approach reproduces a high-fidelity representation of the scene.
It is particularly interesting to note the response for the two challenging sequences \texttt{night\_drive}, \texttt{city09d} and \texttt{interlaken\_01a}.
In both cases, the image is clear and full of detail, despite the high dynamic range of the scene.

The second row computes a (spatial) low pass Gaussian filter of the sequences.
The low pass nature of the response is clear in the image.
Note that if it was desired to compute an image pyramid, then it is a straightforward generalisation of the filter equations to reduce the state dimension at a particular level of the image pyramid by a linear combination of pixel values.
The resulting filter would still be linear and the same filter equations would apply.

The third and fourth rows display the internal filter state for the Sobel kernels in both vertical and horizontal directions.
The fifth row shows the colour coded gradient of the image using the colour wheel as shown in Fig~\ref{fig:kernels}.
The results show that the derivative filter state operates effectively even in very low light and high dynamic range conditions.

Row six displays the Laplacian of the image.
The Laplacian kernel computes an approximation of the divergence of the gradient vector field.
It can be used for edge detection: zero crossings in the Laplacian response correspond to inflections in the gradient and denote edge pixels.
It is also possible to reconstruct an original (log) intensity image from a Laplacian image using Poisson solvers \cite{Agrawal05iccv,Agrawal06eccv}.

It is important to recall that the internal state of the filter is computed directly from the event and image stream in all these cases.
For example, if only the Laplacian is required, then there is no need to compute a grey-scale image or gradient image.


\section{Conclusion}
In this paper, we introduced an asynchronous linear filter architecture that fuses frames with event data asynchronously.
The estimated intensity and spatial convolution can be retrieved at the same temporal resolution as events, allowing for arbitrarily high framerate output.
The proposed filter also reconstructs HDR videos from LDR frames and event data by dynamically adjusting the Kalman filter gain based on a unifying event-frame uncertainty model.
Our method outperforms the state-of-the-art event-based HDR video reconstruction methods on publicly available datasets and our HDR hybrid \eventframe dataset.
Additionally, our architecture is integrated with the event-based spatial convolution of different linear kernels with an asynchronous update.
We believe our asynchronous linear filter architecture has practical applications for video acquisition in HDR scenarios using the extended power of event cameras in addition to conventional frame-based cameras.

\ifCLASSOPTIONcaptionsoff
  \newpage
\fi



%
\ifCLASSOPTIONcaptionsoff
\newpage
\fi

\bibliographystyle{IEEEtran}
\bibliography{pami}

%

\begin{IEEEbiography}[{\includegraphics[width=1in,height=1.25in,clip,keepaspectratio]{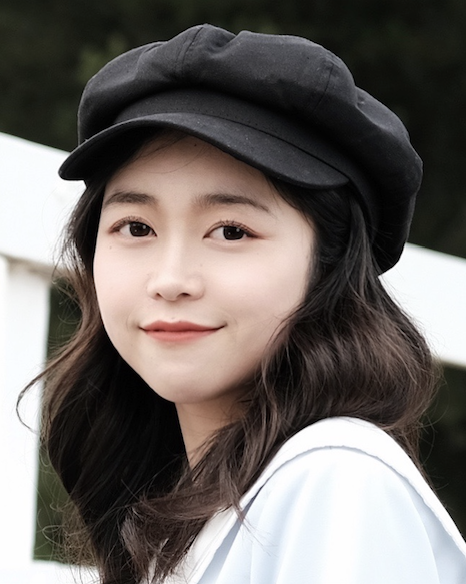}}]{Ziwei Wang}
is current a Ph.D. student in the Systems Theory and Robotics (STR) group, College of Engineering and Computer Science, Australian National University (ANU), Canberra, Australia.
She received her B.Eng degree from ANU (Mechatronics) in 2019.
Her interests include event-based vision, asynchronous image processing and robotics.
\end{IEEEbiography}

\begin{IEEEbiography}[{\includegraphics[width=1in,height=1.25in,clip,keepaspectratio]{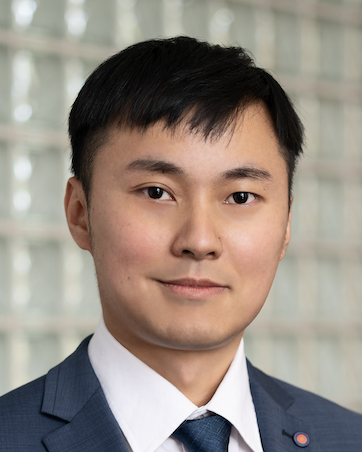}}]{Yonhon Ng}
is a research fellow at the Systems Theory and Robotics (STR) group, the Australian National University (ANU).
He received his BEng(R\&D) in 2013 (Mechatronics) and was awarded with a University Medal from the Australian National University.
He also received his PhD degree in 2018 (Systems Engineering) at the Australian National University.
His research interests include state estimation and control, 3D computer vision and robotics.
\end{IEEEbiography}

\begin{IEEEbiography}[{\includegraphics[width=1in,height=1.25in,clip,keepaspectratio]{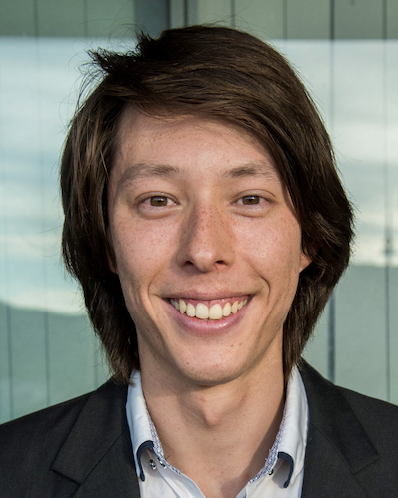}}]{Cedric Scheerlinck} is a Ph.D. graduate from the Systems Theory and Robotics (STR) group, Australian National University. He received his Master of Engineering degree from the University of Melbourne in 2016. His research interests include image reconstruction and deep learning with event cameras.
\end{IEEEbiography}

\begin{IEEEbiography}[{\includegraphics[width=1in,height=1.25in,clip,keepaspectratio]{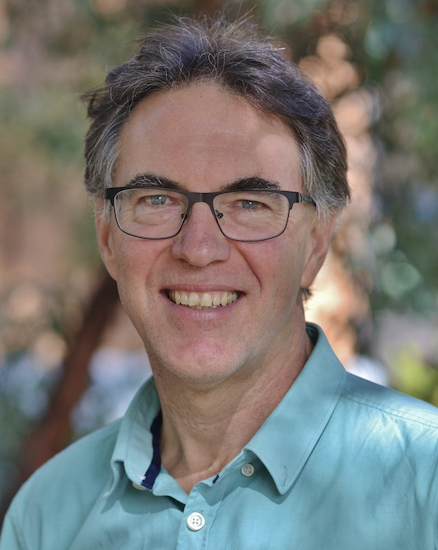}}]{Robert Mahony} is a Professor in the Research School of Engineering at the Australian National University. He is the lead of the Systems Theory and Robotics (STR) group. He received his BSc in 1989 (applied mathematics and geology) and his PhD in 1995 (systems engineering) both from the Australian National University. He is a fellow of the IEEE. His research interests are in non-linear systems theory with applications in robotics and computer vision. He is known for his work in aerial robotics, geometric observer design, robotic vision, and optimisation on matrix manifolds.
\end{IEEEbiography}
\appendices

\section{Event Camera Uncertainty} \label{sec: app Event Camera Uncertainty}
Stochastic models for event camera uncertainty are difficult to develop and justify~\cite{gallego2019event}.
In this paper, we propose a number of simple heuristics to model event noise as the sum of three pixel-by-pixel additive Gaussian processes.

\noindent
\textbf{Process noise:}
Process noise is a constant additive uncertainty in the evolution of the irradiance of the pixel, analogous to process noise in a Kalman filtering model.
Since this noise is realised as an additive uncertainty only when an event occurs, we call on the principles of Brownian motion to model the uncertainty at time $t^s_{\bm{p}}$ as a Gaussian process with covariance that grows linearly with time since the last event at the same pixel.
That is
\begin{align*}
Q^\text{proc.}_{\bm{p}}(t^s_{\bm{p}})
& = \sigma_{\text{proc.}}^2 (t^s_{\bm{p}} - t^{s-1}_{\bm{p}}),
\end{align*}
where $\sigma_{\text{proc.}}^2$ is a tuning parameter associated with the process noise level.

\noindent
\textbf{Isolated pixel noise:}
Spatially and temporally isolated events are more likely to be associated to noise than events that are correlated in group.
The noisy background activity filter~\cite{delbruck2008frame} is designed to suppress such noise and most event cameras have similar routines that can be activated.
Instead, we model an associated noise covariance by
\[
Q^\text{iso.}_{\bm{p}}(t^s_{\bm{p}})
= \sigma_{\text{iso.}}^2 \min \{t^s_{\bm{p}} - t^*_{N(\bm{p})}\},
\]
where $\sigma_{\text{iso.}}^2$ is a tuning parameter and $t^*_{N(\bm{p})}$ is the latest time-stamp of any event in a neighbourhood $N(\bm{p})$ of $\bm{p}$.
If there are recent spatio-temporally correlated events then $Q^\text{iso.}_{\bm{p}}(t^s_{\bm{p}}) $ is negligible, however, the covariance grows linearly, similar to the Brownian motion assumption for the process noise, with time from the most recent event.

\noindent
\textbf{Refractory period noise:}
Circuit limitations in each pixel of an event camera limit the response time of events to a minimum known as the refractory period $\rho > 0$ \cite{yang2015dynamic}.
If the event camera experience fast motion in highly textured scenes then the pixel will not be able to trigger fast enough and events will be lost.
We model this by introducing a dependence on the uncertainty associated with events that are temporally close to each other such that
\[
Q^\text{ref.}_{\bm{p}}(t^s_{\bm{p}}) =
\left\{
\begin{array}{cc}
0 & \text{if } t^s_{\bm{p}} - t^{s-1}_{\bm{p}} > \overbar{\rho}, \\
\sigma_{\text{ref.}}^2 & \text{otherwise},
\end{array}
\right.
\]
where $\sigma_{\text{ref.}}^2$ is a tuning parameter and $\overbar{\rho}$ is an upper bound on the refractory period.

\section{Conventional Camera Uncertainty} \label{sec: app Conventional Camera Uncertainty}
The noise of $I_{\bm{p}}(\tau^k)$ comes from uncertainty in the raw camera response $I^F_{\bm{p}}(\tau^k)$ mapped through the inverse of the Camera Response Function (CRF).
The uncertainty associated with sensing process $I^F_{\bm{p}}(\tau^k)$ is usually modelled as a constant variance Gaussian process
\cite{shin2005block, russo2003method} although for low light situations this should properly be a Poisson process model \cite{hasinoff2014photon}.
The quantisation noise is uniform over the quantisation interval related to the number of bits used for intensity encoding.
Since the CRF compresses the sensor response for extreme intensity values, the quantisation noise will dominate in these situations.
Conversely, for correct exposure settings, the quantisation noise is insignificant and a Gaussian sensing process uncertainty provides a good model \cite{hasinoff2014photon}.
Inverting this noise model through the inverse of the CRF function then we expect the covariance $\bar{R}_{\bm{p}}(\tau^k)$ in equation (3) (main paper) to depend on intensity of the pixel: it should be large for extreme intensity values and roughly constant and small for well exposed pixels.

The CRF can be estimated using an image sequence taken under different exposures~\cite{debevec2008recovering,grossberg2003space,robertson2003estimation}.
For long exposures, pixels that would have been correctly exposed become overexposed and provide information on the nonlinearity of the CRF at high intensity, and similarly, short exposures provide information for the low intensity part of the CRF. We have used this approach to estimate the CRF for the APS sensor on a DAVIS event camera and a FLIR camera.
In the experiment, we use the raw image intensity as the measured camera response.

Following \cite{robertson2003estimation}, the exposure time is linearly scaled to obtain the scaled irradiance in the range of raw camera response.
In this way, the camera response function $\text{CRF}(\cdot)$ is experimentally determined as a function of the scaled irradiance $I$.
The Certainty function $f^{c} (\cdot)$ is defined to be the sensitivity of the CRF with respect to the scaled irradiance
\begin{align}
\label{eq:certainty function}
f^{c} & :=
\frac{
	\mathrm{d} \text{CRF}
}{
\mathrm{d} I
},
\end{align}
and it is renormalised so that the maximum is unity \cite{robertson2003estimation}.
Note that different cameras can have dissimilar camera responses for the same irradiance of the sensor.

\begin{figure}[t]
	\centering
	\resizebox{1\columnwidth}{!}{ 
		\begin{tabular}{c c}
			\\
			\includegraphics[width=0.49\linewidth]{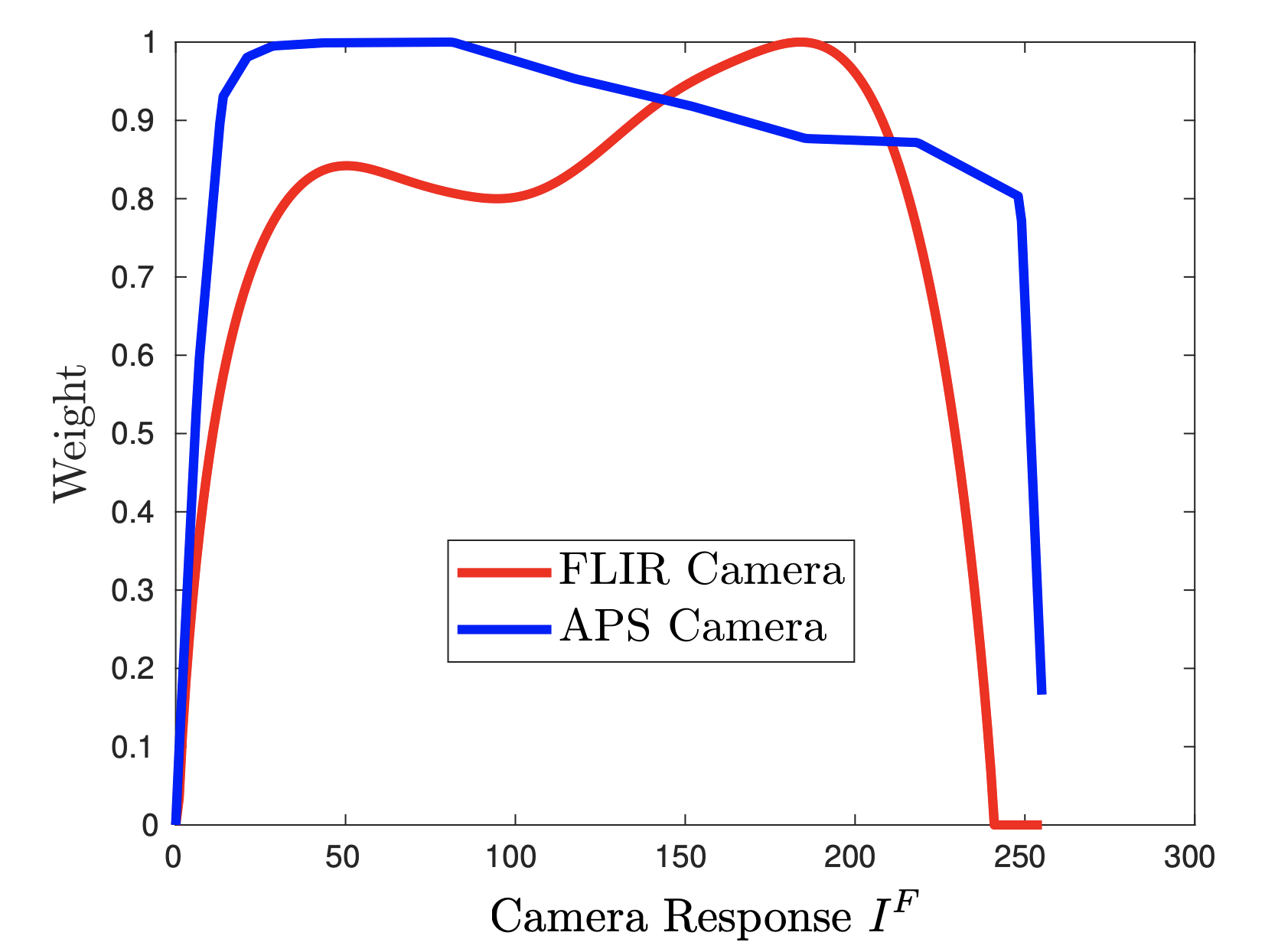} &
			\includegraphics[width=0.49\linewidth]{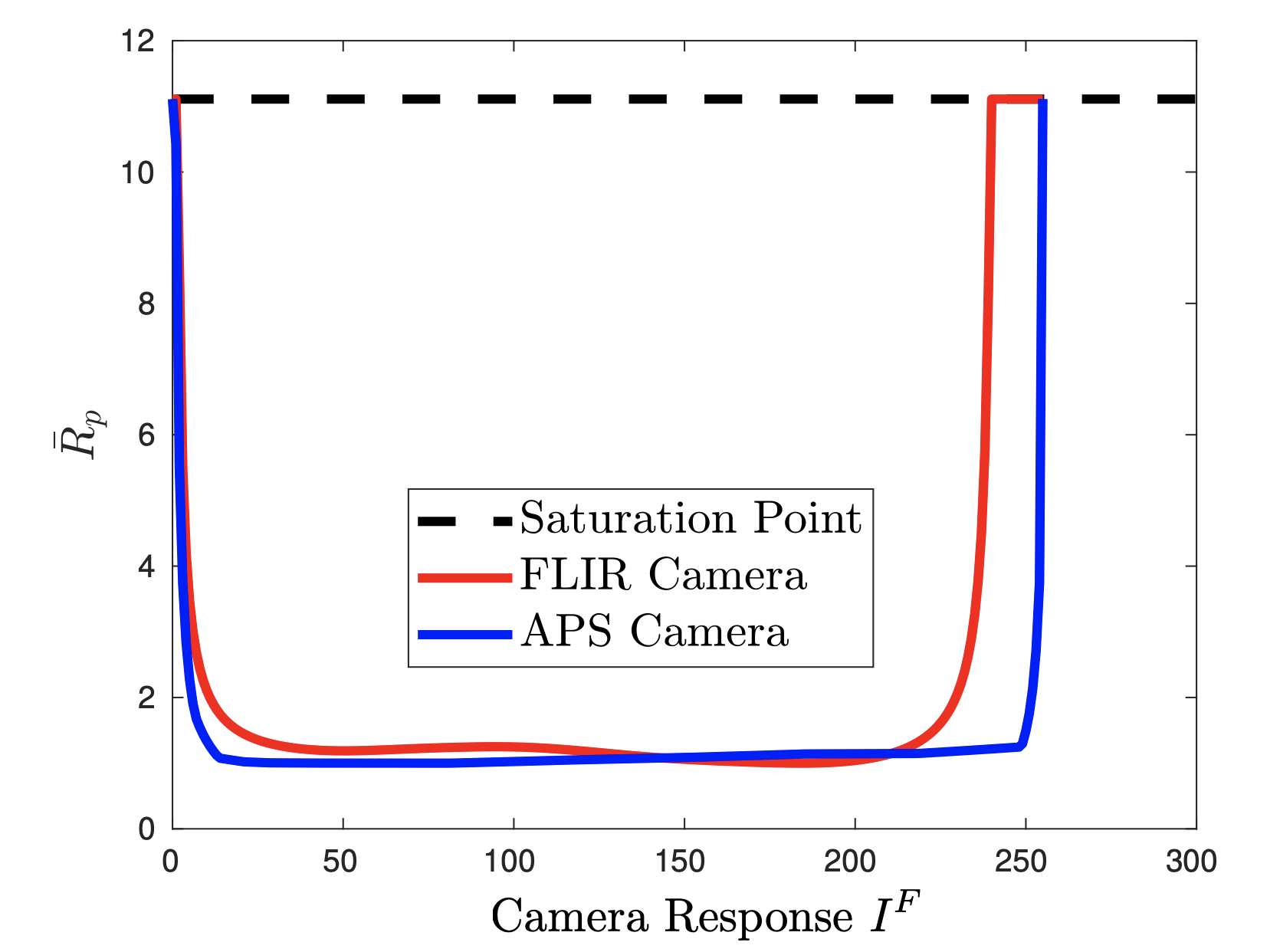}
			\\
			(a) Weighting function  & (b) Image covariance function
			\\
		\end{tabular}
	}
	\caption{\label{fig:crf}
	 Weighting function $f^w(\cdot)$ and image covariance function $\bar{R}_{\bm{p}}$ for the APS camera in a DAVIS \eventframe camera (blue) and the \textit{FLIR} camera (red) used in the experimental studies.
	}\vspace{-2.5mm}
\end{figure}

Remapping the $I$ axis of the Certainty function $f^c(\cdot)$ to camera response $I^F$ defines the Weighting function $f^{w} (\cdot)$ (Fig~\ref{fig:crf}.a) as a function of camera response \cite{robertson2003estimation}
\begin{align}
\label{eq:weight function}
f^{w} & :=
\frac{
	\mathrm{d} \text{CRF}
}{
\mathrm{d} I
} \circ \text{CRF}^{-1},
\end{align}
where $\circ$ defines function composition.

Inspired by \cite{robertson2003estimation}, we define the covariance of noise associated with raw camera response as
\begin{align}
\label{eq:sigma_i}
\bar{R}_{\bm{p}} & :=  {\sigma}_{\text{im.}}^2
\frac{1}{f^{w}(I^F)},
\end{align}
where $\sigma_{\text{im.}}^2$ is a tuning parameter related to the base level of noise in the image
(see Fig. \ref{fig:crf}.b. for $\sigma_{\text{im.}}^2 = 1$).
Note that we also introduce a saturation to assign a maximum value to the image covariance function (Fig.~\ref{fig:crf}.b).
The tuning parameter $\sigma_{\text{im.}}^2$ \eqref{eq:sigma_i} is set to $7\times 10^7$ for the \textit{FLIR} camera and $7\times 10^5$ for the DAVIS240C camera to account for higher relative confidence associated with the intensity value of the \textit{FLIR} camera.

In addition to the base uncertainty model for $I_{\bm{p}}(\tau^{k})$, we will also need to model the uncertainty of frame information in the interframe period and in the log intensity scale for the proposed algorithm.
We use linear interpolation to extend the covariance estimate from two consecutive frames $I_{\bm{p}}(\tau^{k})$ and $I_{\bm{p}}(\tau^{k+1})$ by
\begin{align}
\label{eq: compute R 1}
\bar{R}_{\bm{p}}(t) := \Big(\frac{t-\tau^{k}}{\tau^{k+1} - \tau^{k}}\Big) \bar{R}_{\bm{p}}(\tau^{k+1}) + \Big(\frac{\tau^{k+1}-t}{\tau^{k+1} - \tau^{k}}\Big) \bar{R}_{\bm{p}}(\tau^{k}).
\end{align}

\section{Camera Calibration}
For event cameras, contrast threshold $c$ defines the minimum logarithmic intensity change represented by each event.
In theory, for a certain event camera, contrast threshold of all pixels should be a same value and it is changed by adjusting the event camera sensitivity.
However, the true contrast threshold is difficult to estimate from camera sensitivity, especially for different type of event cameras.
Moreover, due to manufacturing imperfections and circuit noise, contrast threshold may vary among pixels.
Therefore, we calibrate the per-pixel contrast threshold to match the intensity changes between two consecutive frames.
The detailed discussion is shown in the frame augmentation step (Fig.3 main paper).
With matched event and frame data, the intensity jumps and spikes are avoided in the inter-frame high temporal resolution reconstructed videos.
However, due to the calibrated contrast threshold is highly based on intensity frames, the same calibration method fails for state update and deblur on HDR scenarios.
For example, the contrast threshold would always be zero if the bright or dark pixel values in two consecutive frames are saturated.
To avoid limiting the HDR property of events, we estimate the contrast threshold for state update and deblur differently.
We estimate the event sensitively based on the same calibration method between frames but only on pixels without extreme intensity values in frame data.
The median contrast threshold of all pixels in standard dynamic range is chosen.
We experimentally estimate the contrast threshold of our HDR/AHDR dataset to be 0.033, the DESC~\cite{Gehrig21ral} dataset to be 0.05 and all dataset from DAVIS event camera to be 0.1.

\section{Additional Results}

\subsection{Detailed Metric Results:}
It is extremely difficult to measure HDR performance from reproduced images visually, especially after tonemapping or normalisation for print reproduction.
Hence, we provide a detailed metric analysis of HDR performance over state-of-the-art event-based video reconstruction method ECNN~\cite{Stoffregen20eccv}, E2VID \cite{Rebecq20pami}, event-frame video reconstruction method Han \etal~\cite{han2020neuromorphic} with our CF and AKF.
To demonstrate the performance of reconstructing raw scene radiance,
the most relevant metric we provided is the Mean Square Error (MSE).
We also evaluate the structural similarity of the raw intensity using the Structural Similarity Index Measure (SSIM) \cite{Wang04tip} and Q-score \cite{Narwaria15jei} to provide a complete picture of the algorithm performance.

In this section, we provide a detailed result of the summary Table 1 and 2 in the main paper.
The evaluation on each sequence of the HDR dataset (\S\ref{sec: dataset list HDR}) and Artificial HDR dataset (\S\ref{sec: dataset list AHDR}) are summarised in Table \ref{tab:hdr_full}.
The detailed comparison on IJRR dataset \cite{Mueggler17ijrr} (\S\ref{sec: dataset list DAVIS} \#1) is shown in Table~\ref{tab:IJRR_full}.
The quantitative results clearly demonstrate the superior performance of our algorithm CF and AKF versus the compared algorithms ECNN \cite{Stoffregen20eccv}, E2VID \cite{Rebecq20pami} and Han \etal~\cite{han2020neuromorphic}.

\begin{figure*}[h!]
	\vspace{-4mm}
	\centering
	\begin{tabular}{c c c}
		\\
		\includegraphics[width=0.27\linewidth]{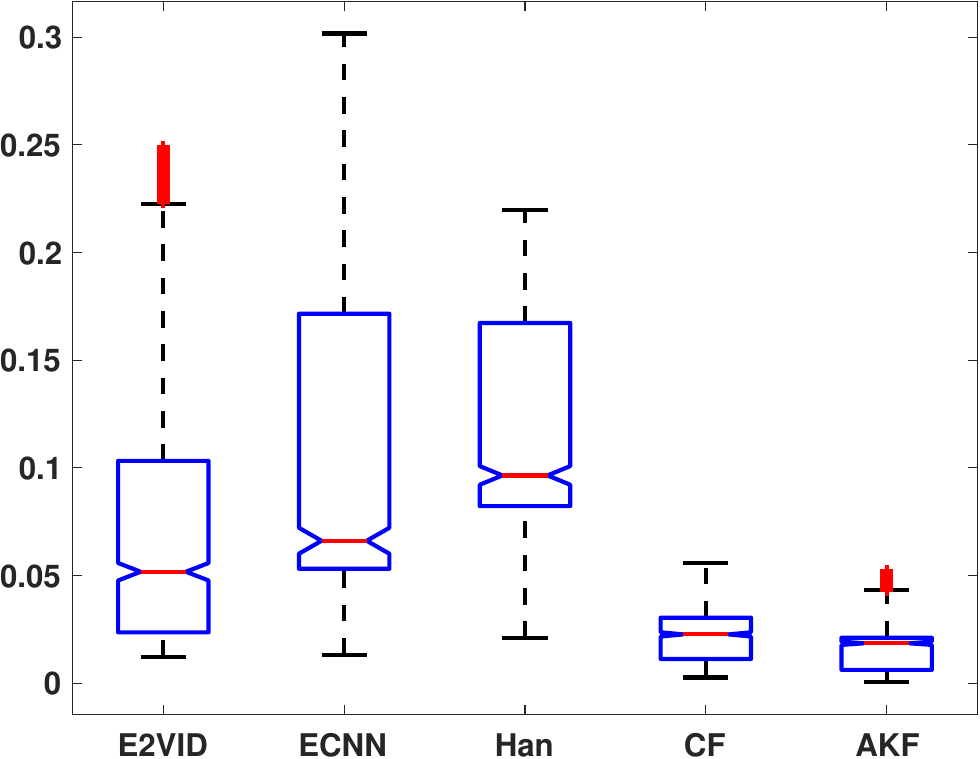} &
		\includegraphics[width=0.27\linewidth]{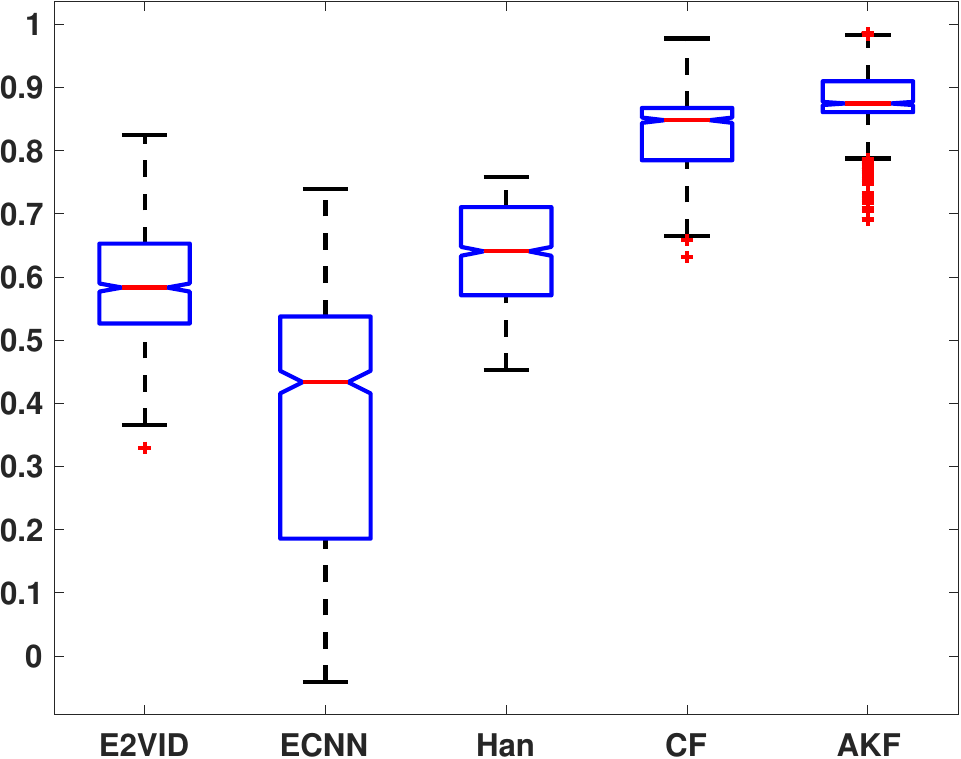} &
		\includegraphics[width=0.27\linewidth]{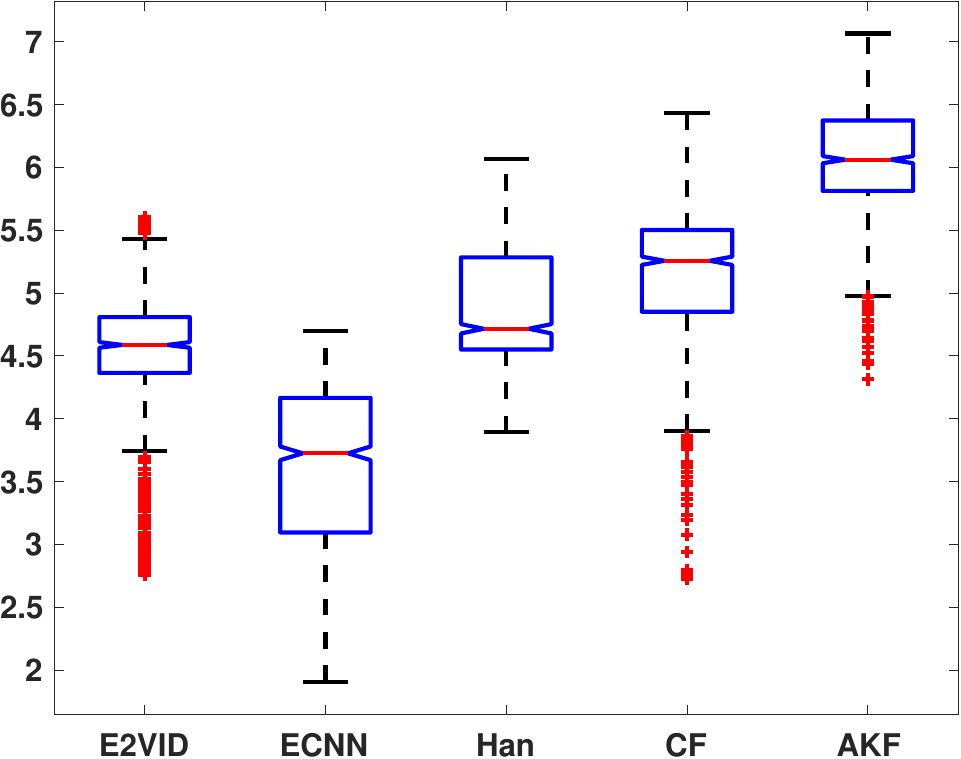} \\ \\
		(a). MSE $\downarrow$ & (b). SSIM $\uparrow$ & (c). Q-score $\uparrow$
	\end{tabular}
	\caption{\label{fig:boxplot}
		Boxplots of raw intensity Mean Square Error (MSE), Structural Similarity Index Measure (SSIM) \cite{Wang04tip} and the HDR Q-score measure \cite{Narwaria15jei}, for ECNN \cite{Stoffregen20eccv}, E2VID \cite{Rebecq20pami}, Han \etal~\cite{han2020neuromorphic} and our algorithm CF and AKF.
		Results are evaluated over the full HDR datasets documented in \S \ref{sec: dataset list HDR}.
	}
\end{figure*}

Fig.~\ref{fig:boxplot} shows a boxplot of the metrics MSE, SSIM and Q-score evaluated on the full HDR dataset (\S\ref{sec: dataset list HDR}) using the reference HDR image as the ground truth.
This provides a clear indication of the relative performance gains of our algorithm, both in \textit{improved average} metric performance and \textit{reduced variance} of the results.

\begin{table*}[t]
	\centering
	\caption{Comparison of state-of-the-art event-based video reconstruction methods E2VID \cite{Rebecq20pami}, ECNN \cite{Stoffregen20eccv} and event-frame video reconstruction method Han \etal~\cite{han2020neuromorphic} with our CF and AKF on the proposed HDR (\S \ref{sec: dataset list HDR}) and AHDR dataset (\S \ref{sec: dataset list AHDR}).
		Our AKF outperforms the compared methods in all scenarios.
		The table expands on the summary results for the HDR-AHDR dataset provided in Table 1 (main paper).
	}
	\label{tab:hdr_full}
	\resizebox{1\textwidth}{!}{ 
		\begin{tabular}{ l | c c c c c | c c c c c | c c c c c }
			\midrule
			\midrule
			Metrics &  \multicolumn{5}{c|}{MSE ($\times 10^{-2}$) $\downarrow$}  &  \multicolumn{5}{c|}{SSIM $\uparrow$}
			&    \multicolumn{4}{c}{Q score $\uparrow$}
			\\
			\midrule
			Methods & E2VID~\cite{Rebecq20pami} & ECNN~\cite{Stoffregen20eccv} & Han~\cite{han2020neuromorphic}  & CF (ours) & AKF (ours) & E2VID~\cite{Rebecq20pami} & ECNN~\cite{Stoffregen20eccv}  & Han~\cite{han2020neuromorphic} & CF (ours) & AKF (ours) & E2VID~\cite{Rebecq20pami} & ECNN~\cite{Stoffregen20eccv} & Han~\cite{han2020neuromorphic} & CF (ours) & AKF (ours)
			\\
			\midrule
			\midrule
			Dataset & \multicolumn{15}{c}{HDR sequences}
			\\
			\midrule

			City
			& $2.05 $ & $1.65 $ & $17.23$ & $0.62 $ & $\textbf{0.25} $
			& $0.60 $ & $0.47 $ & $0.55$ &  $0.78 $ & $\textbf{0.91} $
			& $4.00 $ & $ 2.34$ & $5.58$ & $5.12 $ & $\textbf{6.29} $ \\
			
			Trees 1
			& $2.44 $ & $5.76 $ & $17.34$ & $2.07 $ & $\textbf{1.69} $
			& $0.73 $ & $0.53 $ & $0.55$ & $0.79 $ & $\textbf{0.82} $
			& $4.57 $ & $4.28 $ & $4.42$ & $4.99 $ & $\textbf{6.20} $ \\
			
			Trees 2
			& $9.97 $ & $16.28 $ & $8.33$ & $1.03 $ & $\textbf{0.59} $
			& $0.54 $ & $0.11 $ & $0.71$ & $0.96 $ & $\textbf{0.98} $
			& $4.50 $ & $3.22 $ &  $4.58$ & $5.47 $ & $\textbf{6.17} $ \\
			
			Trees 3
			& $22.50 $ & $26.89 $ &  $9.97$ & $5.00 $ & $\textbf{4.61} $
			& $0.41 $ & $0.15 $ & $0.63$ &  $0.86$ & $\textbf{0.87} $
			& $4.53 $ & $3.19 $ &  $4.63$ & $4.65 $ & $\textbf{5.76} $ \\
			
			Car park 1
			& $10.31 $ & $17.56 $ & $2.31$ &   $2.74 $ & $\textbf{1.89} $
			& $0.56 $ & $0.39 $ &  $0.69$ &  $0.77 $ & $\textbf{0.86} $
			& $4.99 $ & $3.89 $ & $\textbf{5.82}$ &   $5.50 $ & $5.56$ \\
			
			Car park 2
			& $5.08 $ & $5.85 $ & $8.45$ &  $2.80 $ & $\textbf{2.04} $
			& $0.60 $ & $0.56 $ &  $0.72$ &  $0.86 $ & $\textbf{0.91} $
			& $4.59 $ & $4.13 $ & $5.11$ &  $5.26 $ & $\textbf{6.25} $ \\
			
			\midrule
			Mean & $8.73 $ & $12.33 $ & $10.60$ &   $2.37 $  & $\textbf{1.84} $  & $0.58 $ & $0.37 $ &  $0.64$ &  $0.84  $ & $\textbf{0.89} $  & $4.53 $ & $3.51 $  &  $5.02$ & $5.17 $ & $\textbf{6.04} $
			 
			\\
			\midrule
			\midrule
			Dataset & \multicolumn{15}{c}{Artificial HDR sequences (AHDR)}
			\\
			\midrule
			Mountain slow
			& $17.32 $ & $33.00 $ &  $4.31$ & $6.59 $ & $\textbf{6.38} $
			& $0.53 $ & $-0.13 $ &  $0.55$ &  $0.68 $ & $\textbf{0.74} $
			& $5.36 $ & $2.84 $ &  $5.28$ & $6.11 $ & $\textbf{6.60} $ \\
			
			Mountain fast
			& $17.66 $ & $36.54 $ &   $4.34$ & $6.58 $ & $\textbf{6.12} $
			& $0.57 $ & $-0.22 $ &  $0.64$ & $0.66 $ & $\textbf{0.74} $
			& $5.35 $ & $2.77 $ &  $5.33$ & $5.46 $ & $ \textbf{6.24} $ \\

			Lake slow
			& $4.82 $ & $7.97 $ &  $12.06$ & $1.83 $ & $\textbf{1.59} $
			& $0.48 $ & $0.22 $ &  $0.46$ &  $0.79 $ & $\textbf{0.83} $
			& $5.19 $ & $3.82 $ &   $4.82$ & $6.11 $ & $\textbf{6.66} $ \\
			
			Lake fast
			& $6.48 $ & $7.17 $ &  $12.39$ & $2.00 $ & $\textbf{1.66} $
			& $0.43 $ & $0.30 $ &  $0.40$ & $0.78 $ & $\textbf{0.83} $
			& $5.08 $ & $4.00 $ &  $4.62$ & $5.73 $ & $\textbf{6.42} $ \\		
			\midrule
			Mean & $11.57 $ & $21.17 $ &  $8.27$ & $4.25 $  & $\textbf{3.94} $ & $0.50 $ & $0.04 $  & $0.51$ & $0.73  $ & $\textbf{0.79}$  & $5.25 $ & $3.36 $ & $5.01$ &  $5.85 $ & $\textbf{6.48} $
			\\
			\bottomrule[\heavyrulewidth]
		\end{tabular}
	}
\end{table*}


%
\begin{table*}[t!]
	\centering
	\caption{\label{table: detailed IJRR}Comparison of state-of-the art methods of event-based video reconstruction on IJRR~\cite{Mueggler17ijrr} DAVIS datasets (\S \ref{sec: dataset list DAVIS} \#1).
		Though both CF and AKF perform well in the structural similarity metrics SSIM~\cite{Wang04tip} and LPIPS~\cite{Zhang18cvprLPIPS}, our AKF outperforms other methods with a significant margin in the absolute intensity metrics MSE and outperforms the compared methods in the structural similarity metrics SSIM~\cite{Wang04tip} and LPIPS~\cite{Zhang18cvprLPIPS} in most scenarios.
		The table expands on the summary results for the IJRR~\cite{Mueggler17ijrr} DAVIS dataset provided in Table 2 (main paper).
	}
	\label{tab:IJRR_full}
	\resizebox{1\textwidth}{!}{ 
		\begin{tabular}{ l | c c c c | c c c c | c c c c }
			\toprule[\heavyrulewidth]\toprule[\heavyrulewidth]
			Metrics &  \multicolumn{4}{c|}{MSE ($\times 10^{-2}$) $\downarrow$}   &    \multicolumn{4}{c|}{SSIM~\cite{Wang04tip} $\uparrow$}  &    \multicolumn{4}{c}{LPIPS~\cite{Zhang18cvprLPIPS} $\downarrow$}
			\\
			\midrule
			Methods & E2VID~\cite{Rebecq20pami} & ECNN~\cite{Stoffregen20eccv} & CF (ours) & \textbf{AKF (ours)} & E2VID~\cite{Rebecq20pami} & ECNN~\cite{Stoffregen20eccv} & CF (ours) & \textbf{AKF (ours)} & E2VID~\cite{Rebecq20pami} & ECNN~\cite{Stoffregen20eccv} & \textbf{CF (ours)} & \textbf{AKF (ours)}
			\\
			\midrule
			
			boxes\_6dof & $11.87 $ & $3.99 $  & $0.33 $ & $\textbf{0.26} $ & $0.51 $ & $0.62 $  & $0.78 $ & $\textbf{0.80} $ & $0.34 $ & $0.24 $ & $\textbf{0.26}  $ & $\textbf{0.26} $ \\
			
			calibration & $23.46 $ & $3.05 $ & $0.11 $ & $\textbf{0.09} $ & $0.43 $ & $0.65 $  & $\textbf{0.92} $ & $\textbf{0.92} $ & $0.31 $ & $0.18 $ & $0.09  $ & $\textbf{0.07} $ \\
			
			dynamic\_6dof & $30.96 $ & $14.11 $ & $0.13 $ & $\textbf{0.12} $ & $0.24 $ & $0.30 $  & $\textbf{0.88} $ & $0.87 $ & $0.46 $ & $0.36 $ & $\textbf{0.18} $ & $0.20 $ \\
			
			office\_zizang & $17.14 $ & $3.95 $ & $0.26 $ & $\textbf{0.22} $ & $0.40 $ & $0.49 $  & $0.85 $ & $\textbf{0.86}$ & $0.40 $ & $0.25 $  & $\textbf{0.22} $ & $\textbf{0.22} $\\
			
			poster\_6dof & $21.68 $ & $6.86 $  & $0.29 $ & $\textbf{0.26} $ & $0.34 $ & $0.46 $ & $0.78 $ & $\textbf{0.79} $ & $0.35 $ & $0.22 $  & $\textbf{0.22} $ & $0.24 $ \\
			
			shapes\_6dof  & $19.44 $ & $8.77 $  & $0.16 $ & $\textbf{0.11} $ & $0.68 $ & $0.76 $ & $0.92 $ & $\textbf{0.94} $ & $0.31 $ & $0.18 $  & $0.16 $ & $\textbf{0.15}  $\\
			
			slider\_depth  & $20.13 $ & $4.16 $  & $0.21 $ & $\textbf{0.18} $ & $0.44 $ & $0.61 $ & $0.83 $ & $\textbf{0.86} $ & $0.40 $ & $0.23 $  & $0.22 $ & $\textbf{0.21} $\\

			\midrule
			Mean & $20.67 $ & $6.41 $  & $0.21 $ & $\textbf{0.18} $ & $0.43 $ & $0.56 $ & $0.85 $ & $\textbf{0.86} $ & $0.37 $ & $0.24 $  & $\textbf{0.19} $ & $\textbf{0.19} $ \\
			\bottomrule[\heavyrulewidth]
		\end{tabular}
	}
\end{table*}


%
%

\begin{figure*}
	\centering
	\resizebox{1\textwidth}{!}{
		\begin{tabular}{
				>{\centering\arraybackslash}m{1mm}
				>{\centering\arraybackslash}m{6cm}
				>{\centering\arraybackslash}m{6cm}
				>{\centering\arraybackslash}m{6cm}
				>{\centering\arraybackslash}m{6cm}
>{\centering\arraybackslash}m{6cm}
				>{\centering\arraybackslash}m{6cm}}
			\rotatebox{90}{\Large \texttt{(a) Night\_drive}}
			&
			\includegraphics[width=\linewidth]{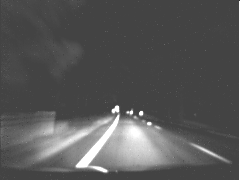}
			&
			\includegraphics[width=\linewidth]{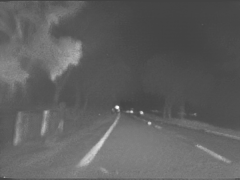}
			&
			\includegraphics[width=\linewidth]{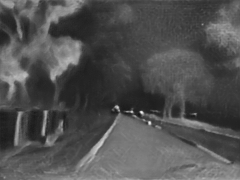}
			&
			\includegraphics[width=\linewidth]{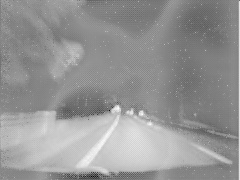}
			&
			\includegraphics[width=\linewidth]{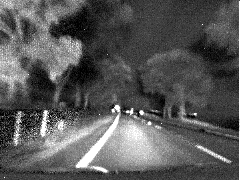}
			&
			\includegraphics[width=\linewidth]{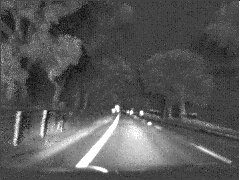}\\
			\rotatebox{90}{\Large \texttt{(b) Boxes\_6dof}}
			&
			\includegraphics[width=\linewidth]{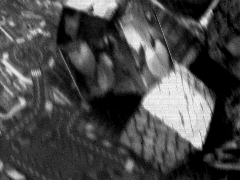}
			&
			\includegraphics[width=\linewidth]{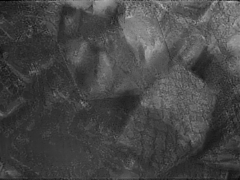}
			&
			\includegraphics[width=\linewidth]{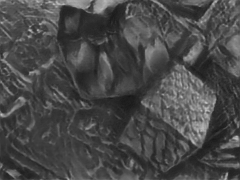}
			&
			\includegraphics[width=\linewidth]{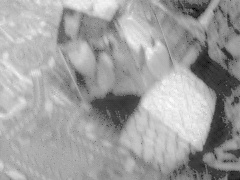}
			&
			\includegraphics[width=\linewidth]{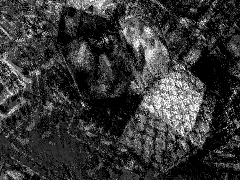}
			&
			\includegraphics[width=\linewidth]{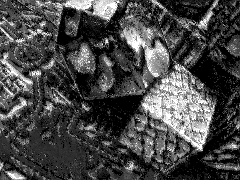}\\
			\rotatebox{90}{\Large \texttt{(c) City\_09b}}
			&
			\includegraphics[width=1\linewidth]{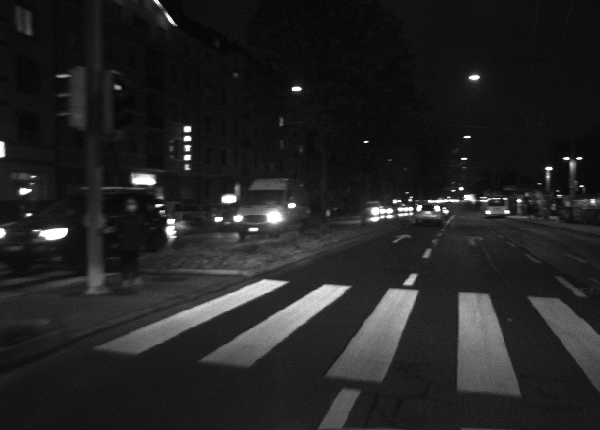}
			&
			\includegraphics[width=1\linewidth]{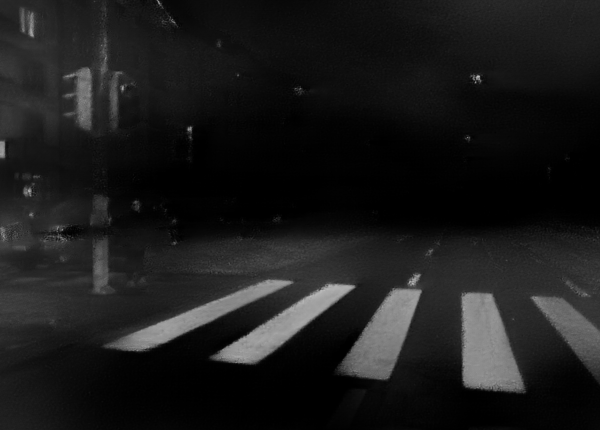}
			&
			\includegraphics[width=1\linewidth]{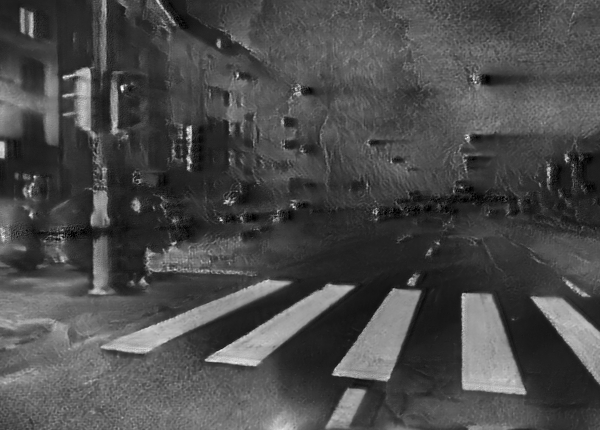}
			&
			\includegraphics[width=1\linewidth]{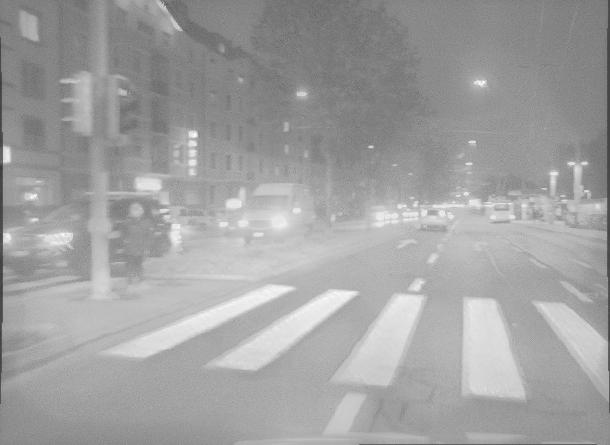}
			&
			\includegraphics[width=1\linewidth]{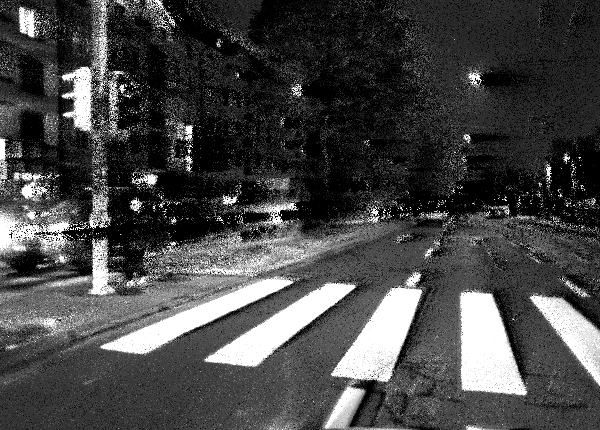}
			&
			\includegraphics[width=1\linewidth]{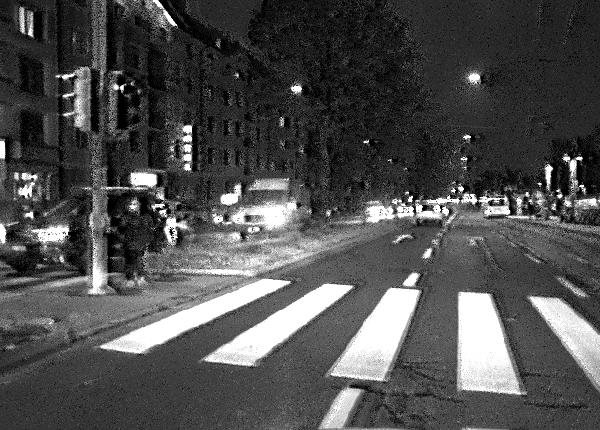}
			\\
			\rotatebox{90}{\Large \texttt{(d) City\_09d}}
			&
			\includegraphics[width=1\linewidth]{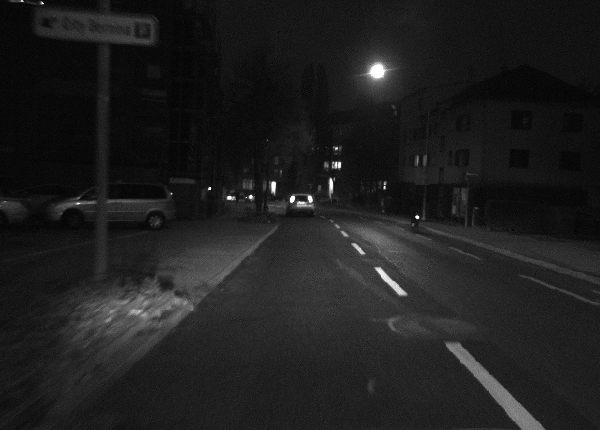}
			&
			\includegraphics[width=1\linewidth]{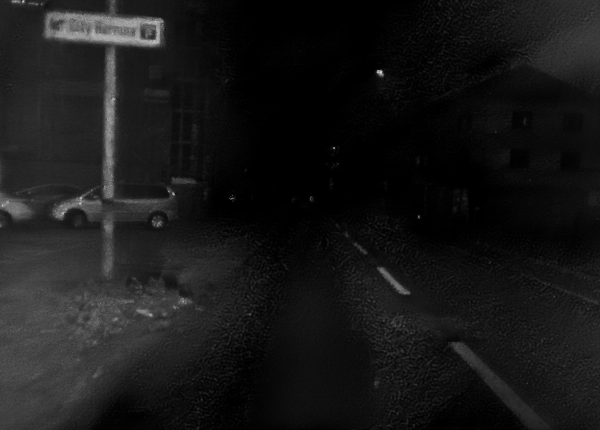}
			&
			\includegraphics[width=1\linewidth]{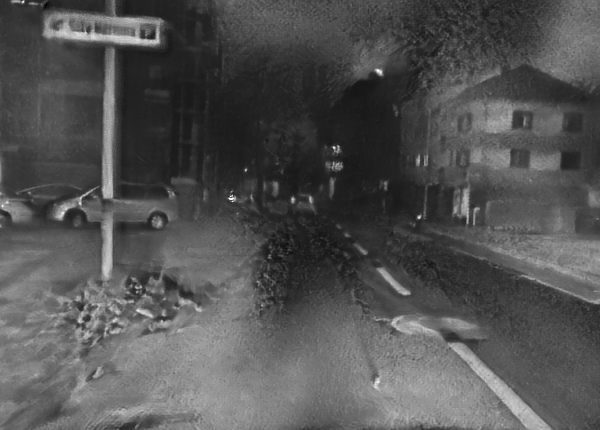}
			&
			\includegraphics[width=1\linewidth]{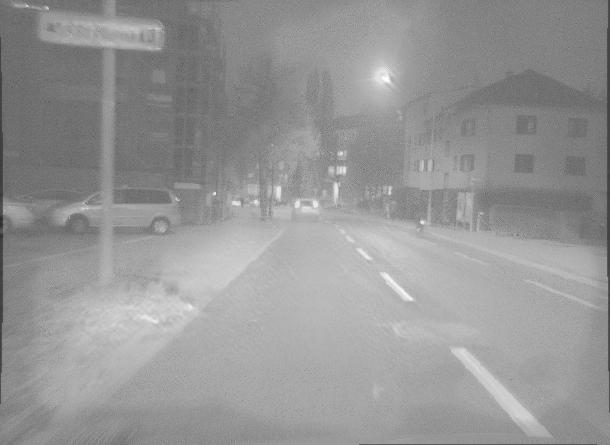}
			&
			\includegraphics[width=1\linewidth]{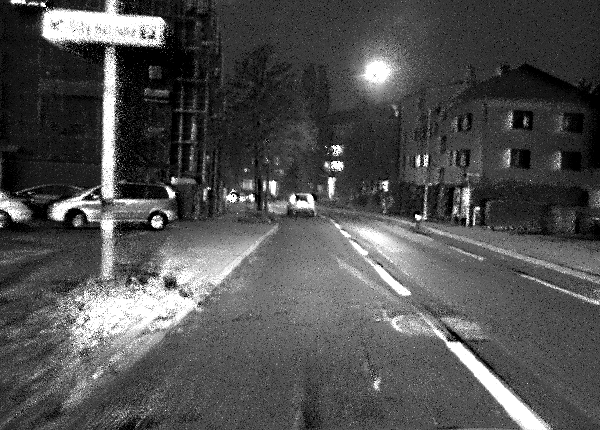}
			&
			\includegraphics[width=1\linewidth]{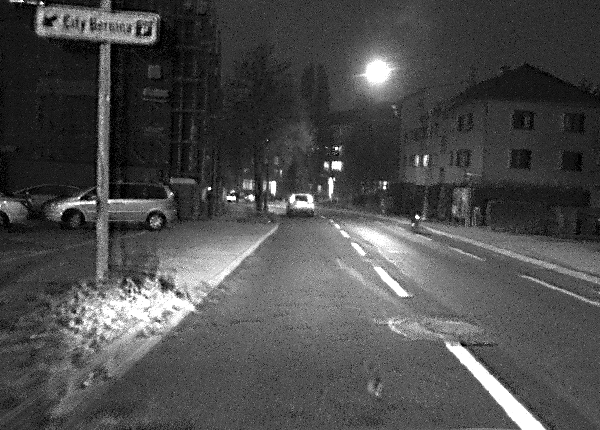}
			\\
			\rotatebox{90}{\Large \texttt{(e) City\_09e}}
			&
			\includegraphics[width=1\linewidth]{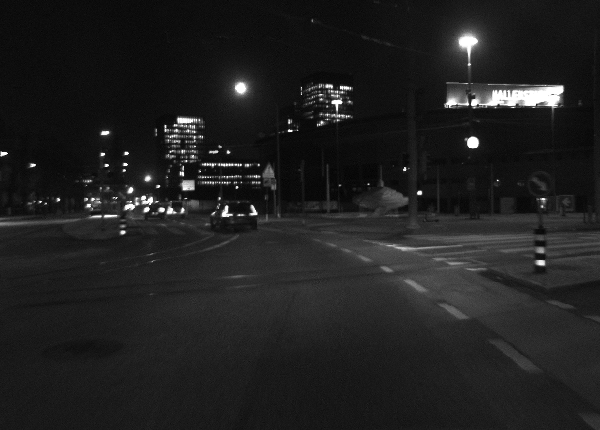}
			&
			\includegraphics[width=1\linewidth]{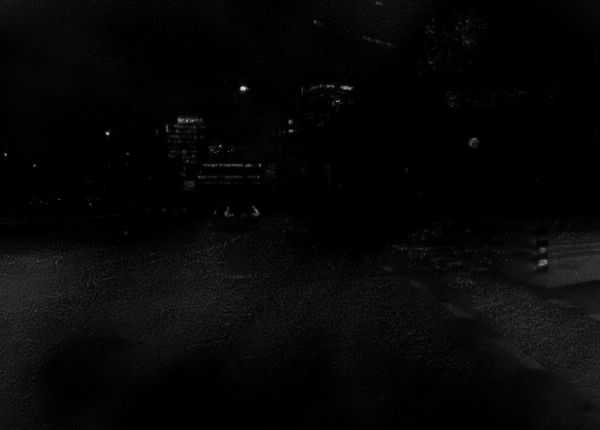}
			&
			\includegraphics[width=1\linewidth]{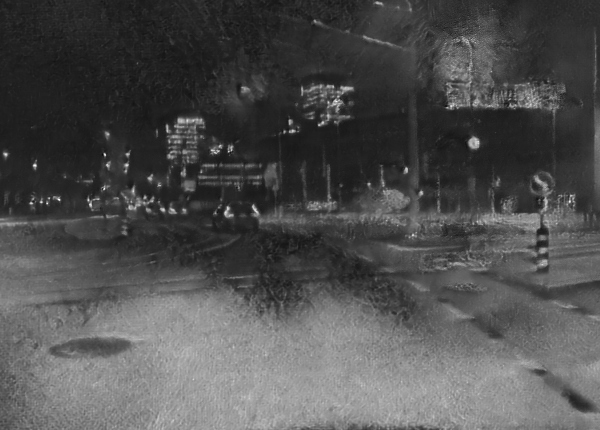}
			&
			\includegraphics[width=1\linewidth]{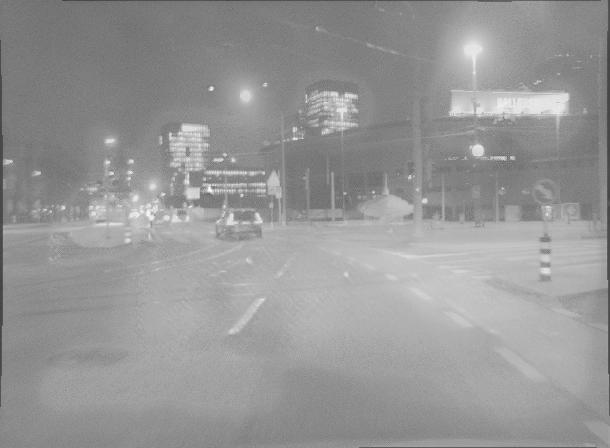}
			&
			\includegraphics[width=1\linewidth]{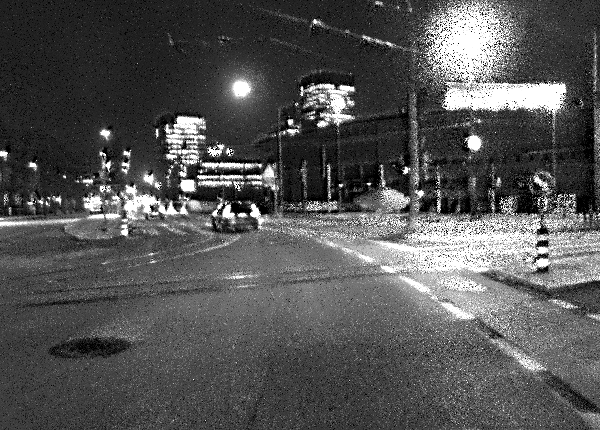}
			&
			\includegraphics[width=1\linewidth]{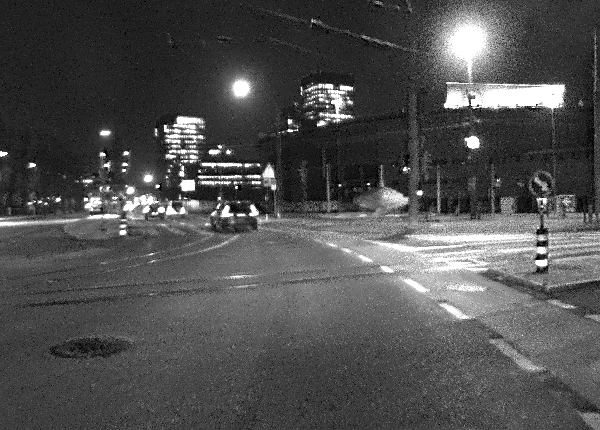}
			\\
			\\
			&\Large DAVIS frame
			&\Large E2VID \cite{Rebecq20pami}
			&\Large ECNN \cite{Stoffregen20eccv}
			&\Large Han \etal~\cite{han2020neuromorphic}
			&\Large CF (ours)
			&\Large \textbf{AKF (ours)}
		\end{tabular}
	}
	\vspace{0.2cm}
	\caption{ 
		Additional results on
		comparison of state-of-the-art event-based video reconstruction methods on sequences with challenging lighting conditions and fast motions, drawn from the open-source datasets ACD~\cite{Scheerlinck18accv}, IJRR~\cite{Mueggler17ijrr} and DSEC~\cite{Gehrig21ral}.
		We demonstrate the different time instances of the same dataset in the main paper.
		E2VID \cite{Rebecq20pami} fails to recover trees on the right-hand side in (a) and details on the
		carpet in (b).
		It fails on DSEC~\cite{Gehrig21ral} dataset by generating black reconstruction.
		The method provides washed-out reconstructions that fail to capture the true image intensity.
		ECNN~\cite{Stoffregen20eccv} is able to reconstruct objects under extreme lighting conditions but it is very sensitive to event noise, leading to some unrealistic artefacts on road in (a), carpet textures in (b), and `dirty mist' in (c)-(e).
		Han \etal~\cite{han2020neuromorphic}, CF and AKF achieve good HDR reconstruction by taking advantage of both event and frame data. Han \etal~\cite{han2020neuromorphic} and 
		CF results suffer from blurry, such as poles in (a) and boxes in (b).
		Our AKF provides sharper, cleaner and higher dynamic reconstructions under challenging scenarios, such as box textures in (b), cars in (c), the road sign in (d).
		Video comparisons are provided in the supplementary video materials.
	}
	\label{fig:davis_supp}
\end{figure*}

\subsection{Visual Analysis of HDR Performance:}
In the main paper, we have visually evaluated the algorithms on a set of public DAVIS event camera datasets that highlight specific issues in the HDR reconstruction.
\texttt{Night\_drive} in ACD~\cite{Scheerlinck18accv} Dataset \#2, \texttt{Outdoors\_running} and \texttt{Boxes\_6dof} in IJRR~\cite{Mueggler17ijrr} Dataset \#1 and a hybrid event-frame dataset DSEC~\cite{Gehrig21ral} in Dataset \#4 of \S \ref{sec: dataset list DAVIS}.
The evaluated datasets include a range of challenging scenarios such as night/bright outdoor lighting conditions, driving, running, fast/slow motions, static/moving background, dynamic/static objects, highly reflective/high contrast environment.
In Fig.~\ref{fig:davis_supp}, we demonstrate more examples on sequences \texttt{Night\_drive}, \texttt{Boxes\_6dof} and three different sequences in
DSEC~\cite{Gehrig21ral} dataset at different time instances of the same dataset in the main paper.

In Fig.~\ref{fig:davis_supp} (a), the DAVIS frames only capture blurry roadside poles and a small part of the background trees.
E2VID~\cite{Rebecq20pami} produces washed-out reconstructions of the near trees and the roadside poles.
The background trees on the right-hand side can not be captured properly.
The other learning-based method ECNN~\cite{Stoffregen20eccv} is able to capture more details of the night driving scene but still leads to unrealistic textures on road.
Han \etal~\cite{han2020neuromorphic} fuses events and frames. 
They use pure events to reconstruct intensity images using a known algorithm~\cite{Rebecq20pami}, and fuse this HDR information with LDR absolute intensity images.	
The output images are highly affected by the blurry input LDR images, such as the poles and trees.
The output reconstructions lose the absolute brightness information of LDR images, generating `washed-out' images with inaccurate intensities.
The network also introduces noisy artefacts which affects the overall quality.
By fusing DAVIS frames, our CF captures more detailed background trees and provides more accurate intensities.
However, the reconstructions of CF include `hot pixels' (constantly fired pixels) and
noisy shadows trailing behind the fast-moving objects (obvious in our supplementary video).
Also, the trade-off between fusing frames and event data of the CF algorithm is fixed by a predefined constant gain in the filter which limits the performance.
Our AKF overcomes the limitations of the simple CF model by dynamically adjusting the Kalman gain based on the event and frame uncertainty.
AKF encodes the sensor uncertainties in both frames and event data.
It allows the filter to exploit the full information in the event stream where the image frame information is not reliable (\eg underexposed or overexposed), while still exploiting the image frame information where it is reliable, \eg the visible trees, the roadside poles, and the road markings.
For CF, the constant triggered `hot pixels' affect the reconstruction performance,
Therefore, AKF relies more on frame data or filter state on these pixels so that is not affected by the `hot pixels' in the \texttt{Night\_drive} sequence.

In Fig.~\ref{fig:davis_supp} (b), the fast camera motion causes blurry DAVIS frames and the lack of textures on boxes and the carpet.
The batch method E2VID~\cite{Rebecq20pami} and ECNN~\cite{Stoffregen20eccv} accumulate events within a certain time period before generating a frame reconstruction from the event batch.
The reconstruction quality is highly dependent on the quantity and quality of the corresponding event data.
When the camera is moving quickly in a highly textured scene, events will not be triggered fast enough due to the refractory period~\cite{yang2015dynamic}, so the event quality is degraded.
This leads to the poor reconstruction for E2VID~\cite{Rebecq20pami} and ECNN~\cite{Stoffregen20eccv}, while our AKF models this type of noise (main paper \S 3.1.1) and managed to produce better image reconstruction.
Han \etal~\cite{han2020neuromorphic} use both events and frames as input. 
The overall structure of over- and under-exposed regions is well reconstructed, but the algorithm does not generate images with realistic intensity value levels and does not tackle motion blur.
Our CF fuses the image frame data and consequently remains faithful to the true intensity values in the image. However, the simple zero-order hold assumption (fusing event data with the previous frame data) cannot account for fast camera motion.
This leads to obvious `double edges' where image frame data from previous frames are not fully compensated in the reconstruction.
In comparison, our AKF provides sharp and clean reconstruction in challenging scenarios with high dynamic range and fast motions.

In Fig.~\ref{fig:davis_supp} (c)-(e), we provide additional examples on each sequence of the DSEC~\cite{Gehrig21ral} data presented in the main paper.
E2VID~\cite{Rebecq20pami} generates very smooth but mostly black reconstructions.
ECNN~\cite{Stoffregen20eccv} outperforms E2VID~\cite{Rebecq20pami} in this dataset,
because ECNN~\cite{Stoffregen20eccv} is able to recover textures of the background buildings, road sign in \texttt{city\_09d} and \texttt{city\_09e}.
However, ECNN~\cite{Stoffregen20eccv} is very sensitive to event noise, leading to some unrealistic artefacts like `dirty mist', such as the undesired noise on the background of dataset \texttt{city\_09b} and the upper right corner of dataset \texttt{city\_09d}.
The intensity values of these reconstructed pixels are too high to match the correct dynamic range of the overall image.
By exploiting the complementary characteristics of frames and events, Han \etal~\cite{han2020neuromorphic}, CF and AKF fuse the two sensor modalities to achieve much cleaner reconstructions.
The dark buildings are well recovered in dataset \texttt{city\_09d} and \texttt{city\_09e}.
However, absolute brightness information is not correct in Han \etal~\cite{han2020neuromorphic}. The reconstructions of Han \etal~\cite{han2020neuromorphic} and CF are blurry, such as the road sign in \texttt{city\_09d}.
Also, CF still suffers from `shadows' trailing behind moving objects. For instance, the road sign in \texttt{city\_09d} is ruined by bright `shadows'.
The HDR performance is also limited, such as the details of the dark buildings in \texttt{city\_09b} are not well reconstructed.
AKF outperforms CF because it recovers sharper and clearer objects in dark, such as cars, traffic lights and passengers in the lower-left corner in \texttt{city\_09b}, clear road sign and roadside grass in \texttt{city\_09d}, less shadow of the white dividing lines and pedestrian crossing in \texttt{city\_09e}.
Please refer to our supplementary video for more comparisons.

\section{Color Event Camera Datasets}
\begin{figure}[h!]
	\centering
	\begin{tabular}{c c}
		\\
		\includegraphics[width=0.46\linewidth]{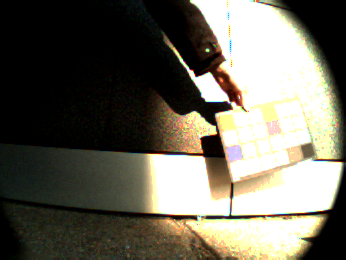} &
		\includegraphics[width=0.46\linewidth]{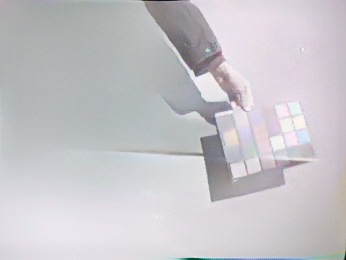} \\
		(a) CED Raw Image  & (b) E2VID \\
		\includegraphics[width=0.46\linewidth]{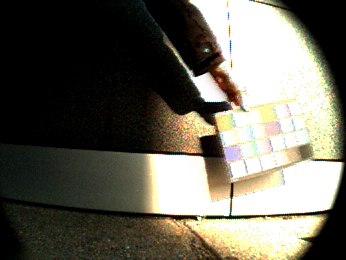} &
		\includegraphics[width=0.46\linewidth]{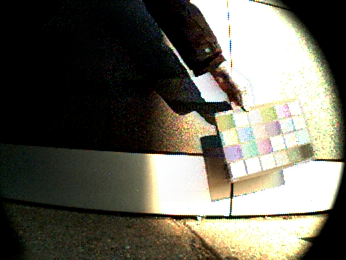} \vspace{-1mm}
		\\ (c) CF (ours) & (d) AKF (ours)
	\end{tabular}
		\caption{\label{fig: CED}
		Image reconstruction evaluation on color event camera CED~\cite{scheerlinck2019ced}.
		}
\end{figure}

Reconstructing colored video from event data is a possible future research direction when color frame-event data is readily available.
Presently, the only color event camera dataset is the CED~\cite{scheerlinck2019ced} dataset.
We evaluate the algorithms on the HDR sequence draw from the dataset in Figure~\ref{fig: CED}.
Our CF and AKF process events in a pixel-wised manner which preserves the Bayer pattern well for color reconstruction.
AKF outperforms CF in the highly reflective board area because in these pixels, the dynamically adjusted filter gain relies less on the saturated frame data but filter state and HDR events.
Without using frame information, E2VID~\cite{Rebecq20pami} fails to compute the static background of \texttt{Shadow}, and only provides washed-out reconstructions.
In the main paper, we show preliminary results (Fig. 6 main paper) by applying chrominance from RGB frame data to color HDR greyscale reconstructions.

\section{Derivation of $\hat{L}_{\vb*{p}}(t)$} \label{section: AKF ODE}
Within a time-interval $t \in [t^{i}_{\bm{p}},t^{i+1}_{\bm{p}})$, the ODE of $\hat{L}_{\bm{p}}(t)$ is
\begin{align*}
&\dot{\hat{L}}_{\bm{p}}(t) = - \frac{R_{\bm{p}}^{-1}(t) \cdot [\hat{L}_{\bm{p}}(t) - L_{\bm{p}}^F(t)]}{P_{\bm{p}}^{-1}(t^{i}_{\bm{p}}) + R_{\bm{p}}^{-1}(t) \cdot  (t - t^{i}_{\bm{p}})}, \\
&\frac{\mathrm{d} [\hat{L}_{\bm{p}}(t) - L_{\bm{p}}^F(t)]}{\hat{L}_{\bm{p}}(t) - L_{\bm{p}}^F(t)} = -\frac{R_{\bm{p}}^{-1}(t)}{P_{\bm{p}}^{-1}(t^{i}_{\bm{p}}) + R_{\bm{p}}^{-1}(t) \cdot (t - t^{i}_{\bm{p}})} \mathrm{d} t.
\end{align*}
Integrate from the event time $t^{i}_{\bm{p}}$ to time $t$
\begin{align*}
&\ln (\hat{L}_{\bm{p}}(t) - L_{\bm{p}}^F(t)) \\
& = \int_{t^{i}_{\bm{p}}}^{t} -\frac{R_{\bm{p}}^{-1}(t)}{P_{\bm{p}}^{-1}(t^{i}_{\bm{p}}) + R_{\bm{p}}^{-1}(t) \cdot (\gamma - t^{i}_{\bm{p}})} \mathrm{d} \gamma\\
&= -\ln (P_{\bm{p}}^{-1}(t^{i}_{\bm{p}}) + R_{\bm{p}}^{-1}(t) \cdot (t - t^{i}_{\bm{p}})) + C_2,
\end{align*}
where $C_2$ is a constant number. Take exponential of both sides
\begin{align*}
&\hat{L}_{\bm{p}}(t) - L_{\bm{p}}^F(t) = \frac{1}{P_{\bm{p}}^{-1}(t^{i}_{\bm{p}}) + R_{\bm{p}}^{-1}(t) \cdot  (t - t^{i}_{\bm{p}})} \cdot e^{C_2},
\end{align*}
and we define $C_3 = e^{C_2}$.\\
Let $t = t^{i}_{\bm{p}}$ and we have
\begin{align*}
&\hat{L}_{\bm{p}}(t^{i}_{\bm{p}}) - L_{\bm{p}}^F(t^{i}_{\bm{p}}) = \frac{1}{P_{\bm{p}}^{-1}(t^{i}_{\bm{p}}) } \cdot  C_3, \\
&C_3 =  [\hat{L}_{\bm{p}}(t^{i}_{\bm{p}}) - L_{\bm{p}}^F(t^{i}_{\bm{p}})] \cdot P_{\bm{p}}^{-1}(t^{i}_{\bm{p}}).
\end{align*}
The solution of the ODE of $\hat{L}_{\bm{p}}(t)$ is
\begin{align*}
\hat{L}_{\bm{p}}(t) = \frac{ [\hat{L}_{\bm{p}}(t^{i}_{\bm{p}}) - L_{\bm{p}}^F(t^{i}_{\bm{p}})] \cdot P_{\bm{p}}^{-1}(t^{i}_{\bm{p}})}{P_{\bm{p}}^{-1}(t^{i}_{\bm{p}}) + R_{\bm{p}}^{-1}(t) \cdot  (t - t^{i}_{\bm{p}})} + L_{\bm{p}}^F(t).
\end{align*}

\section{Derivation of $P_{\vb*{p}}(t)$}
Within a time-interval $t \in [t^{i}_{\bm{p}},t^{i+1}_{\bm{p}})$, the ODE of $P_{\bm{p}}(t)$ is
\begin{align*}
\frac{1}{P_{\bm{p}}^2(t)} \dv{P_{\bm{p}}(t)}{t} = -R_{\bm{p}}^{-1}(t).
\end{align*}
Moving d$t$ to the right hand side yields
\begin{align*}
\frac{1}{P_{\bm{p}}^2(t)} \mathrm{d}P_{\bm{p}}(t) = -R_{\bm{p}}^{-1}(t) \mathrm{d}t.
\end{align*}
Integrate from event time $t^{i}_{\bm{p}}$ to time $t$
\begin{align*}
&\int_{t^{i}_{\bm{p}}}^{t} \frac{1}{P_{\bm{p}}^2(t)} \mathrm{d}P_{\bm{p}}(t) = \int_{t^{i}_{\bm{p}}}^{t}-R_{\bm{p}}^{-1}(t) \mathrm{d} \gamma,\\
&-P_{\bm{p}}^{-1}(t) = -R_{\bm{p}}^{-1}(t) \cdot  (t - t^{i}_{\bm{p}}) + C_1, \\
&P_{\bm{p}}(t) = \frac{1}{R_{\bm{p}}^{-1}(t) \cdot  (t - t^{i}_{\bm{p}}) - C_1 },
\end{align*}
where $C_1$ is a constant number. \\
Let $t = t^{i}_{\bm{p}}$ and we have
\begin{align*}
&{P(t^{i}_{\bm{p}})} = \frac{1}{-C_1}, \\
&C_1=  -{P_{\bm{p}}^{-1}(t^{i}_{\bm{p}})}.
\end{align*}
The solution of the ODE of $P_{\bm{p}}(t)$ is
\begin{align*}
P_{\bm{p}}(t) = \frac{1}{ P_{\bm{p}}^{-1}(t^{i}_{\bm{p}}) + R_{\bm{p}}^{-1}(t) \cdot  (t - t^{i}_{\bm{p}})}.
\end{align*}

\begin{table*}
	\centering
	\caption{\label{tab:other_datasets}Open-source Event Camera Dataset Used in the Paper for Qualitative Evaluation of HDR Reconstruction Performance.}
	\resizebox{1\textwidth}{!}{ 
		\begin{tabular}{ c | c | c | c | c | c}
			\midrule
			\midrule
			HDR Dataset & \# of images &  Speed & Description & HDR Scene & Reference Image
			\\
			\midrule
			\midrule
			\textbf{Dataset \#1} & \multicolumn{4}{c}{\textbf{IJRR dataset}~\cite{Mueggler17ijrr}, DAVIS 240C event camera}
			\\
			\midrule
			boxes\_6dof & 1298 & increasing speed &  highly textured environment &  boxes and carpet & \xmark
			\\
			\midrule
			outdoors\_running & 1573 & running speed & sunny urban environment & buildings & \xmark
			\\
			\midrule
			calibration & 1422 & slow & checkerboard (6x7, 70mm) & - & \xmark
			\\
			\midrule
			dynamic\_6dof & 1267 & increasing speed & office with moving person & - & \xmark
			\\
			\midrule
			office\_zizang & 248 & slow & office environment & - & \xmark
			\\
			\midrule
			poster\_6dof & 1358 & increasing speed & wallposter & - & \xmark
			\\
			\midrule
			shapes\_6dof & 1356 & increasing speed & simple shapes on a wall & - & \xmark
			\\
			\midrule
			slider\_depth & 87 & constant speed & objects at different depths & - & \xmark
			\\
			\midrule
			\midrule
			\textbf{Dataset \#2} & \multicolumn{4}{c}{\textbf{ACD dataset}~\cite{Scheerlinck18accv}, DAVIS 240C event camera}
			\\
			\midrule
			night\_drive & 1058 & high speed & low-light driving &  roadside signs, poles and trees & \xmark
			\\
			\midrule
			\midrule
			\textbf{Dataset \#3} & \multicolumn{4}{c}{\textbf{CED dataset}~\cite{scheerlinck2019ced}, DAVIS 346 color event camera}
			\\
			\midrule
			shadow & 576 & slow & static background, moving board &  board & \xmark  \\
			\midrule
			\midrule
			\textbf{Dataset \#4} & \multicolumn{4}{c}{\textbf{DSEC dataset}~\cite{Gehrig21ral}, stereo-hybrid Prophesee event camera}
			\\
			\midrule
			Interlaken 00b & 1617 & driving speed & driving in daylight & tunnel & \xmark
			\\
			\midrule
			Interlaken 01a & 2263 & driving speed & driving at dusk  & highly reflective road & \xmark
			\\
			\midrule
			City 09 b & 367 & driving speed & city night driving view & dark buildings, trees, cars and people &
			\xmark
			\\
			\midrule
			City 09 d & 1699 & driving speed & city night driving view & dark buildings, trees, cars and people &
			\xmark
			\\
			\midrule
			City 09 e & 817 & driving speed & city night driving view & dark buildings, trees, cars and people &
			\xmark
			\\
			\midrule
			\midrule
		\end{tabular}
	}
\end{table*}

\section{Dataset List}\label{sec: dataset list}
In this section, we list all datasets we used in the paper and supplementary material.
We evaluate the compared methods on some challenging sequences from the popular open-source DAVIS event camera datasets and stereo hybrid event-frame dataset.
The datasets developed in this paper are the HDR (\S\ref{sec: dataset list HDR}) and AHDR (\S\ref{sec: dataset list AHDR}) dataset with HDR references.

\subsection{Evaluated Open-source Event Camera Dataset}\label{sec: dataset list DAVIS}
For completeness, we selected a number of challenging sequences (with HDR scene, driving, fast motion or dynamic objects) from the existing open-source datasets for evaluation.
The evaluated datasets include the popularly DAVIS event camera IJRR datasets~\cite{Mueggler17ijrr}, HDR sequences drawn from ACD datasets~\cite{Scheerlinck18accv} and CED~\cite{scheerlinck2019ced}, the newest stereo-hybrid Prophesee event camera DSEC dataset~\cite{Gehrig21ral}, with details as shown in Table~\ref{tab:other_datasets}.

\begin{table*}
	\caption{\label{tab:HDR}Details of Our HDR Hybrid Event-Frame Dataset}
	\resizebox{1\textwidth}{!}{ 
		\begin{tabular}{ c | c | c | c | c | c}
			\midrule
			\midrule
			HDR Dataset & \# of images &  Speed & Description & HDR Scene & Reference Image
			\\
			\midrule
			\midrule
			\textbf{Dataset} & \multicolumn{4}{c}{\textbf{HDR hybrid event-frame sequences}}
			\\
			\midrule
			city & 150 & medium & rooftop overlooking city buildings & dark buildings & \checkmark
			\\
			\midrule
			trees 1 & 210 & slow and fast & a car under tree shadow & tree shadow & \checkmark
			\\
			\midrule
			trees 2 & 150 & medium & parking lot, buildings and clouds & bright far-away buildings & \checkmark
			\\
			\midrule
			trees 3 & 150 & medium & trees partially covered by shadow and far away buildings & buildings and tree shadows & \checkmark
			\\
			
			\midrule
			car park 1 & 200 & medium & clouds, cars and buildings & bright sky and road surface & \checkmark
			\\
			\midrule
			car park 2 & 200 & medium & clouds, cars, buses and buildings & bright sky, road surface and cars & \checkmark
			\\
			\midrule
			\midrule
		\end{tabular}
	}
\end{table*}

\vspace{0.3cm}
\begin{figure*}
	\centering
	\resizebox{1\linewidth}{!}{ 
		\begin{tabular}{c c c c c c}
			\\
			\includegraphics[width=1\linewidth]{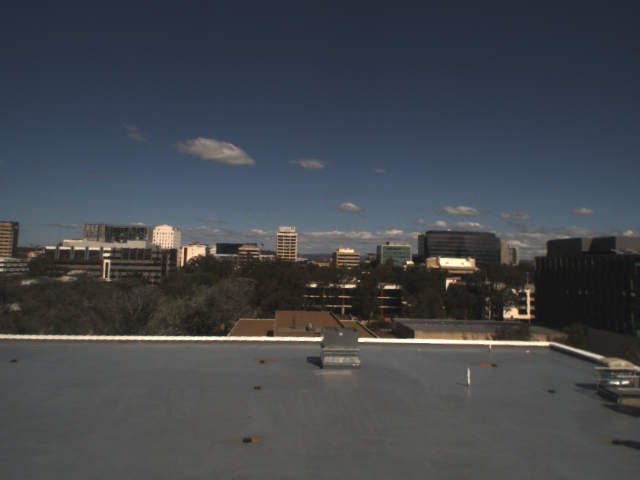} &
			\includegraphics[width=1\linewidth]{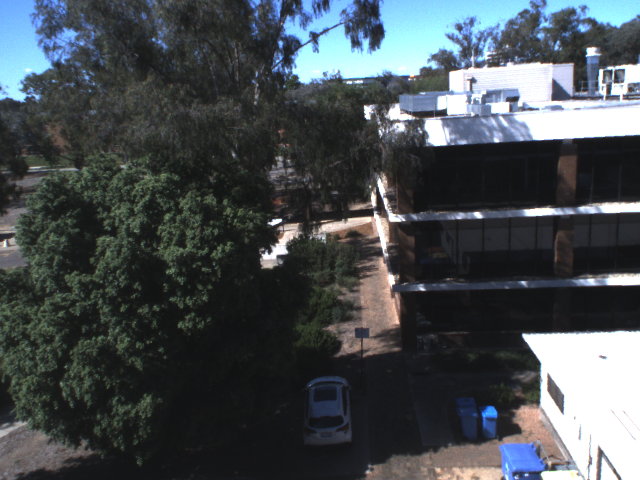} &
			\includegraphics[width=1\linewidth]{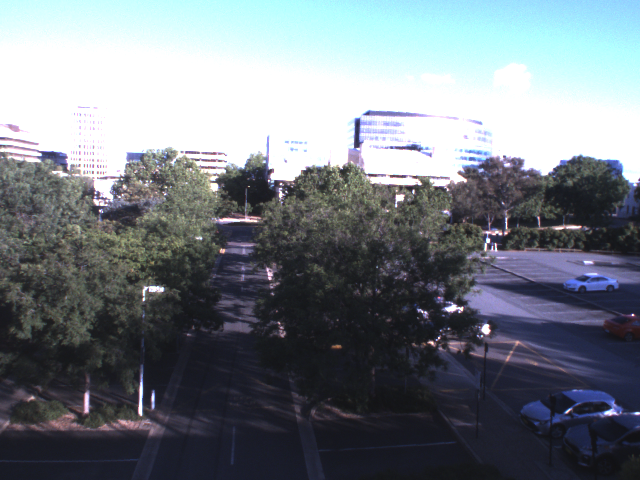} &
			\includegraphics[width=1\linewidth]{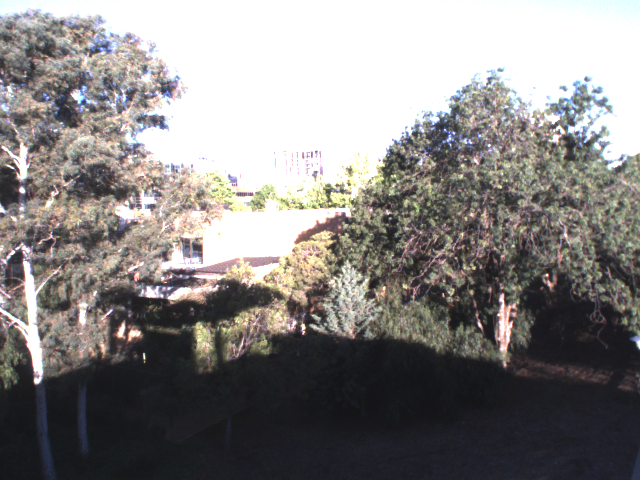} &
			\includegraphics[width=1\linewidth]{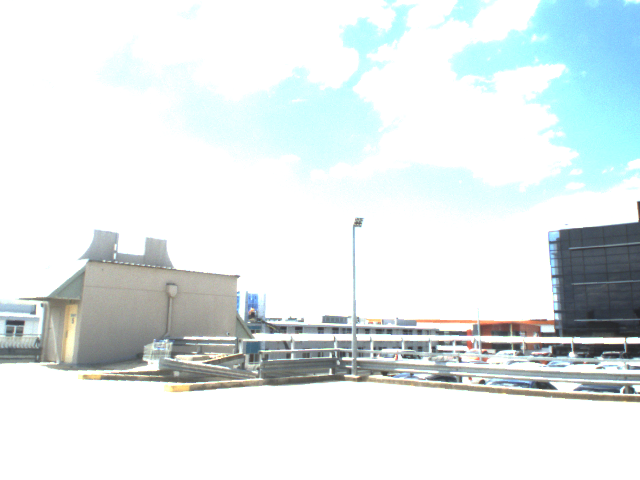} &
			\includegraphics[width=1\linewidth]{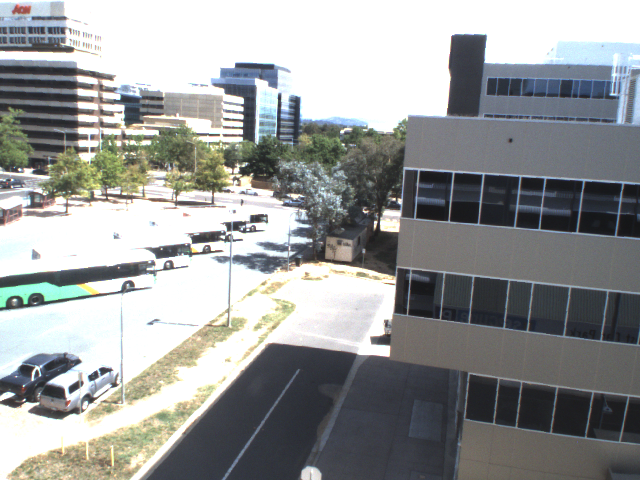}
			\vspace{0.5cm}
			\\
			\includegraphics[width=1\linewidth]{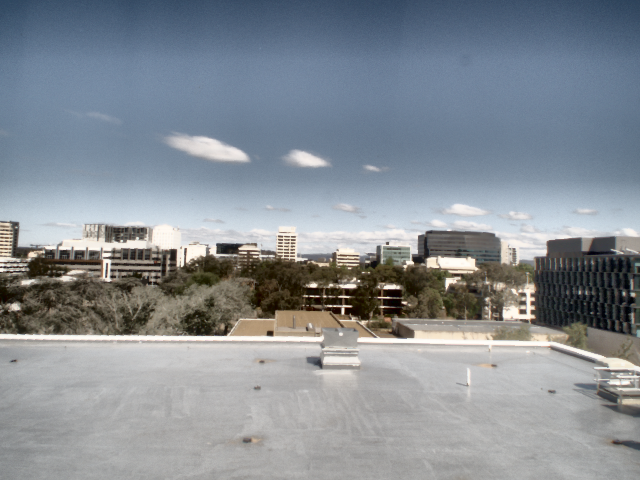} &
			
			\includegraphics[width=1\linewidth]{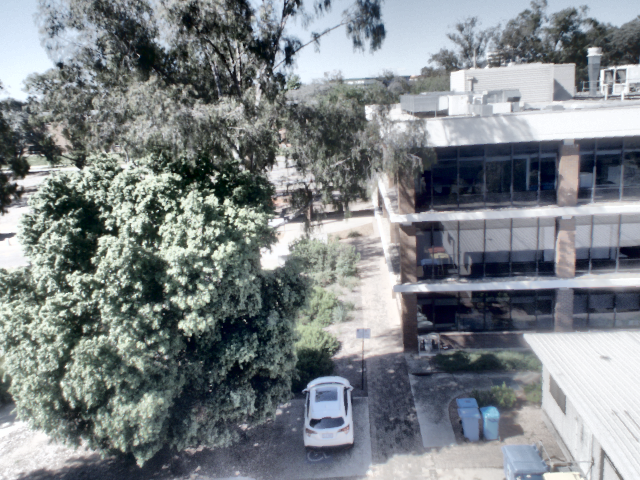} &
			
			\includegraphics[width=1\linewidth]{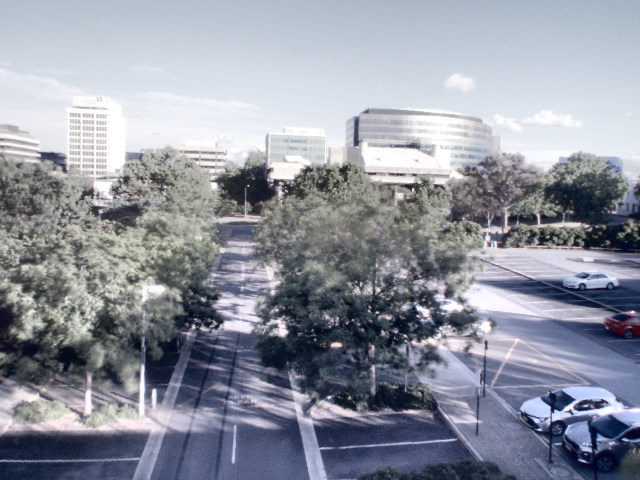} &
			
			\includegraphics[width=1\linewidth]{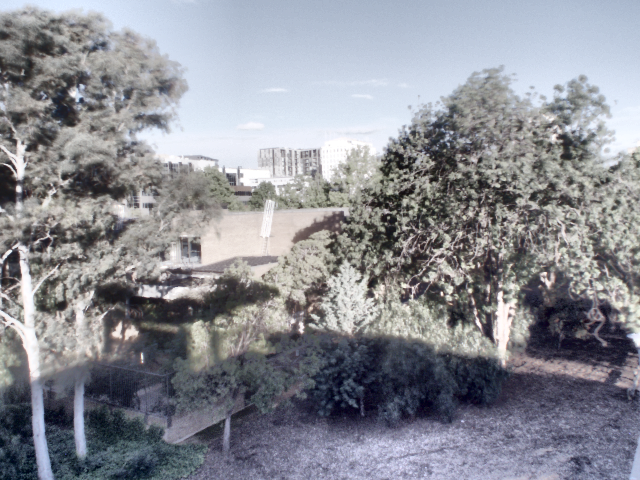} &
			
			\includegraphics[width=1\linewidth]{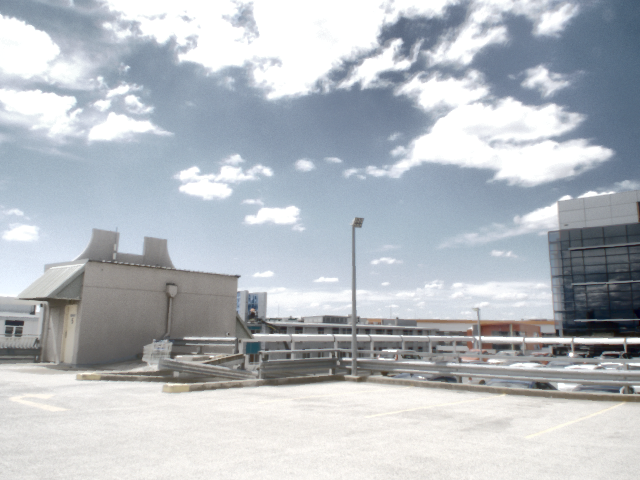} &
			
			\includegraphics[width=1\linewidth]{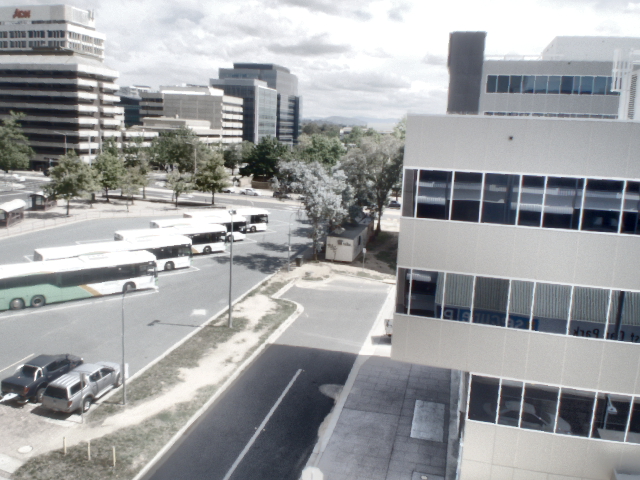}
\\
			{\fontsize{50}{50}\selectfont (a). city} &
			{\fontsize{50}{50}\selectfont (b). tree 1} &
			{\fontsize{50}{50}\selectfont (c). tree 2} &
			{\fontsize{50}{50}\selectfont (d). tree 3} &
			{\fontsize{50}{50}\selectfont (e). car park 1} &
			{\fontsize{50}{50}\selectfont (f). car park 2}
			\\
		\end{tabular}
	}
	\caption{\label{fig:HDR}Selected Images from Our HDR Hybrid Event-Frame Dataset. First row shows the low dynamic range frames and the second row shows the high dynamic range ground truth (with tone-mapping for display only).}
\end{figure*}

\subsection{Our HDR Hybrid \EventFrame Dataset}\label{sec: dataset list HDR}
Evaluating HDR reconstruction for hybrid \eventframe cameras requires a dataset including synchronised events, low dynamic range video and high dynamic range reference images.
The dataset associated with the recent work by~\cite{han2020neuromorphic} is patent protected and not publicly available.
Published datasets lack high quality HDR reference images, and instead rely on low dynamic range sensors such as the APS component of a DAVIS for groundtruth \cite{Stoffregen20eccv,Zhu18ral,Mueggler17ijrr}.
Furthermore, these datasets do not specifically target HDR scenarios.
DAVIS cameras used in these datasets also suffer from shutter noise (noisy events triggered by APS frame readout) due to undesirable coupling between APS and DVS components of pixel circuitry \cite{brandli2014240}.

\begin{figure}[H]
	\centering
	\includegraphics[width=0.7\linewidth]{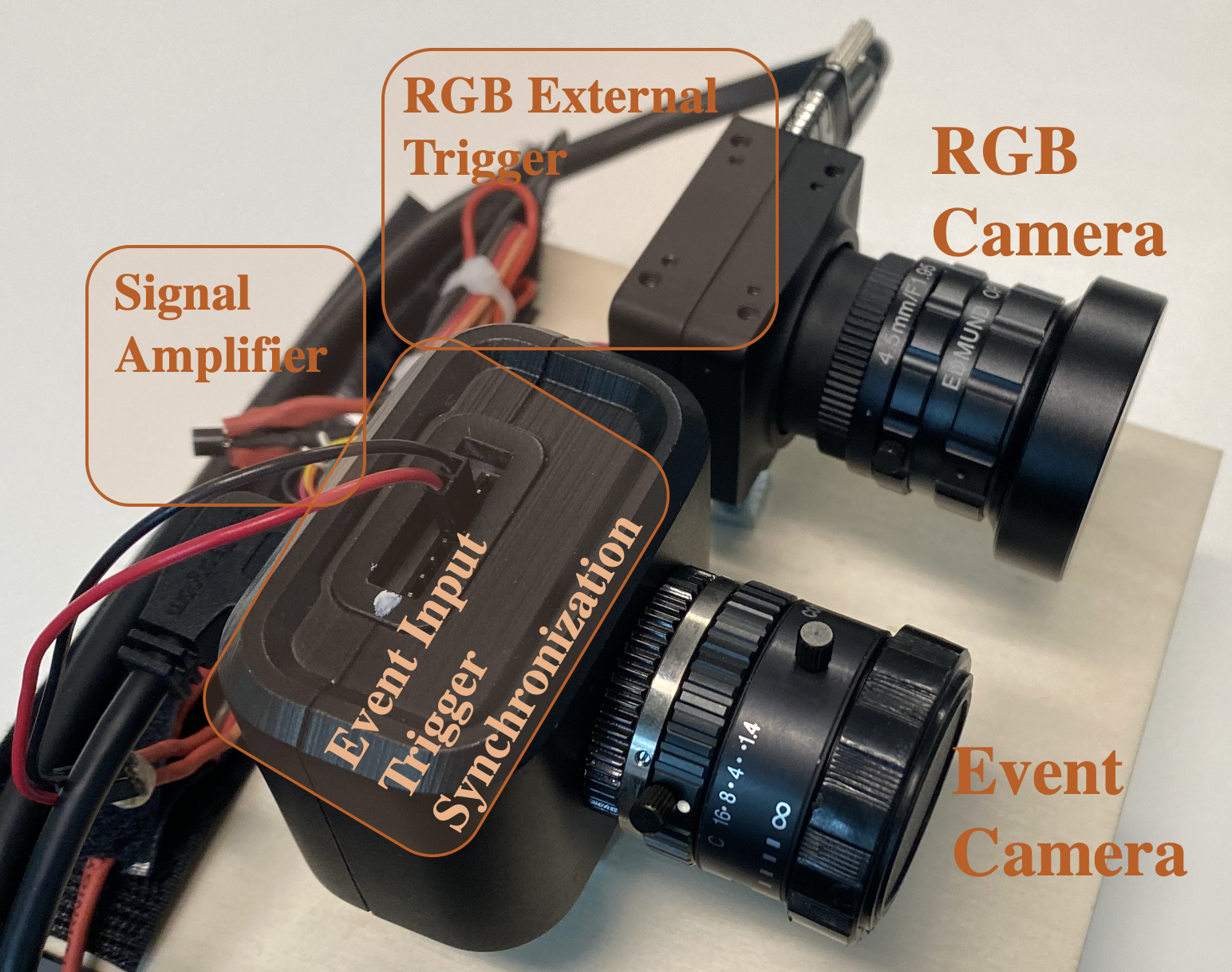}
	\caption{\label{fig:camera rig} Our Hybrid Event-Frame System}
\end{figure}

To address these limitations, we built a hybrid \eventframe camera system
consisting of two separate high quality sensors, a Prophesee event camera (VGA, 640$\times$480 pixels) and a \textit{FLIR} RGB frame camera (Chameleon3 USB3, 2048$\times$1536 pixels, 55FPS, lens of 4.5mm/F1.95), mounted side-by-side.
The stereo hybrid \eventframe camera prototype we built is shown in Fig. \ref{fig:camera rig}.
We use the camera system to collect events, frames and HDR reference images.
The dataset sequences focus on different HDR scenes with different camera motion speeds, which is challenging for all event-based HDR image reconstruction methods.

\begin{table*}
	\centering
	\caption{\label{tab:AHDR}Details of AHDR Dataset}
	\resizebox{1\textwidth}{!}{ 
		\begin{tabular}{ c | c | c | c | c | c}
			\midrule
			\midrule
			HDR Dataset & \# of images &  Speed & Description & HDR Scene & Reference Image
			\\
			\midrule
			\midrule
			\textbf{Dataset} & \multicolumn{4}{c}{\textbf{Artificial HDR sequences (AHDR)}}
			\\
			\midrule
			mountain $\times$ 2 & 150 & slow and fast &  mountain with road overlooking city &  dark grass and road & \checkmark
			\\
			\midrule
			lake $\times$ 2 & 200 & slow and fast & side of lake with trees and road during cloudy day &  trees and clouds & \checkmark
			\\
			\midrule
			\midrule
		\end{tabular}
	}
\end{table*}

We calibrated the hybrid system using a blinking checkerboard video and computed camera intrinsic and extrinsic matrices following \cite{heikkila1997four, zhang2000flexible}.
We synchronised the two cameras by sending an external signal from the frame camera to trigger timestamped zero magnitude events in the event camera.
We obtained a HDR reference image for quantitative evaluation of a sequence via traditional multi-exposure image fusion followed by an image warp to register the reference image with each frame.
The scene in the proposed dataset is chosen to be static and far away from the camera due to the 6-centimeter baseline between event and frame camera, so that SURF feature matching \cite{bay2006surf} and homography estimation are sufficient for the image registration.
But a more general scene can be used when event and camera sensors are placed closely on a circuit board in the future.
For a near field scene, registration of stereo event-frame camera data requires disparity estimation,
however, it is beyond the scope of the present paper that focuses on HDR reconstruction. By using far-field datasets, we are able to provide compelling evidence for the proposed algorithm without adding complexity to the problem.
Despite this, we believe that our proposed HDR dataset is currently the state-of-the-art event/frame HDR dataset and will encourage the development of more datasets and algorithms in the field.
The details of our proposed HDR \eventframe dataset are summarised in Table~\ref{tab:HDR} and Fig~\ref{fig:HDR}.

\subsection{Our AHDR Dataset}\label{sec: dataset list AHDR}
We also provide an artificial HDR (AHDR) dataset that was generated by simulating a low dynamic range (LDR) camera by applying an artificial camera response function and using the original images as HDR references.
We synthesised LDR images in this manner to provide additional data to verify the performance of our algorithm.

\begin{figure}
	\centering
	\begin{tabular}{c c}
		\\
		\includegraphics[width=.47\linewidth]{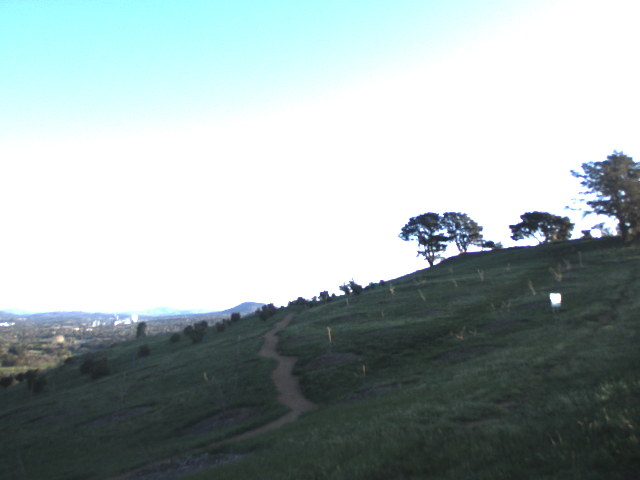} &
		\includegraphics[width=.47\linewidth]{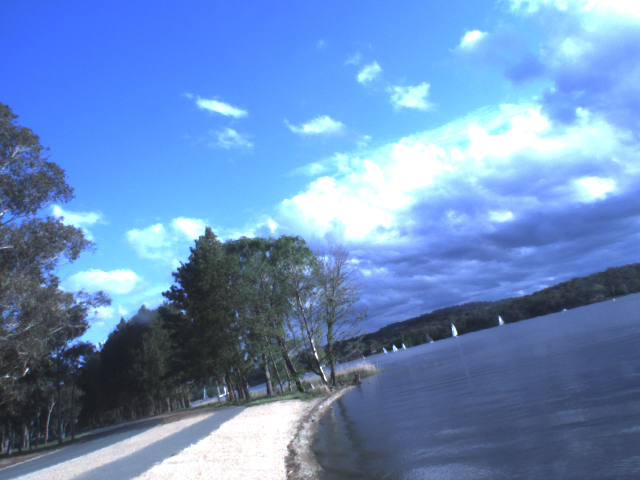} \\
		(a). mountain & (b). lake \\
	\end{tabular}
	\caption{\label{fig:AHDR}Selected Images from AHDR Dataset}
\end{figure}

For our AHDR dataset, we apply an artificial camera response function to RGB camera output frames to simulate a low dynamic range camera (see Fig. \ref{fig:ahdr_tonemap}).
Following the process introduced in Appendix B of the main paper, we experimentally determine the corresponding camera uncertainty function of the low dynamic range camera, where the resulting image noise covariance is high for the `cropped' intensity values. Details of the AHDR dataset are as shown in Table~\ref{tab:AHDR} and Fig.~\ref{fig:AHDR}.

\begin{figure}[H]
	\centering
	\includegraphics[width=0.95\linewidth]{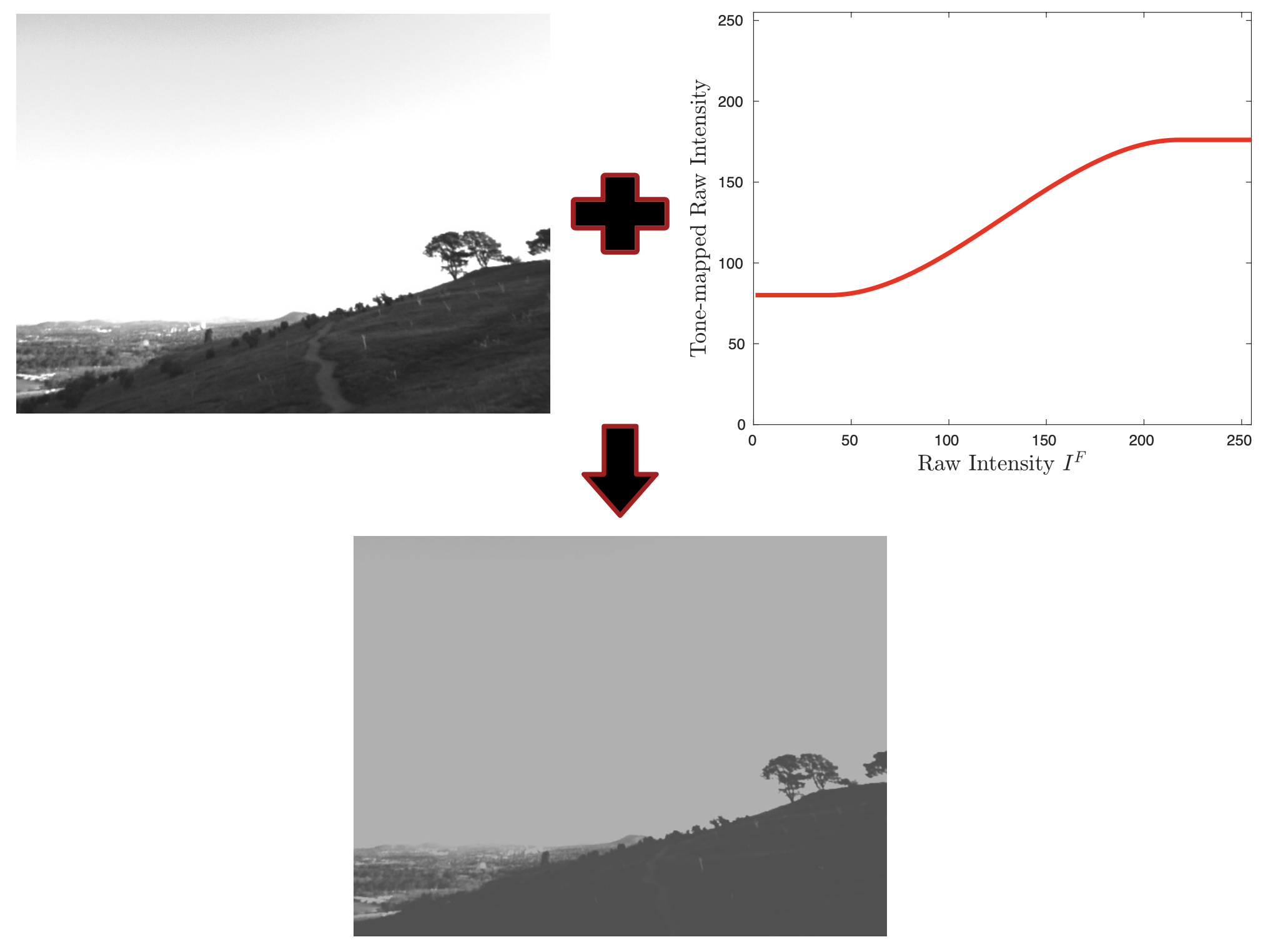}
	\caption{\label{fig:ahdr_tonemap} Process of generating our AHDR dataset. We simulate a low dynamic range camera by applying an artificial camera response function to real images.}
\end{figure}

\end{document}